\documentclass[runningheads]{llncs}

\usepackage{eccv}

\usepackage{eccvabbrv}

\usepackage{graphicx}
\usepackage{booktabs}

\usepackage[accsupp]{axessibility}  

\usepackage[pagebackref,breaklinks,colorlinks,citecolor=eccvblue]{hyperref}

\usepackage{amsmath,amsfonts,bm}

\def\eqref#1{equation~\ref{#1}}

\def\1{\bm{1}}

\def\rvx{{\mathbf{x}}}

\def\vf{{\bm{f}}}
\def\vg{{\bm{g}}}

\def\vw{{\bm{w}}}

\def\vy{{\bm{y}}}

\DeclareMathAlphabet{\mathsfit}{\encodingdefault}{\sfdefault}{m}{sl}
\SetMathAlphabet{\mathsfit}{bold}{\encodingdefault}{\sfdefault}{bx}{n}

\def\sD{{\mathbb{D}}}

\def\sG{{\mathbb{G}}}

\def\sR{{\mathbb{R}}}
\def\sS{{\mathbb{S}}}

\def\sY{{\mathbb{Y}}}

\newcommand{\E}{\mathbb{E}}

\DeclareMathOperator*{\argmax}{arg\,max}
\DeclareMathOperator*{\argmin}{arg\,min}

\usepackage[table]{xcolor}
\usepackage{tikz,pgfplots}
\usepgfplotslibrary{groupplots}

\definecolor{tableheader}{RGB}{228,224,221} 
\definecolor{defaultrow}{RGB}{242,240,255} 
\definecolor{ligtable}{RGB}{235,240,235} 
\definecolor{todo}{RGB}{200,100,100}

\makeatletter
\newcommand{\mylabel}[2]{#2\def\@currentlabel{#2}\label{#1}}
\makeatother

\begin{document}

\title{The Label Imitation Game: Turing Test Network for Zero-Shot Pseudo-Label Pruning}

\titlerunning{Turing Test Network}

\author{Brent A. Griffin\inst{1} \and
Jason J. Corso\inst{1,2}
}

\authorrunning{B.~Griffin and J.~Corso}

\institute{Voxel51, \email{\{brent,jason\}@voxel51.com}  \and
University of Michigan
}

\maketitle

\begin{abstract}
Foundation model pseudo-labeling---labeling data strictly via zero-shot inference---enables massive scale, but performance is undermined by hallucinations that evade standard thresholds.
To eliminate these errors, we introduce the Turing-inspired \textbf{Label Imitation Game (LIG)}, a framework that formalizes pseudo-label pruning as an adversarial interrogation. 
Rather than filtering labels via isolated thresholds, we use the LIG to train a \textbf{Turing Test Network (TTN)}, a task-agnostic ``judge'' that evaluates candidate pseudo-labels within a dataset-wide context. 
Experiments across four diverse datasets demonstrate the TTN’s robustness, consistently enhancing label accuracy for three state-of-the-art vision-language models without costly supervision or retraining. 
Crucially, we demonstrate that learned semantic-contextual logic is a robust alternative to spatial-geometric verification, enabling a unique zero-shot task transfer capability---a TTN trained strictly on image classification datasets can effectively prune complex object detection pseudo-labels. 
This pruning yields $F_1$-score gains of 28\% for the worst-performing baseline categories and 44\% with task-specific fine-tuning. 
Significantly, we also observe \textbf{Category Revival}, where the TTN pruning ``detoxifies'' the training signal for downstream models and enables them to recover from zero recall on transfer-vulnerable classes. 
The pre-trained TTN models and code are available at \url{https://github.com/voxel51/ttn}.


\end{abstract}

\section{Introduction}
\label{sec:intro}

\begin{figure}
	\centering
	\begin{minipage}{1\textwidth}
		\begin{picture}(0,0)
			\put(16,0){\rotatebox{0}{\bf \scriptsize VOC (GDINO)}}
			\put(96,0){\rotatebox{0}{\bf \scriptsize COCO (YOLOE)}}
			\put(196,0){\rotatebox{0}{\bf \scriptsize LVIS and BDD (YOLOW)}}
		\end{picture}
	\end{minipage}\\  \vspace{-0.05em}
	\begin{minipage}{0.016\textwidth}
		\begin{picture}(0,0)
			\put(0,25){\rotatebox{90}{\bf \scriptsize Reject}}
			\put(0,-40){\rotatebox{90}{\bf \scriptsize Accept}}
		\end{picture} \vspace{-1.17em}
	\end{minipage}
	\begin{minipage}{0.1925\textwidth}
		\includegraphics[width=0.975\textwidth]{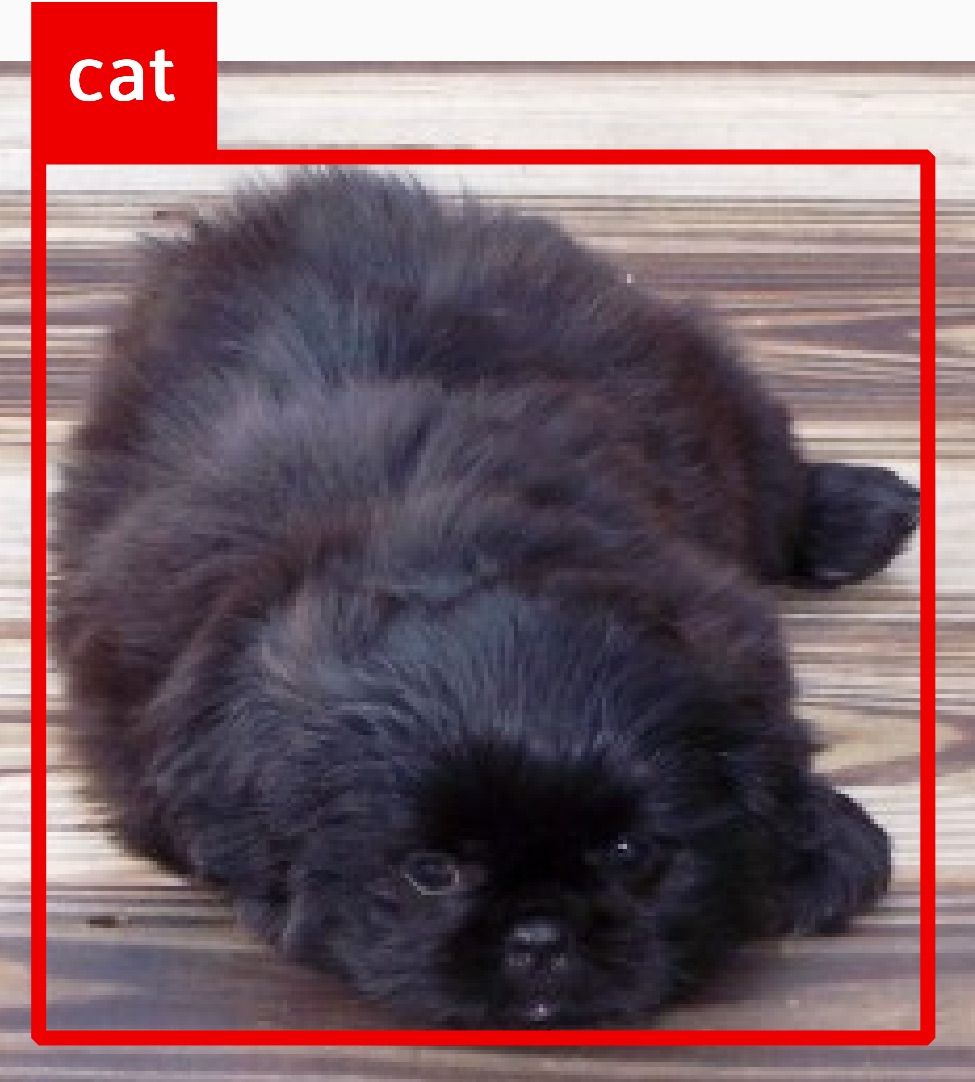}\\ \vspace{-1.1em}
		\includegraphics[width=0.975\textwidth]{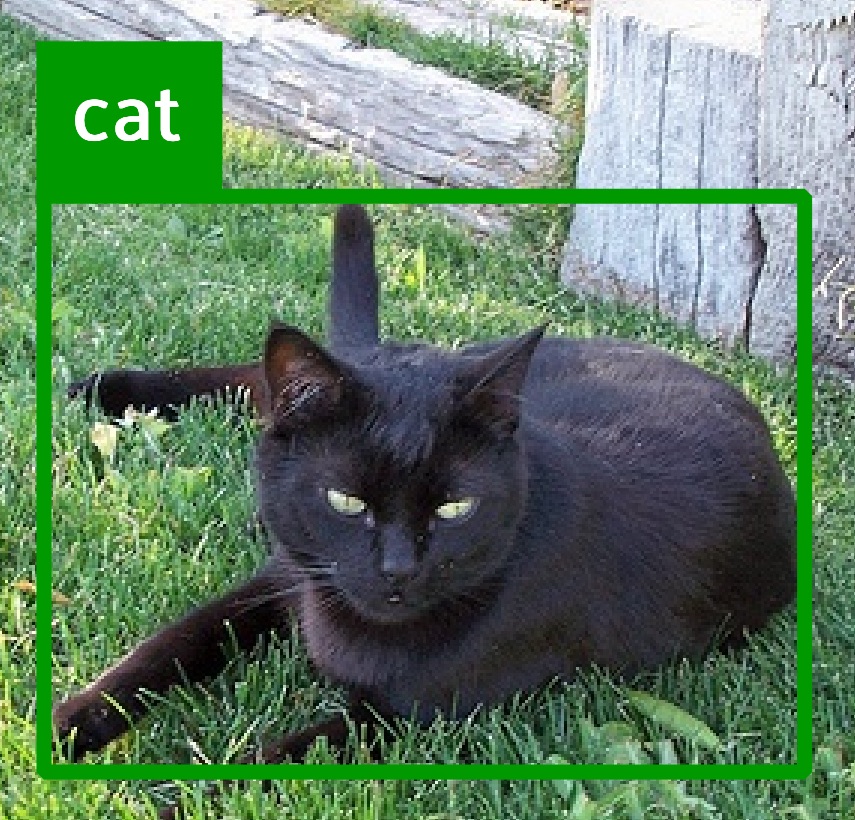} 
	\end{minipage} \hspace{-0.3em}
	\begin{minipage}{0.195\textwidth}
		\includegraphics[width=0.975\textwidth]{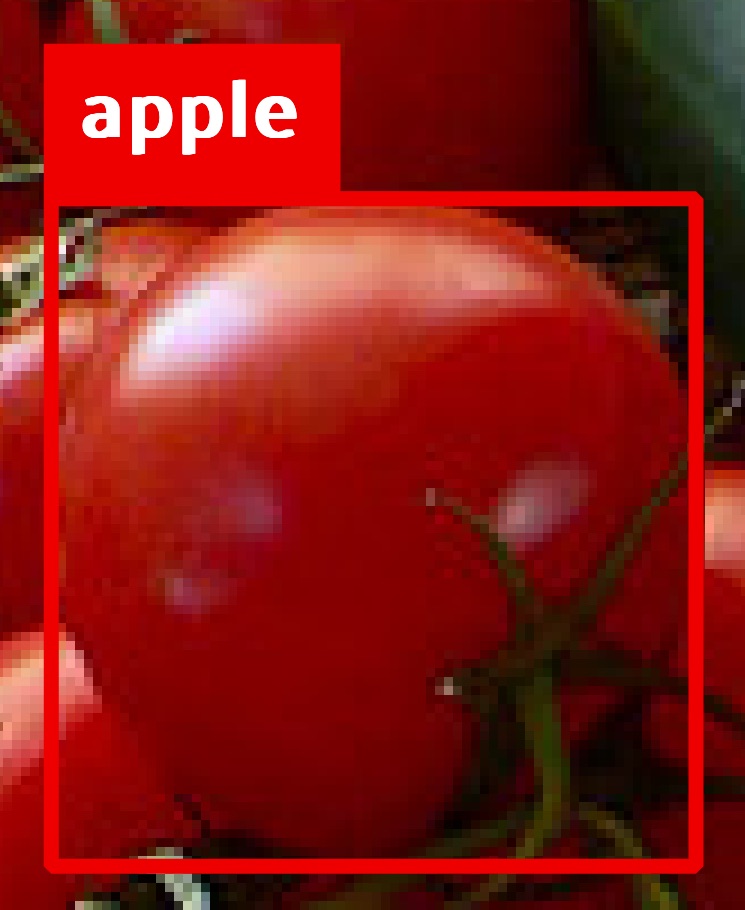} \\ \vspace{-1.1em}
		\includegraphics[width=0.975\textwidth]{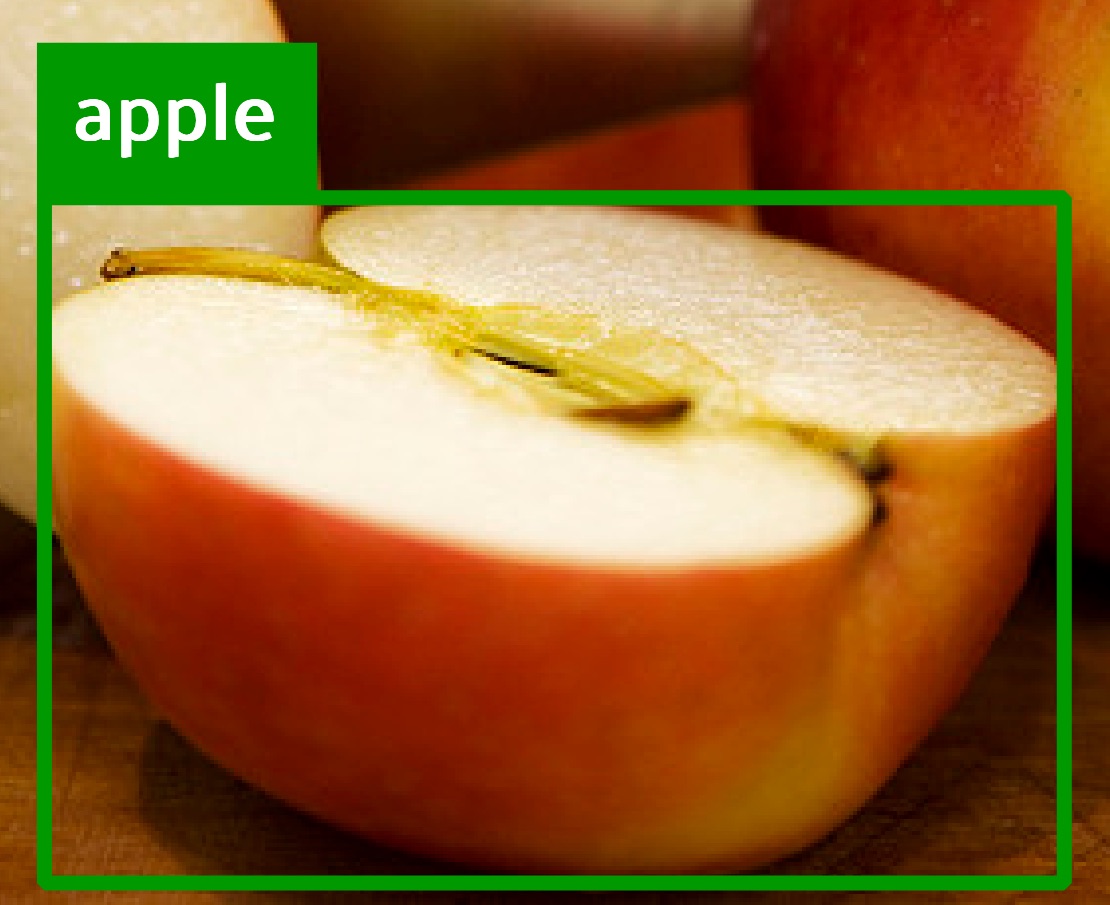}
	\end{minipage}%
	\begin{minipage}{0.077\textwidth} 
		\includegraphics[width=0.935\textwidth]{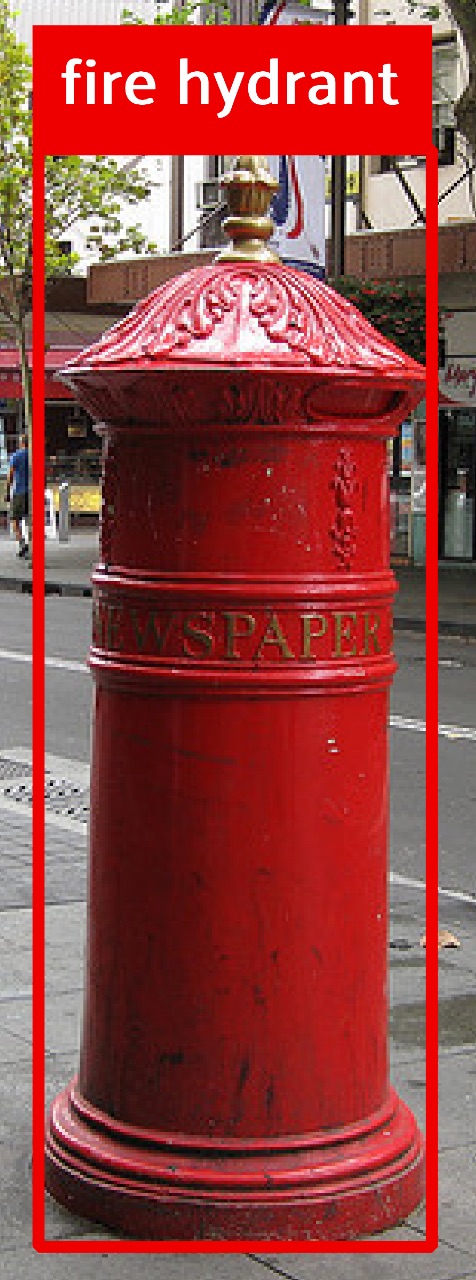} \\ \vspace{-1.1em}
		\includegraphics[width=0.935\textwidth]{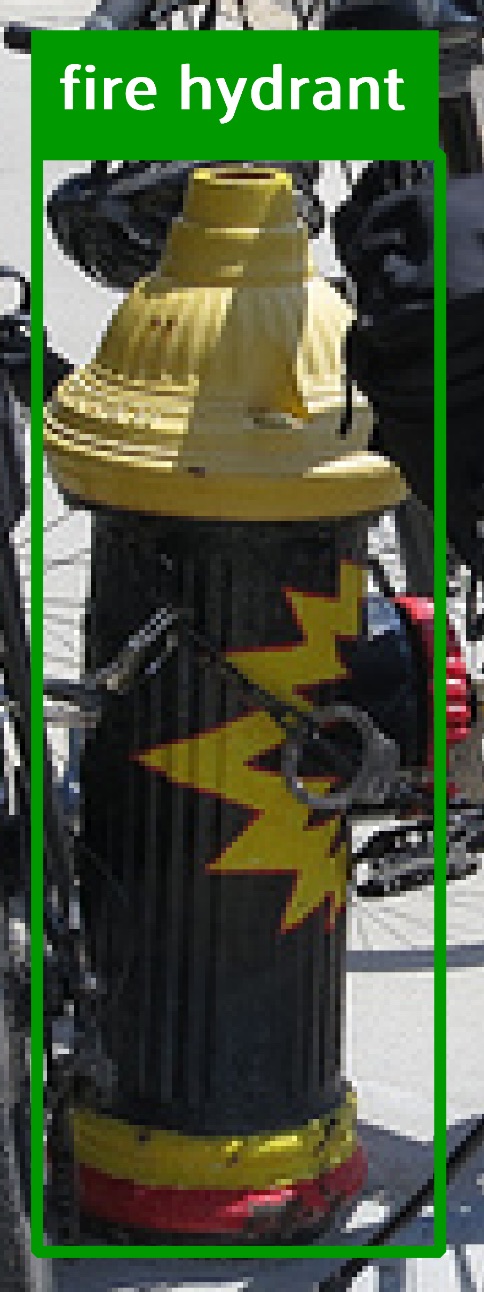}
	\end{minipage}  \hspace{-0.3em}
	\begin{minipage}{0.123\textwidth}
		\includegraphics[width=1\textwidth]{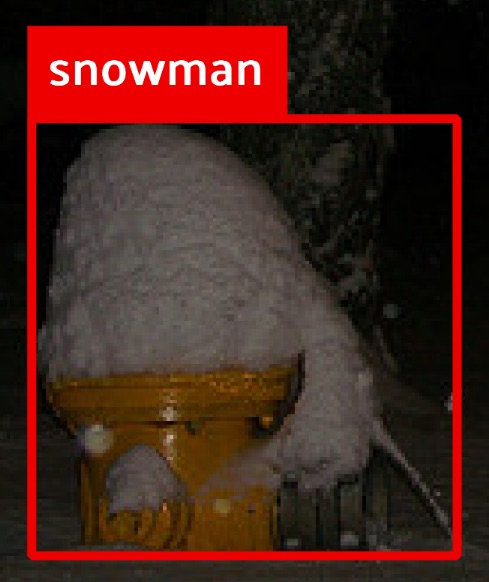} \\ \vspace{-1.1em}
		\includegraphics[width=1\textwidth]{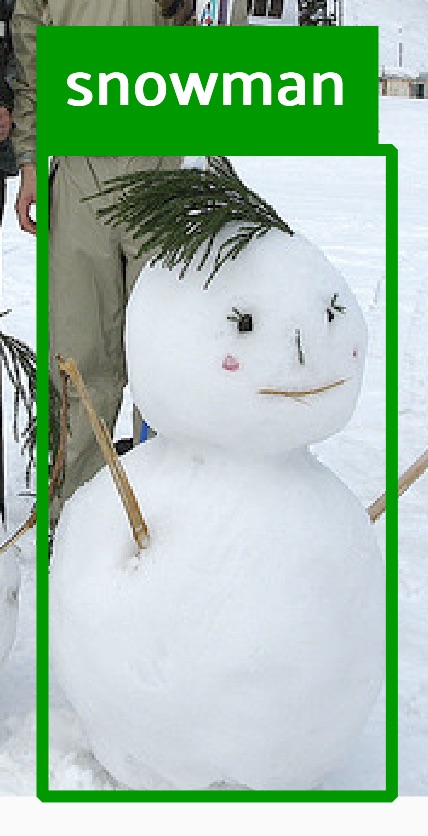}
	\end{minipage} \hspace{-0.16em}
	\begin{minipage}{0.171\textwidth} 
		\includegraphics[width=0.97\textwidth]{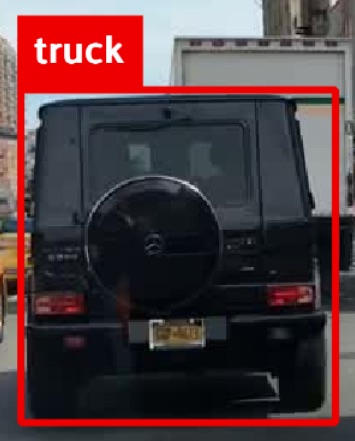} \\ \vspace{-1.1em}
		\includegraphics[width=0.97\textwidth]{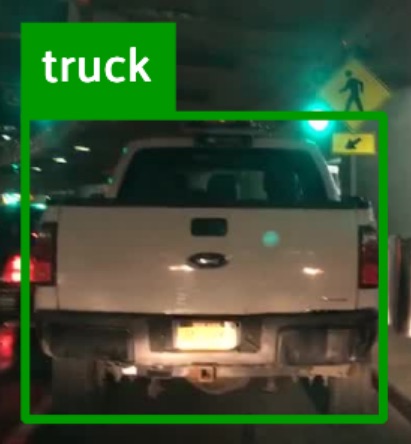}
	\end{minipage}%
	\begin{minipage}{0.151\textwidth}
		\includegraphics[width=1\textwidth]{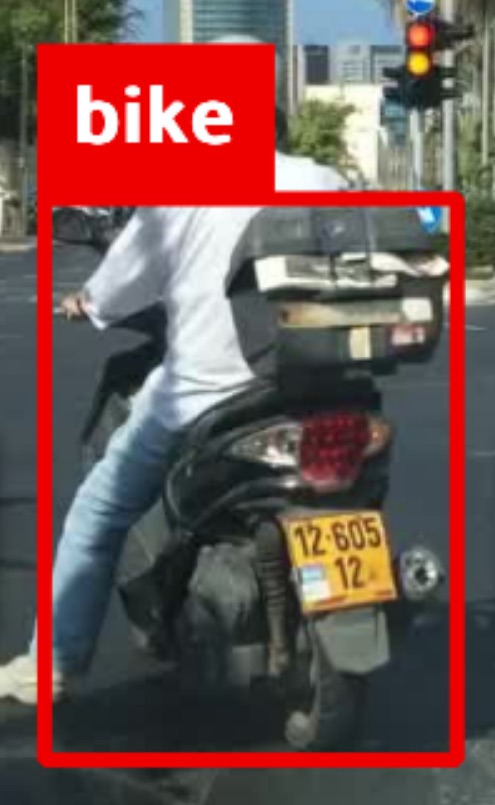} \\ \vspace{-1.1em}
		\includegraphics[width=1\textwidth]{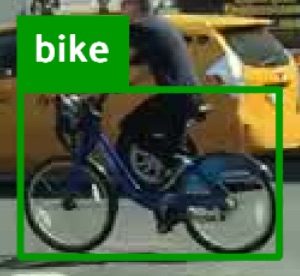}
	\end{minipage} \\ \vspace{-0.05em}
	\begin{minipage}{1\textwidth}
		\includegraphics[width=0.992\textwidth]{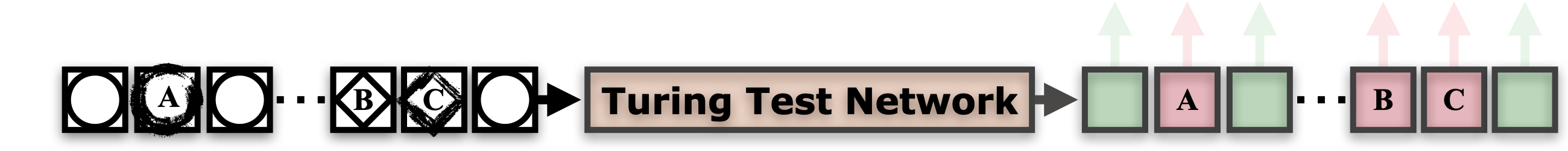}
	\end{minipage} 
				\begin{minipage}{1\textwidth}
		\begin{picture}(0,0)
			\put(25.5,22){\rotatebox{0}{\bf \scriptsize Pseudo-Label Tokens}}
			\put(162,22){\rotatebox{0}{\bf \scriptsize Judge}}
			\put(260.5,22){\rotatebox{0}{\bf \scriptsize Accept/Reject}}
		\end{picture}
	\end{minipage}
	\vspace{-2.5em}
	\caption{{\bf  Zero-Shot Pseudo-Label Pruning}. 
		A single Turing Test Network (TTN) trained strictly on image-classification (bottom) finds and rejects systemic VLM hallucinations across diverse detection datasets and pseudo-label architectures (top) while accepting accurate labels (middle). 
		TTN rejects labels for spatial inaccuracy (\textbf{A}), semantic inconsistency (\textbf{B}), or both (\textbf{C}).
		Visualizations generated using FiftyOne \cite{moore2020fiftyone}.
	}
	\label{fig:front}
\end{figure}
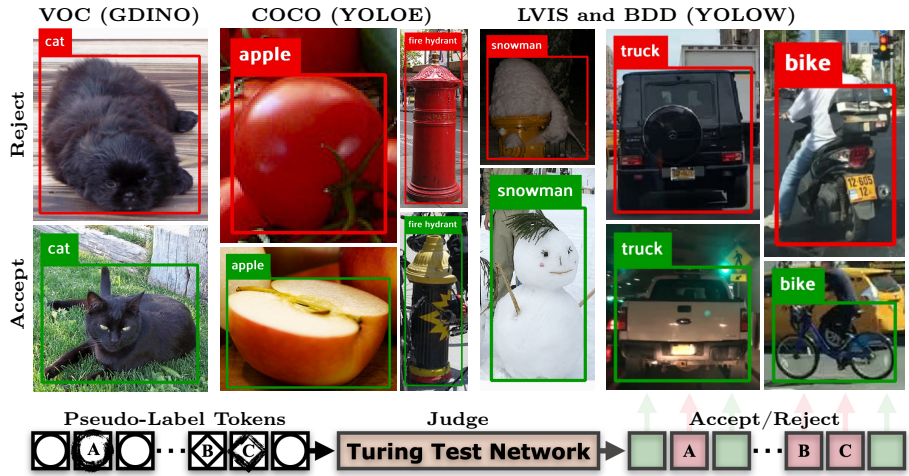

Can machines annotate data?
Foundation vision-language models (VLMs), while powerful for tasks like object detection \cite{gu2022openvocabulary,owl23_vlm,Fu_2025_CVPR}, often suffer systemic hallucinations when applied to open-world problems \cite{li-etal-2023-evaluating,zhou2024analyzing,Gunjal_Yin_Bas_2024}.
For pseudo-labeling \cite{pseudo13}, these hallucinations add training noise that degrades downstream model performance relative to expensive (but accurate) human labeling \cite{popp25,nagase2025annotation}.
Accordingly, traditional strategies use fixed thresholds to mitigate noise \cite{griff25AL,pmlr-v80-jiang18c,NEURIPS2018_a19744e2}.
However, our experiments find this often discards authentic, low-confidence labels while retaining structured errors that are ``confidently wrong.''
To address this, there is a critical need for a lightweight, post-hoc ``interrogator'' that differentiates accurate labels from hallucinations, \textit{ideally} without any costly human supervision or changes to the underlying VLM architecture and downstream training pipeline.

Inspired by Alan Turing’s ``Imitation Game'' \cite{Turing1950}, we propose the Label Imitation Game (LIG), which formalizes pseudo-label pruning as an adversarial interrogation where a judge must distinguish between accurate labels and hallucinations. 
To operationalize this judge, we introduce the Turing Test Network (TTN)---a transformer-based module that rejects erroneous labels by evaluating their contextual plausibility against a dataset-wide reference set (see \cref{fig:front}).

Our primary contributions are:
\textbf{(i) The Label Imitation Game (LIG):} a novel framework that formalizes label pruning as an adversarial interrogation between a judge and candidate labels (\cref{sec:method}); 
\textbf{(ii) The Turing Test Network (TTN):} a lightweight, transformer-based architecture that \textit{automates} the role of the LIG judge by utilizing non-masked self-attention to evaluate semantic and spatial consistency without class labels~(\cref{sec:ttn}); and 
\textbf{(iii) Zero-Shot Task Transfer:} a cross-task evaluation demonstrating that a TTN trained strictly on image classification can prune object detection pseudo-labels without \textit{any} human input, yielding $F_1$-score gains of 28\% and Category Revival---enabling downstream detectors to recover from zero recall on up to 27 classes~(\cref{sec:exp}).

\section{Related Work}
\label{sec:relate}

\noindent \textbf{Foundation Model Pseudo Labeling} is a scalable alternative to costly manual annotation \cite{griff25AL,popp25,nagase2025annotation}. 
After earlier work established the benefits of unified Vision-Language Models (VLMs) \cite{zhou2020unified}, grounded language-image pre-training \cite{Li_2022_CVPR} reformulated object detection as a phrase grounding task---aligning text phrases to image regions. 
This VLM innovation enables massive zero-shot transfer by treating arbitrary detection categories as contextualized text prompts, which is enhanced in open-vocabulary detectors like Grounding DINO \cite{shilong2024} and YOLO-World \cite{Cheng_2024_CVPR}. 
These VLMs are powerful tools for ``data engines'' \cite{mcity25,Xiao_2024_CVPR}, with innovative processes like using high-confidence predictions in weakly-augmented views to generate pseudo-labels for strongly-augmented ones \cite{NEURIPS2020_06964dce}.
Nonetheless, VLMs are fallible and generally limited by a precision-recall trade-off \cite{griff25AL}.

\vspace{0.2em}
\noindent \textbf{Hallucinations and Label Noise}---predicting entities that are visually absent or contextually inconsistent \cite{li-etal-2023-evaluating,zhou2024analyzing,Gunjal_Yin_Bas_2024}---are the Achilles heel of VLMs, even occurring at high confidence thresholds \cite{chen24eccv} (\cref{fig:qualconf}). 
Thankfully, VLM hallucinations can be treated as a specialized label noise problem \cite{FREIRE2024112544}. 
While noise-robust methods commonly use fixed loss thresholds \cite{pmlr-v80-jiang18c,NEURIPS2018_a19744e2}, self-adaptive and class-balanced approaches can explicitly identify noisy data \cite{sheng24ECCV}.
For pseudo-label pruning, specifically, other techniques maximize re-labeling accuracy by selecting subsets that maintain neighborhood prediction confidence \cite{NEURIPS2023_ebb6bee5} or iteratively refining boxes to improve localization \cite{he2023pseudolabelcorrectionlearningsemisupervised}. 
Other spatial-geometric verification methods use embeddings to prune redundant detections \cite{featurenms}, but rely on local feature similarity.
To expand beyond local heuristics, we leverage a dataset-wide semantic context to selectively prune inaccurate pseudo-labels, addressing global inconsistencies that traditional methods overlook.

\vspace{0.2em}
\noindent \textbf{Adversarial Interrogation} provides a novel framework for pruning. 
The concept of a Visual Turing Test was first proposed to evaluate scene interpretation through binary interrogation \cite{geman2014pnas}, while later adversarial variants for dialogue use a ``Judge'' to discriminate between human and machine responses \cite{gao2021adversariallylearnedturingtestdialog}.
This draws a direct analogy to generative adversarial networks \cite{Goodfellow2014GenerativeAN}, which popularized the minimax game between a generator and discriminator.
This emerging ``LLM-as-a-Judge'' paradigm provides a scalable alternative to human evaluation by using foundation model reasoning to assess output quality across diverse tasks \cite{li2024llmsasjudgescomprehensivesurveyllmbased,NEURIPS2023_91f18a12}.
As an alternative to high-cost LLMs, adversarial filters use an iterative greedy algorithm to remove classification data biases \cite{pmlr-v119-bras20a}, but this approach precludes reasoning for complex spatial-semantic interrogation.
We address this with the Label Imitation Game (LIG), a general framework that formalizes label pruning as an adversarial interrogation independent of source models. 
By decoupling evaluation from specific task heads, the LIG enables our lightweight Turing Test Network to act as a task-agnostic judge that interrogates the semantic plausibility of candidate pseudo-labels within a global reference context.

\section{Problem Formulation: The Label Noise Challenge}

Assume a model generates pseudo-labels $\hat{\sY}$ on an unlabeled dataset $\sD_\text{\tiny U}=\{\rvx_i\}^M_{i=1}$.
Here, data examples $\rvx_i$ are drawn i.i.d. from an underlying distribution $P$, and $\hat{\sY} = \{ \hat{\vy}_i\}^M_{i=1}$ comprises variable-length label sets ($|\hat{\vy}_i| \geq 0$) for each $\rvx_i$.

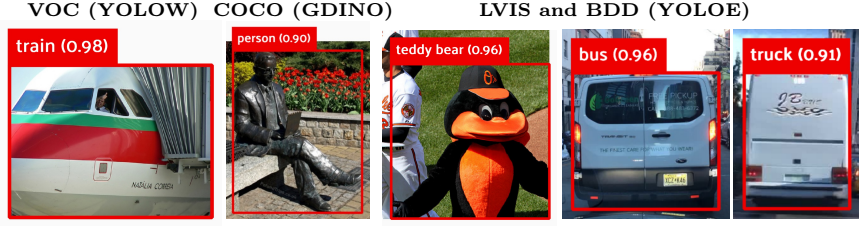
\begin{figure}[t!]
	\centering
	\begin{minipage}{1\textwidth}
		\begin{picture}(0,0)
			\put(20,0){\rotatebox{0}{\bf \scriptsize VOC (YOLOW)}}
			\put(90,0){\rotatebox{0}{\bf \scriptsize COCO (GDINO)}}
			\put(190,0){\rotatebox{0}{\bf \scriptsize LVIS and BDD (YOLOE)}}
		\end{picture}
	\end{minipage}\\  \vspace{-0.15em}
	\begin{minipage}{0.245\textwidth}
		\includegraphics[width=0.975\textwidth]{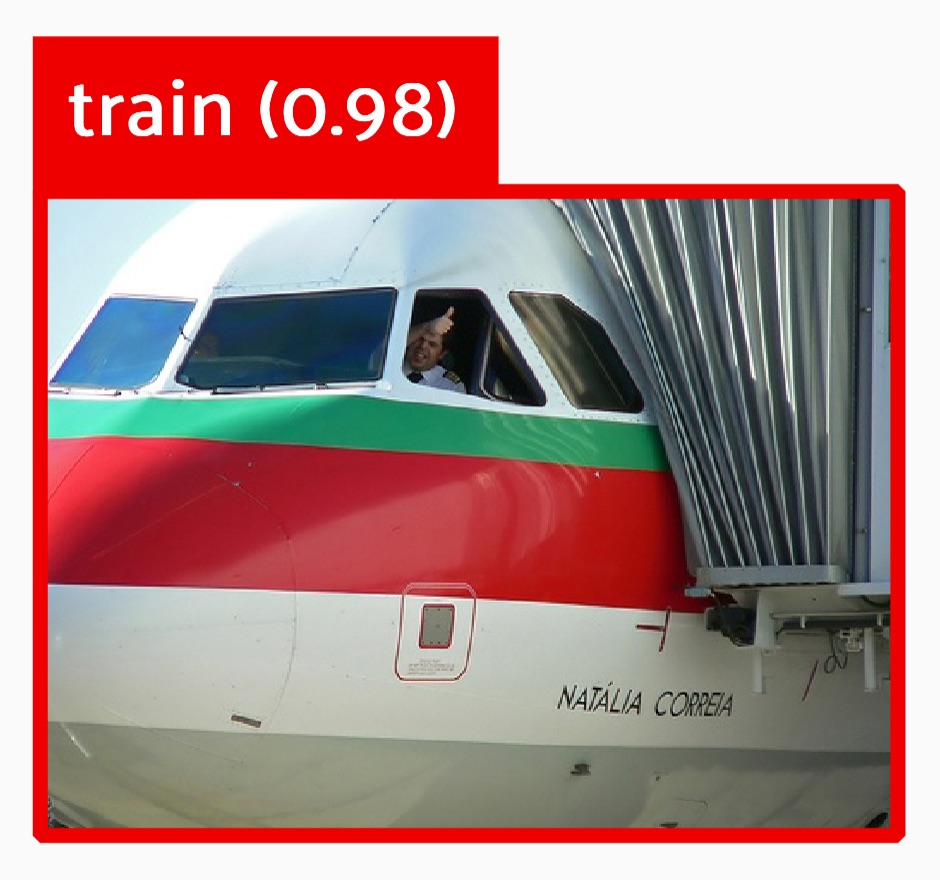}
	\end{minipage} \hspace{-0.7em}
	\begin{minipage}{0.159\textwidth} 
		\includegraphics[width=0.975\textwidth]{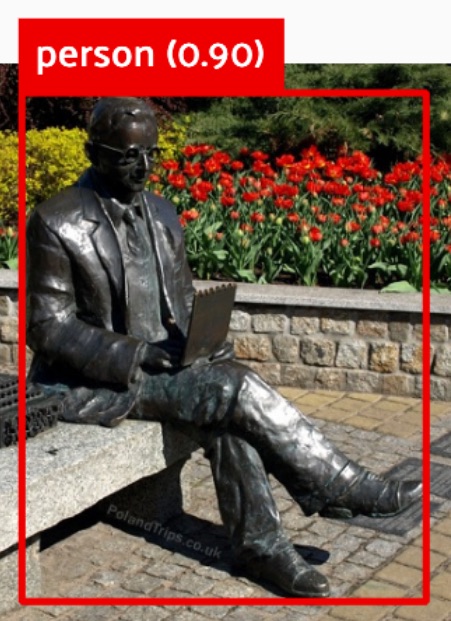}
		\vspace{-0.9em}
	\end{minipage}  \hspace{-0.3em}
	\begin{minipage}{0.194\textwidth} 
		\includegraphics[width=0.975\textwidth]{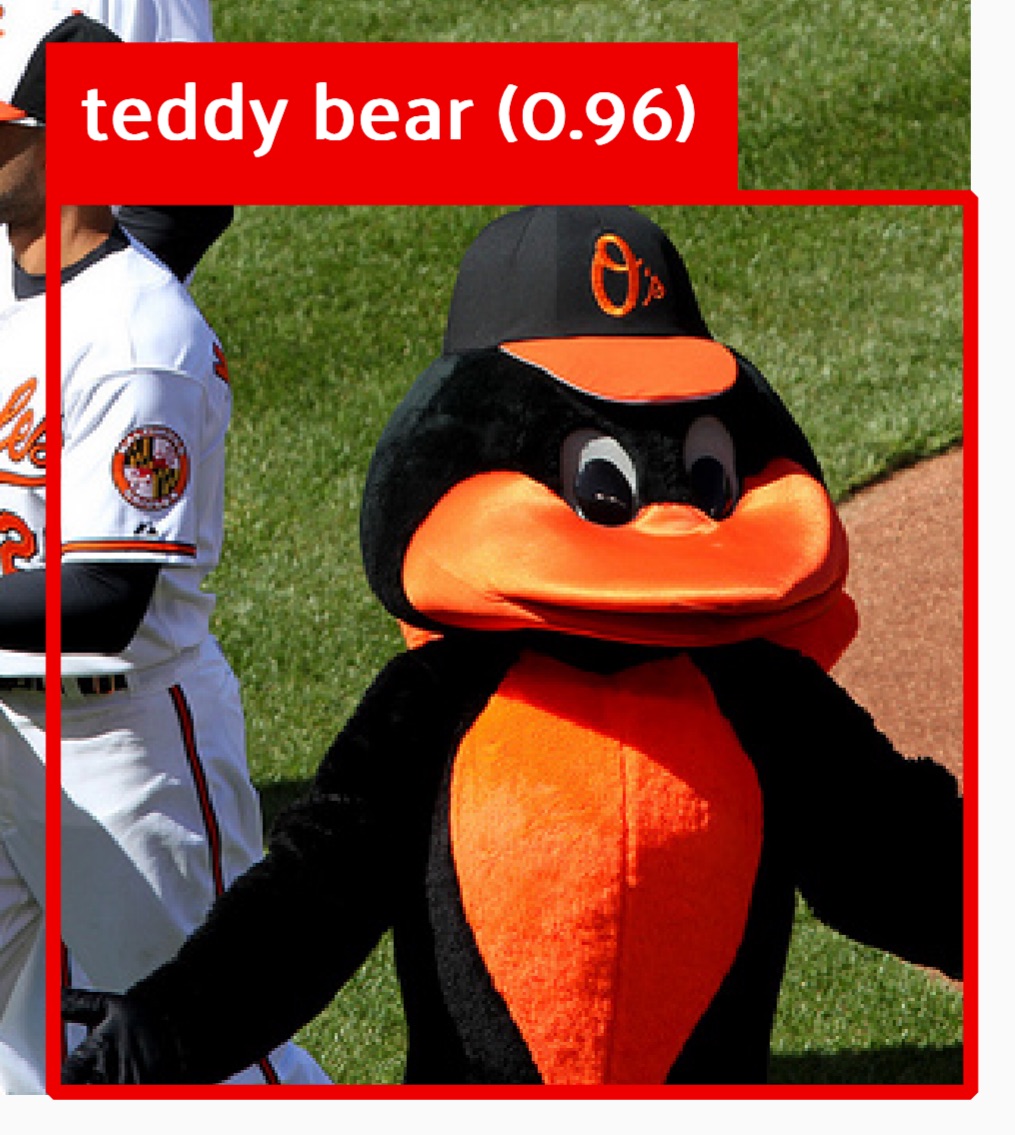}
		\vspace{-1.5em}
	\end{minipage} \hspace{-0.6em}
	\begin{minipage}{0.185\textwidth} 
		\includegraphics[width=0.975\textwidth]{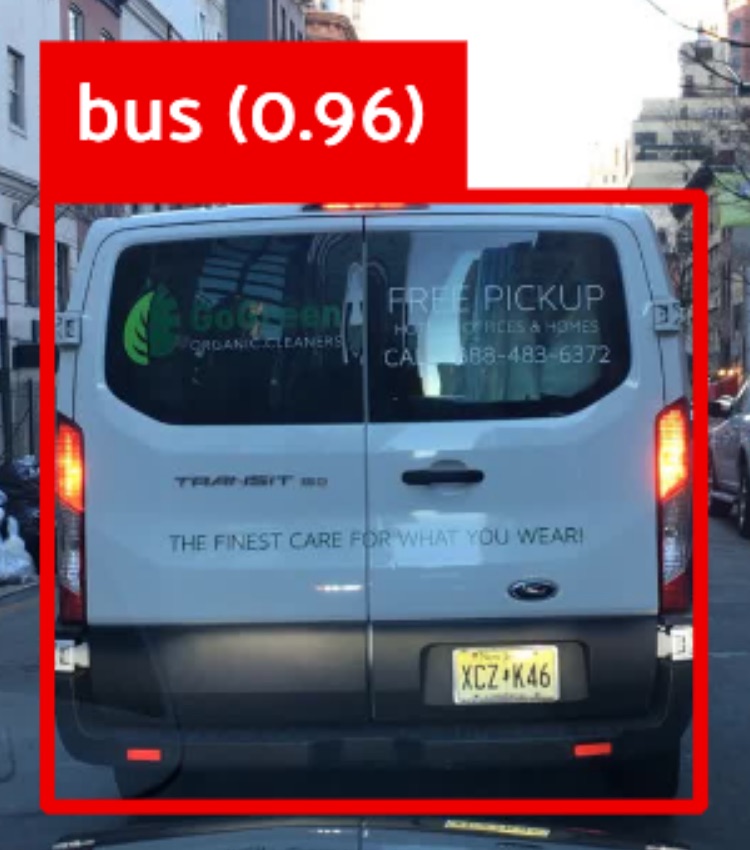}
		\vspace{-1.2em}
	\end{minipage}%
	\begin{minipage}{0.139\textwidth}
		\includegraphics[width=0.975\textwidth]{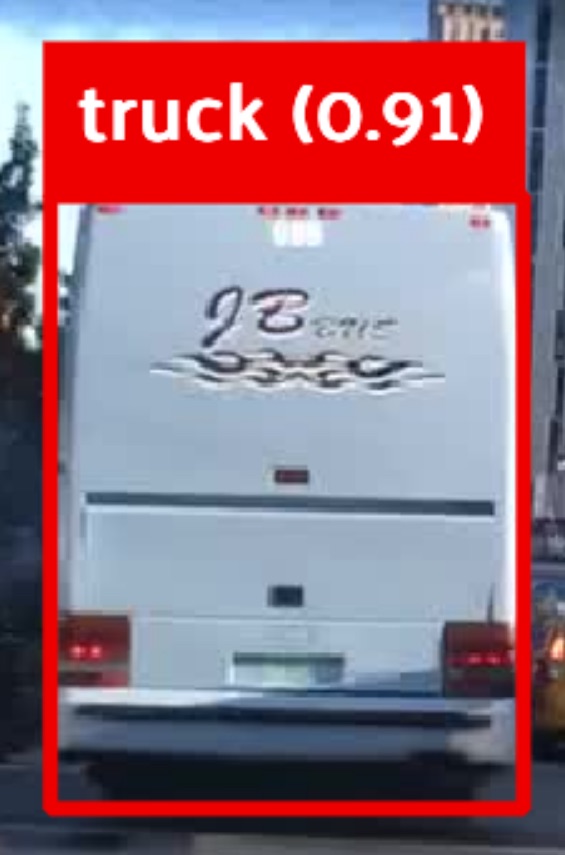}
	\end{minipage} 
	\vspace{-1em}
	\caption{{\bf  TTN Zero-Shot Rejections on High-Confidence Pseudo-Labels}
	}
	\label{fig:qualconf}
\end{figure}

In practice, pseudo-label accuracy is highly variable and, even with high-confidence thresholds, models often produce hallucinations (see \cref{fig:qualconf}).
Our goal is to (i) selectively prune these inaccurate labels to increase precision, while (ii) preserving accurate labels to maintain recall.
Accordingly, we evaluate and prune $\hat{\sY}$ to select label subset $\tilde{\sY} = \{ \tilde{\vy}_i \}^M_{i=1}$ and formulate the \textbf{pseudo-label pruning} problem as
\begin{equation}
	\begin{aligned}
		\argmax_{ \tilde{\sY} \subseteq \hat{\sY}} \E_{\rvx, \vy \sim P} [ F_1 (\vy; \tilde{\vy} )], \\
	\end{aligned}
	\label{eq:f1}
\end{equation}
where $\vy$ is the ground truth and the $F_1$ score is the harmonic mean of precision and recall. 
Specifically, $F_1 = {\textstyle 2\frac{\text{precision}\cdot\text{recall}}{\text{precision + recall}}} \in [0,1]$,
precision {\footnotesize $=\frac{\text{TP}}{\text{TP+FP}}~\in~[0,1]$}, recall {\footnotesize $=\frac{\text{TP}}{\text{TP+FN}} \in [0,1]$},
true positives ({\footnotesize TP}) is the number of accurate pseudo-labels, and false negatives ({\footnotesize FN}) is the number of $\vy$ labels without a corresponding {\footnotesize TP}.
In plain words, {\bf precision} is the frequency of pseudo-labels being correct, {\bf recall} is the frequency of ground-truth instances being correctly pseudo-labeled, and the goal of \cref{eq:f1} is maximizing the {\bf $F_1$ score}, which emphasizes label performance across \textit{both} metrics ($F_1=1$ for perfect precision and recall).

Pseudo-labeled datasets are commonly used for \textbf{downstream model training}.
However, if left unpruned, pseudo-labels corresponding to hallucinations will induce \textbf{inductive interference} that misleads downstream model convergence. 
Thus, we combine pruned pseudo-labels $\tilde{\sY}$ with $\sD_\text{\tiny U}$ to generate the ``detoxified'' dataset $\tilde{\sD} = \{ (\rvx_i, \tilde{\vy}_i) \}^M_{i=1}$.
We evaluate $\tilde{\sD}$ by reformulating pruning from \cref{eq:f1} explicitly for downstream model training as
 \begin{align}
 	\argmin_{\tilde{\sY} \subseteq \hat{\sY} } \E_{\rvx, \vy \sim P} [ \ell (\rvx, \vy; f_{( \tilde{\sD} )} ) ],
 	\label{eq:downstream}
 \end{align}
 where $\ell$ is the task-specific loss function and $f_{( \tilde{\sD} )}$ is a model trained on $\tilde{\sD}$.
 In plain words, the goal of \cref{eq:downstream} is to prune pseudo-labels ($\tilde{\sY}\subseteq \hat{\sY}$) to train a downstream model ($f$) that most accurately predicts ground-truth labels ($\vy$) for unseen images ($\rvx$) drawn from the underlying data distribution ($P$).

\section{Adversarial Interrogation via the Label Imitation Game}
\label{sec:method}

The Turing test evaluates a machine's ability to exhibit intelligent behavior indistinguishable from that of a human \cite{Turing1950}.
Likewise, our goal is to evaluate which pseudo-labels are indistinguishable from those of a human annotator, implicitly denoting high accuracy.
We formalize this as the \textbf{Label Imitation Game} (LIG).

We formulate label tests in the context of a game involving a set of \textbf{Ideal} and \textbf{Candidate} labels and a \textbf{Judge}.
The Judge is given all labels and must determine which belong to each set.
We use LIG to train the Turing Test Network, which later serves as the Judge at inference to reject and prune pseudo-labels, solving $\tilde{\sY} \subseteq \hat{\sY}$ in Eqs. (\ref{eq:f1}) \& (\ref{eq:downstream}).
We use a tokenized representation (\cref{sec:token}) for the LIG (\cref{sec:lig}), but these games are general to the original labels and data.

\subsection{Tokenizing Labels \& Data}
\label{sec:token}

The transformer architecture (the \textit{de facto} standard for NLP \cite{attention}) operates on a tokenized representation of data to iteratively compare context across all inputs before determining final outputs.
To utilize this iterative comparison capability for the LIG, we fuse labels $\vy$ and data $\rvx$ into tokens, specifically
\begin{equation}
	\begin{aligned}
 g (\rvx, \vy) = \{ \vg_1, \vg_2, \dots, \vg_n \}, \\
	\end{aligned}
	\label{eq:g}
\end{equation}
where $g$ is a function operating on $\rvx$ and $\vy$, $n = |\vy|=|g (\rvx, \vy)|$, and each output token $\vg_i$ is a lower-dimension representation than $\rvx$ (we specify $g$ in \cref{sec:ttn}).
Extending \cref{eq:g}, we can generate arbitrary tokens across $m$ data examples as
\begin{equation}
	\sG = \{ \vg_1, \vg_2, \dots, \vg_N \} = \{ g (\rvx_i, \vy_i) \}^m_{i=1}.
	\label{eq:setg}
\end{equation}

Using \cref{eq:setg}, we reformulate label pruning as \textbf{token pruning}, i.e., selecting $\tilde{\sG} \subseteq \hat{\sG}$, where $\hat{\sG} = \{ \hat{\vg}_i \}^{\hat{N}}_{i=1} = \{ g (\rvx_i, \hat{\vy}_i) \}^M_{i=1}$ and $\hat{N}= | \hat{\sY} |$.
Furthermore, because there is a one-to-one correspondence between tokens and labels, tokens can map directly back to labels as $g^{-1}: \tilde{\sG} \rightarrow \tilde{\sY}$ in \cref{eq:f1} for bijective equivalence.

\subsection{Label Imitation Games}
\label{sec:lig}

For \textbf{Game 1}, we use \cref{eq:setg} to generate a set of tokens  $\sG^* = \{ \vg^*_i \}^{N^*}_{i=1}$ from accurately labeled data $\{\rvx_i, \vy^*_i \}^{m^*}_{i=1}$ and $\hat{\sG} = \{ \hat{\vg}_i \}^{\hat{N}}_{i=1}$ from pseudo-labeled data $\{ \rvx_i, \hat{\vy}_i \}^m_{i=1}$.
The \textbf{Judge's} objective is to predict which tokens in the combined set correspond to pseudo-labels.
We formulate this \textbf{Pseudo-Label Game} as
\newtheorem{imitategame}{Game}
\begin{imitategame}[PLG]
	Given $\sG^* \cup \hat{\sG}$, identify which tokens belong to $\hat{\sG}$.
\label{plg}
\end{imitategame}
The intuition of PLG is if a \textbf{Candidate} token $\hat{\vg}_i \in \hat{\sG}$ is distinguishable from the \textbf{Ideal} set $\sG^*$, its corresponding pseudo-label is likely inaccurate and should be rejected.
Conversely, as $\hat{\vg}_i$ approaches $\sG^*$ accuracy, it becomes effectively indistinguishable.
In fact, experiments show that carefully chosen $\hat{\vg}_i$ can even serve as Ideal tokens $\sG^*$ (we call this ``Self-Referential Pruning'' in \cref{sec:zs}).

For \textbf{Game 2}, assume each token is associated with a latent category $c_i \in \mathcal{C}$ from its underlying label and data.
Extending \cref{eq:setg}, we denote $c_i \ \forall \ \vg_i$ as
\begin{equation}
	\label{eq:c}
		 \sG = \{ \vg^{(c_1)}_1, \vg^{(c_2)}_2, \dots, \vg^{(c_N)}_N \}= \{ g (\rvx_i, \vy_i) \}^m_{i=1}.
\end{equation}
In this game, by construction, the \textbf{Ideal} tokens $\sG^{(c^*)}$ are homogeneous w.r.t. a target category $c^*$, s.t. $c_i = c^*, \forall \ \vg_i^{(c_i)} \in \sG^{(c^*)}$.
Conversely, \textbf{Candidate} tokens $\sG^{(\neg c^*)}$ are strictly disjoint from $c^*$, s.t. $c_i \neq c^*, \forall \ \vg_i^{(c_i)} \in \sG^{(\neg c^*)}$.
The \textbf{Judge's} objective is to predict which tokens in the combined set belong to disjoint categories $c_i \neq c^*$.
We formulate this \textbf{Out-of-Category Game} as
	\begin{imitategame}[OCG]
	Given $\sG^{(c^*)} \cup \sG^{(\neg c^*)}$, identify which tokens belong to $\sG^{(\neg c^*)}$.
	\label{ocg}
\end{imitategame}
Notably, the OCG Judge is not informed of the target category $c^*$.
To ensure the objective is not underdetermined, we implement OCG s.t. $| \sG^{(c^*)} | > | \sG^{(\neg c^*)} |$.
This constraint provides sufficient referential density for the Judge to infer target identity, resolve categorical ambiguity, and distinguish the Ideal tokens from Candidates.
The intuition of OCG is if a token $\vg_i \in \sG^{(\neg c^*)}$ is distinguishable from the ideal set $\sG^{(c^*)}$, its corresponding pseudo-label is semantically inconsistent or noisy and should be rejected.
A substantial benefit of OCG is generating numerous training examples for the Turing Test Network (details in \cref{sec:ttn}).

Finally, denoting PLG~\ref{plg} and or OCG~\ref{ocg} as $h$, we find $\tilde{\sY} \subseteq \hat{\sY}$ in \cref{eq:f1} as
\begin{equation}
	\tilde{\sY} = g^{-1}(\tilde{\sG}) = g^{-1} \big(  h ( \sG^* \cup  \hat{\sG}) \cap  \hat{\sG} \big), 
\label{eq:h}
\end{equation}
where $\tilde{\sG}$ are Candidate pseudo-label tokens $\hat{\vg}_i \in \hat{\sG}$ accepted by $h$ when provided $\sG^* \cup  \hat{\sG}$ and $g^{-1}$ maps the tokens back to their original labels (\cref{eq:setg}).

\section{Learning to Judge: The Turing Test Network}
\label{sec:ttn}

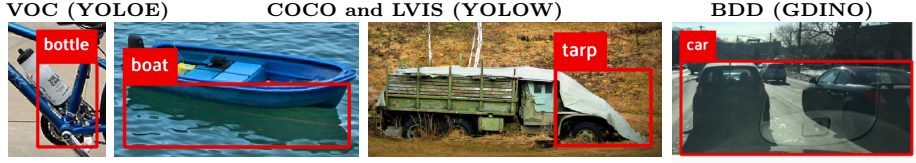
\begin{figure}[t!]
	\centering
	\begin{minipage}{1\textwidth}
		\begin{picture}(0,0)
			\put(0,0){\rotatebox{0}{\bf \scriptsize VOC (YOLOE)}}
			\put(98,0){\rotatebox{0}{\bf \scriptsize COCO and LVIS (YOLOW)}}
			\put(265,0){\rotatebox{0}{\bf \scriptsize BDD (GDINO)}}
		\end{picture}
	\end{minipage}\\  \vspace{-0.1em}
	\begin{minipage}{0.108\textwidth}
		\includegraphics[width=0.975\textwidth]{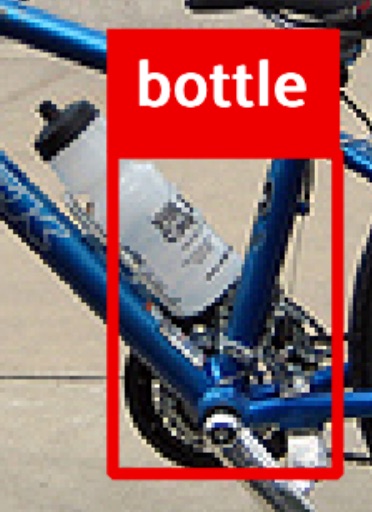}
	\end{minipage} \hspace{-0.4em}
	\begin{minipage}{0.271\textwidth} 
		\includegraphics[width=0.975\textwidth]{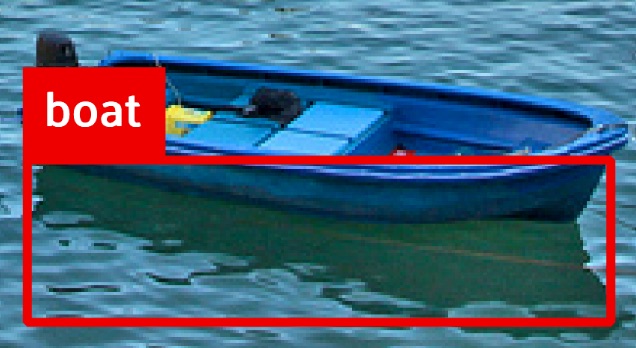}
	\end{minipage}  \hspace{-0.5em}
	\begin{minipage}{0.325\textwidth} 
		\includegraphics[width=0.975\textwidth]{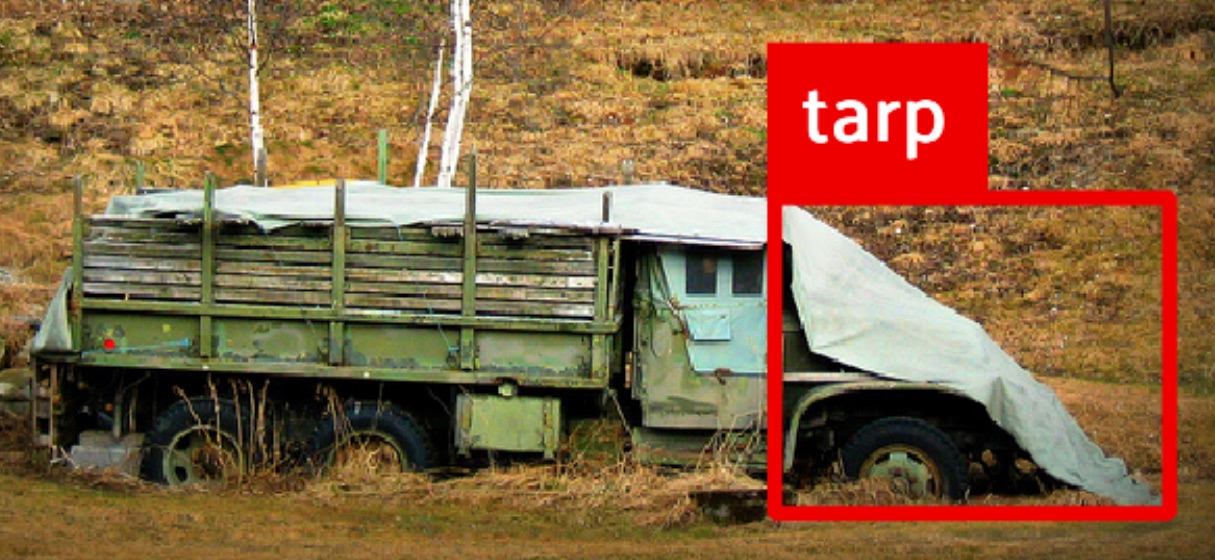}
	\end{minipage} \hspace{-0.53em}
	\begin{minipage}{0.276\textwidth} 
		\includegraphics[width=0.975\textwidth]{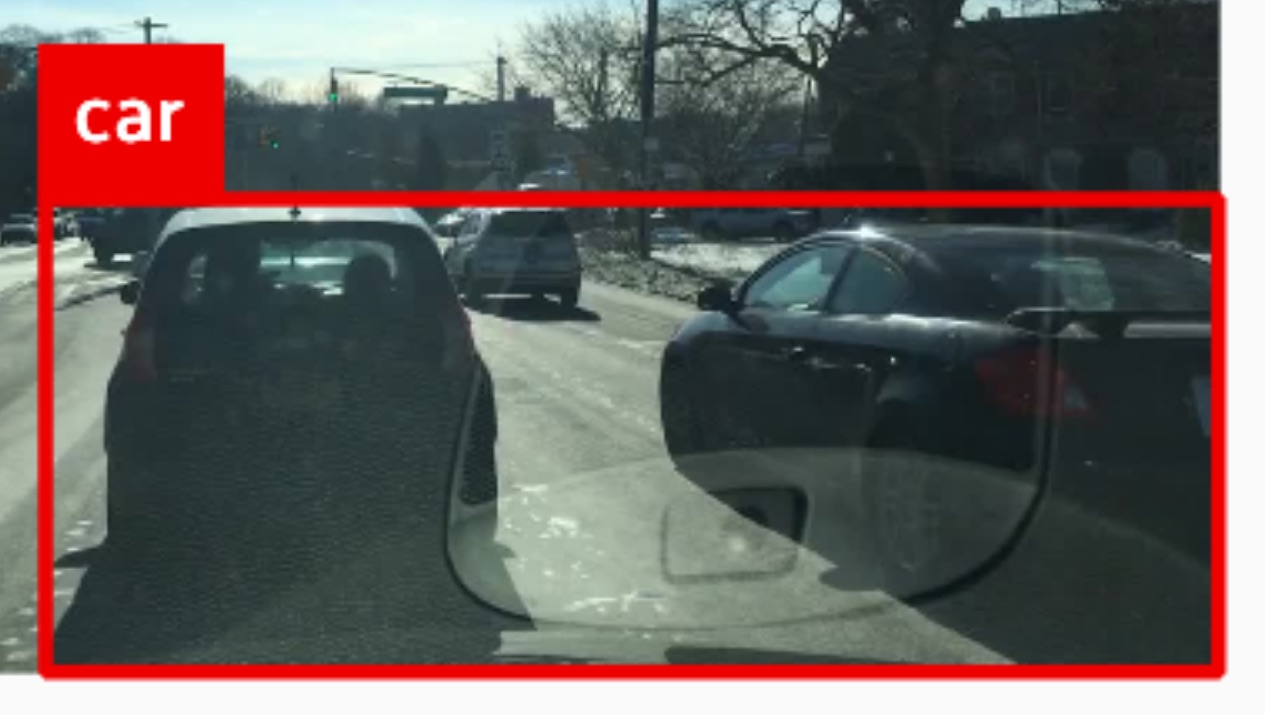}
		\vspace{-0.25em}
	\end{minipage}%
	\vspace{-0.85em}
	\caption{{\bf  TTN Zero-Shot Rejections on Poorly-Localized Pseudo-Labels}
	}
	\label{fig:qualbox}
\end{figure}

We introduce the \textbf{Turing Test Network} (TTN) to \textit{automate} the role of LIG Judge.
Specifically, we employ a transformer-based discriminator trained on a LIG-based binary cross-entropy loss to identify label errors (\cref{sec:train}).
We train the baseline TTN strictly on classification data but in a class-agnostic manner, withholding categorical labels across 1,303 classes from eight datasets to ensure the TTN learns latent semantic relationships rather than basic class mapping.
After training, the TTN exhibits an emergent capacity to build and evaluate semantic (\cref{fig:qualconf}) and spatial context (\cref{fig:qualbox}) solely from input sequences, enabling zero-shot transfer to object detection without supervision (\cref{sec:zs}).
Finally, we introduce a task-specific variant (TTN$_\text{\tiny D}$) fine-tuned on detection data, pseudo-labels, and $\leq 100$ human labels per class for enhanced performance~(\cref{sec:ft}). 

\vspace{0.2em} \noindent \textbf{Network Architecture}:
TTN's architecture consists of (i) a frozen CLIP \cite{clip} feature extractor that encodes label-defined image patches as $\vf_i \in \sR^{\text{768}}$ and (ii) a custom-learned tokenizer $\vg_i(\vf_i) \in \sR^{\text{512}}$ that enables the (iii) transformer-based reasoning block to evaluate each token and output Accept/Reject logits $z_i$ (\cref{fig:schematic}).
For full details, please refer to \cref{sec:supnet} in the Supplementary Material.

\subsection{Training via the Label Imitation Game}
\label{sec:train}

We train the TTN by minimizing a weighted binary cross-entropy (BCE) loss over LIG outcomes.
For a sequence of $n = | \sG |$ tokens, the objective function is
\begin{equation}
\mathcal{L}_\text{\tiny BCE}(\vw) := - \sum_{i=1}^{n} \Big[ p_w y_i \cdot \log \big( \sigma (z_i) \big) + (1-y_i) \cdot \log \big(1 - \sigma(z_i) \big) \Big],
\label{eq:loss}
\end{equation}
where $\vw$ are trainable parameters of the tokenizer and transformer, 
$y_i \in \{0,1\}$ is the ground-truth indicator assigning $y_i = 0$ to the \textbf{Accept} set and $y_i=1$ to the \textbf{Reject} set,  
$p_w$ is a positive weight applied to Reject training examples (conceptually, pruning aggression), 
$z_i$ is the output logit per token $\vg_i \in \sG$,
and $\sigma(z_i) = \frac{1}{1 + e^{-z_i}}$ is the model's estimated probability that $\vg_i$ should be rejected.

\begin{figure}[t!]
	\centering
	\includegraphics[width=0.975\textwidth]{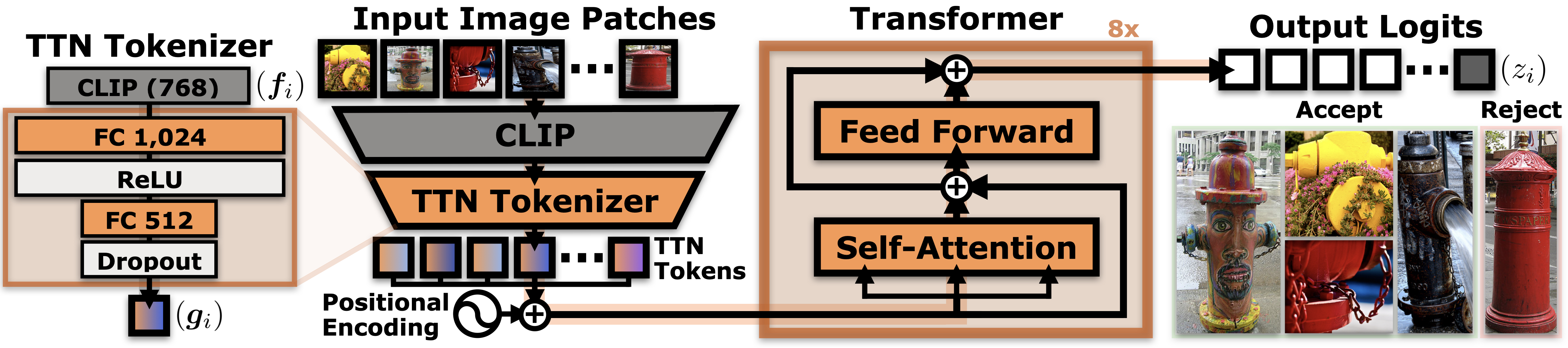}
	\vspace{-0.8em}
	\caption{{\bf Turing Test Network (TTN) Architecture}.
		The TTN processes image patches via a TTN Tokenizer and an eight-layer Transformer.
		Using non-masked self-attention, TTN evaluates relational and sequential context to produce Accept/Reject logits ($z_i$) for each input without any class labels, enabling zero-shot task transfer.
	}
	\label{fig:schematic}
\end{figure}

To ensure the TTN generalizes across diverse failure modes, we construct training batches by sampling tokens $\vg_i$ from a categorical \textbf{training distribution} $P(\sS)$ over the following four mutually exclusive states $\sS \in \{1,2,3,4\}$:
\begin{enumerate}
	\item[] $\sS=$\textcolor{white}{\tiny a}\mylabel{s:1}{$1$} (Ideal): High-fidelity human-label token with target category $c^*$.
	\item[] $\sS=~$\mylabel{s:2}{$2$} (OCG-Human): Human-label token with disjoint category $c_i \neq c^*$.
	\item[] $\sS=~$\mylabel{s:3}{$3$} (OCG-Pseudo): Pseudo-label token with disjoint category $c_i \neq c^*$.
	\item[] $\sS=~$\mylabel{s:4}{$4$} (PLG): Pseudo-label token of any fidelity and target category $c^*$.
	\end{enumerate}
By construction, $y_i=1$ if $\sS=1$ and $y_i=0$ otherwise in \cref{eq:loss}.
We denote the sampling probability for each state as $\pi_k = P(\sS=k)$, where $\sum \pi_k =1$.
This formulation enables control of LIG difficulty by adjusting the training failure distribution.
For OCG ($\sS = \{2,3\}$), all disjoint categories $c_i \neq c^*$ have an equal chance of being selected, but we only use one disjoint category per game.

\vspace{0.2em}
\noindent \textbf{Training Protocol}:
Each training sequence includes references for context prior to candidates. 
Specifically, we assign the first 10 positions sampled exclusively from the Ideal state ($\sS=1$) to the \textbf{Reference Set}. 
To ensure a robust reference context, we limit Reference Sets to only include categories with $N > 10$ instances.
The final position is reserved for the \textbf{Candidate Slot}, which is sampled from the full distribution $P(\sS)$.
This forces the TTN to utilize the Reference Set's context to interrogate the Candidate's fidelity. 
While experiments indicate the TTN can handle an arbitrary number of Reference and Candidate Slots, we utilize this 10:1 ratio for all training results.
Notably, a disjoint category may still serve as a Reject Candidate ($\sS \neq 1$) even if it fails to meet the Reference Set threshold (i.e., $N \leq 10$ instances), provided it contains at least one representative label.

\vspace{0.2em}
\noindent \textbf{Optimization}: 
We optimize the baseline TTN using $p_w=1$ in \cref{eq:loss}, a batch size of 256, and an SGD optimizer ($5 \times 10^{-4}$ weight decay, $5 \times 10^{-3}$ learning rate, implemented via PyTorch \cite{NEURIPS2019_bdbca288}).
Each distinct dataset contributes 1,000 training sequences per epoch, with a 20\% held-out validation set for model selection.
To ensure the TTN is a balanced and domain-agnostic judge, we select the final parameters $\vw$ from the checkpoint achieving the highest validation accuracy.
Specifically, we evenly weight the Accept ($\sS = 1$) and Reject ($\sS \neq 1$) accuracies per dataset, then average across all datasets in \cref{tab:icdataset}.
This dual-averaging strategy prevents the TTN from over-fitting to specific image domains or majority-class distributions.
Notably, rotating disjoint categories $c_i \neq c^*$ evenly within each dataset also reduces memorizing specific features for the 1,303 overall classes.
The TTN trains for 100\textrm{K} epochs (hold-out selection at the 90\textrm{K} checkpoint) in 61.05 hours on a single NVIDIA L40S GPU.
This modest computational requirement---relative to the foundation models it evaluates---highlights TTN efficiency as a general lightweight pruning module.

\subsection{Zero-Shot Cross-Task Transfer}
\label{sec:zs}

\setlength{\tabcolsep}{5.1pt}
\begin{table} [t!]
	\centering
	\caption{{\bf TTN Classification LIG Evaluation}. TTN trains on 80\% of images from all categories and is evaluated on the remaining 20\% for validation.
		The Accept/Reject validation accuracies are for Ideal ($\sS= \ $\ref{s:1}) and OCG ($\sS= \ $\ref{s:2}) Candidates, respectively.}
	\vspace{-0.75em}
	\scriptsize
	\begin{tabular}{| c | r | r | c | r | r | c | }
		\hline 
		\rowcolor{tableheader} \rule{0pt}{3ex} & \multicolumn{2}{ c |}{\bf \footnotesize Number of}   &  \multicolumn{1}{ c |}{\bf \footnotesize Image} & \multicolumn{3}{c |}{\bf \footnotesize TTN 80/20 Split Acc. } \\ 
		\rowcolor{tableheader} \rule{0pt}{2.5ex} \footnotesize \bf Dataset & \multicolumn{1}{ c }{\bf \footnotesize Classes} & \multicolumn{1}{ c |}{\bf \footnotesize Images} & \multicolumn{1}{ c |}{\bf \footnotesize Size} & \multicolumn{1}{ c }{\bf \footnotesize Accept} & \multicolumn{1}{ c }{\bf \footnotesize Reject} & \multicolumn{1}{ c |}{\bf \footnotesize Overall} \\ \hline
		ImageNet \cite{imagenet}	&	1,000	&	1,281,167	&	Varied	&	97.7\% &	94.8\% &	96.2\% \\ \hline
		Food-101 \cite{bossard14} &	101	&	101,000	&	$\leq$512	&	97.7\% &	95.3\% &	96.5\% \\ \hline
		CIFAR100 \cite{cifar}	&	100	&	50,000	&	32$\times$32	&	77.6\% &	85.4\% &	81.5\% \\ \hline
		MIT Indoor Scenes \cite{indoorscenes}	&	67	&	15,620	&	$\geq$200	&	98.9\% &	96.5\% &	97.7\% \\ \hline
		Fashion-MNIST \cite{fashionmnist} &	10	&	60,000	&	28$\times$28	&	97.2\% &	92.5\% &	94.8\% \\ \hline
		CIFAR10 \cite{cifar}	&	10	&	50,000	&	32$\times$32	&	98.3\% &	97.8\% &	98.0\% \\ \hline
		EuroSAT	\cite{eurosat} &	10	&	27,000	&	64$\times$64	&	98.6\% &	97.6\% &	98.1\% \\ \hline
		VTID2 \cite{vtid2}	&	5	&	4,356	&	Varied	&	100.0\% &	99.8\% &	99.9\% \\ \hline
	\end{tabular}
	\label{tab:icdataset}
\end{table}

TTN's zero-shot cross-task transfer performance---pruning object detection labels with a model trained purely on image classification (\cref{fig:front,fig:qualconf,fig:qualbox,fig:schematic})---stems directly from our training protocol, which forces the TTN to identify robust semantic boundaries purely from diverse, high-fidelity input sequences (\cref{tab:icdataset}).
Specifically, the TTN training sampling distribution $P(\sS)$ uses only human-verified sources, $\pi_\text{\ref{s:1}}=\pi_\text{\ref{s:2}}=\frac{1}{2}$, with zero pseudo-label probabilities, $\pi_\text{\ref{s:3}}=\pi_\text{\ref{s:4}}=0$.
By excluding noisy pseudo-labels and detection-specific formats during this phase, we ensure the TTN internalizes a robust intrinsic semantic logic. 
This foundation generalizes to novel tasks and categories in a zero-shot manner, requiring no task-specific supervision.
Furthermore, including low-resolution datasets (e.g., CIFAR, Fashion-MNIST) acts as an inherent curriculum for scale variation.

\vspace{0.2em}
\noindent \textbf{TTN LIG Evaluation}:
As detailed in \cref{tab:icdataset}, the TTN achieves high validation accuracy ($>94\%$ Overall on 7/8 datasets) across diverse classification domains. 
Notably, the model maintains robust performance even on low-resolution data (e.g., 98.0\% on CIFAR10), which is representative of the visual degradations typical of tiny object pseudo-labels. 
By learning to judge pixelated and high-fidelity labels for numerous classes, the TTN develops a domain-agnostic semantic logic that facilitates effective zero-shot transfer from classification to detection.

\vspace{0.2em}
\noindent \textbf{Self-Referential Pruning}:
After foundation model pseudo-labeling, the TTN prunes detection labels \textit{without} human supervision by constructing a dataset-spanning ``Ideal'' set from the 1,000 highest-confidence pseudo-labels per target category $c^*$, $|\sG^{(c^*)}| \leq 1,000$. 
We then perform a Self-Referential LIG for each category, where the TTN evaluates \textit{every} pseudo-label candidate $\hat{\vg}_i^{(c^*)} \in \hat{\sG}^{(c^*)}$ (including ``Ideal'' selections) against 10 reference samples drawn from $\sG^{(c^*)}$.
We aggregate the Accept/Reject logits for each candidate ($z_i$) across multiple games to cover all available references and increase robustness; if a candidate's aggregate score is $>0$, it is pruned (\cref{eq:h}).
This iterative process allows the TTN to identify and reject systemic hallucinations, even if they had high initial confidence (\cref{fig:qualconf}).
Categories lacking sufficient reference labels ($|\sG^{(c^*)}| < 10$) are preserved with their original pseudo-labels to prevent uncontextualized pruning.

\subsection{Task-Specific Fine-Tuning}
\label{sec:ft}

\setlength{\tabcolsep}{3.8pt}
\begin{table}[t!]
	\centering
	\caption{{\bf TTN$_\text{\tiny D}$ Detection LIG Evaluation}. 
		We train TTN$_\text{\tiny D}$ with VLM pseudo-labels and $\leq$ 100 human labels per class. 
		Accept/Reject validation accuracies are for $\sS=~$\ref{s:1}/$\sS=\{2,3,4\}$ Candidates, respectively. 
		TTN$_\text{\tiny D}$ uses a lower $p_w$ (\cref{eq:loss}) to prioritize Accept accuracy; Reject accuracy varies with instance density and class complexity.
		}
	\vspace{-0.75em}
		\scriptsize
	\begin{tabular}{| c | r | r | r | c | r | r | c | }
		\hline 
		\rowcolor{tableheader} \rule{0pt}{3ex} & \multicolumn{3}{ c |}{\bf \footnotesize Number of Training}   & \bf \footnotesize Objects & \multicolumn{3}{c |}{\bf \footnotesize TTN$_\text{\tiny D}$ 80/20 Split Acc. } \\ 
		\rowcolor{tableheader} \rule{0pt}{2.5ex} \bf \footnotesize Dataset & \multicolumn{1}{ c }{\bf \footnotesize Classes} & \multicolumn{1}{ c }{\bf \footnotesize Images} & \multicolumn{1}{ c |}{\bf \footnotesize Objects} & \bf \footnotesize per Image & \multicolumn{1}{ c }{\bf \footnotesize Accept} & \multicolumn{1}{ c }{\bf \footnotesize Reject} & \multicolumn{1}{ c |}{\bf \footnotesize Overall} \\ \hline
		VOC	\cite{EvEtAl10} &	20	&	16,551 & 40,058	& ~~2.42 & 95.1\%	&	84.2\%	&	89.7\%	\\ \hline
	COCO \cite{coco}	&	80	&	118,287 & 849,945	& ~~7.19 &	96.8\%	&	59.2\%	&	78.0\%	\\ \hline
	LVIS \cite{Gupta_2019_CVPR}	&	1,203	& 100,170	&	1,270,141 & 12.68 & 95.8\%	&	50.6\%	&	73.2\%	\\ \hline
	BDD \cite{Yu_2020_CVPR}	&	10	& 70,000 & 1,286,871	& 18.38 &	93.0\%	&	63.0\%	&	78.0\%	\\ \hline
	\end{tabular}
	\label{tab:detectdataset}
\end{table}

To maximize detection pruning performance, we test task-specific fine-tuning (TTN$_\text{\tiny D}$) with the full LIG training sampling distribution $P(\sS)$.
From TTN weights, we fine-tune a single TTN$_\text{\tiny D}$ model on all detection datasets in \cref{tab:detectdataset} simultaneously, spanning web images to autonomous driving.
This multi-domain approach ensures $\text{TTN}_{\text{\tiny D}}$ is a robust generalist across diverse visual contexts while mastering the specific spatial and semantic nuances of the object detection task.

For the Ideal set ($\sS= \ $\ref{s:1}), TTN$_\text{\tiny D}$ uses up to 100 human-verified detection labels for each target category (subject to availability), which are split 80/20 for the train/validation protocol.
For label efficiency, the 80 training labels are reused as the Ideal set for pseudo-label pruning at inference ($|\sG^{(c^*)}| \leq 80$).

To generate the remaining training distributions ($\sS=\{2,3,4\}$), TTN$_\text{\tiny D}$ uses three Vision-Language Models (VLMs): YOLO-World (YOLOW) \cite{Cheng_2024_CVPR}, YOLOE \cite{wang2025yoloerealtimeseeing}, and Grounding DINO-Tiny (GDINO)  \cite{shilong2024}, each sampled at a confidence threshold of $\tau=0.1$.
To maximize the hallucination signal relative to the Ideal set, we limit training examples from each VLM to the lowest-confidence quartile of pseudo-labels per class.
Each VLM-dataset combination contributes 1,000 training sequences per epoch, exposing TTN$_\text{\tiny D}$ to the diverse error characteristics and spatial inaccuracies of multiple VLM architectures.

TTN$_\text{\tiny D}$ adopts the baseline TTN training configuration with three modifications:
(i) a non-uniform sampling distribution of $\pi_\text{\ref{s:1}}=\frac{1}{2}$, $\pi_\text{\ref{s:2}}=\frac{1}{4}$, $\pi_\text{\ref{s:3}}=\pi_\text{\ref{s:4}}=\frac{1}{8}$;
(ii) a reduced pruning aggression ($p_w=0.2$ in \cref{eq:loss}) to instill a conservative bias that minimizes false rejections, ensuring the TTN$_\text{\tiny D}$ only prunes when there is strong evidence of a hallucination;
and (iii) a 0.02 learning rate for 4\textrm{K} epochs (with selection at the 2\textrm{K} checkpoint), requiring only 4.28 hours of fine-tuning.

After fine-tuning, $\text{TTN}_{\text{\tiny D}}$ achieves high validation accuracies (73--90\%, \cref{tab:detectdataset}).
Accept accuracy (Candidate $\sS=1$) consistently exceeds 93\% across all domains. 
Reject accuracy ($\sS=\{2,3,4\}$) naturally reflects the varying instance density and class complexity of the target datasets, which we explore further in \cref{sec:exp}.

\section{Experimental Evaluation of the LIG and TTN}
\label{sec:exp}

\subsection{Experimental Setup}
\label{sec:setup}

Our experimental setup is motivated by the comprehensive \textbf{Auto-Labeling} (AL) benchmark \cite{griff25AL}.
This foundation model pseudo-labeling benchmark uses GDINO, YOLOE, and YOLOW VLMs across various confidence thresholds $\tau$ and four diverse detection datasets: VOC, COCO, LVIS, and BDD.
We use $\tau \in [0.05, 0.5]$ to improve peak AL $F_1$-score performance via TTN pruning ($ \tilde{\sY} \subseteq \hat{\sY}$ in \cref{eq:f1}).
The AL benchmark includes training downstream detectors strictly on pseudo-labels, which provides another evaluation for TTN performance gains ($\tilde{\sD}$ in \cref{eq:downstream}).
In addition to TTN, we include \textbf{LIG$^*$} as an Oracle judge that prunes all candidate pseudo-labels with $\text{IoU}<0.5$.
This represents an IoU-specific upper bound for LIG-based pruning, though it preserves spatial label redundancies.
See \textbf{Supplementary Material for method comparisons, ablations, runtime, extended benchmark results,} and full experiment details (\cref{sec:ablate,sec:compare,sec:compute,sec:supexp,sec:supsetup}).

\subsection{Interrogating the Signal: Pseudo-Label Evaluation}
\label{sec:labeleval}

\begin{figure}[t!]
	\begin{minipage}{1\textwidth}
		\begin{tikzpicture}
	
	\definecolor{darkgrey176}{RGB}{176,176,176}
	\definecolor{gainsboro247}{RGB}{253,253,253}
	\definecolor{lightgrey204}{RGB}{204,204,204}
	
	\definecolor{lightgrey204}{RGB}{204,204,204}
	\definecolor{orangered2381020}{RGB}{238,102,0}
	\definecolor{slategrey119119153}{RGB}{119,119,153}
	
	\tikzstyle{every node}=[
	font=\scriptsize,
	]
	
	\begin{groupplot}[
		group style={
			group size=4 by 2,
			horizontal sep=0.65cm, 
			vertical sep=0.3cm,
		},
		legend cell align={left},
		legend style={
			draw opacity=1,
			text opacity=1,
			anchor=south west,
			draw=none,
			fill=gainsboro247,
			at={(2.56,-0.33)}, 
			legend columns=3,
			column sep = 1pt,
			fill opacity=0,
		},
		width=3.9cm,
		height=3.9cm,
		ytick align=inside,
		grid=both,
		x grid style={darkgrey176},
		y grid style={darkgrey176},
		title style={at={(0.5,1.125)},anchor=north},
		xtick style={draw=none},
		ytick style={draw=none},
		xminorticks=true,
		minor x tick num=1,
		yminorticks=true,
		minor y tick num=1,
		axis background/.style={fill=gainsboro247},
		yticklabel style = {xshift=1.5ex}, 
		xticklabel style = {yshift=1.2ex}, 
		y label style={at={(axis description cs:0.25, 0.5)}}, 
		xticklabels={},
		x label style={at={(axis description cs:-0.16, 0.1)}},
		yticklabel style={/pgf/number format/fixed zerofill,/pgf/number format/fixed, /pgf/number format/precision=1},
		]
\nextgroupplot[
tick align=outside,
tick pos=left,
title={\bf \footnotesize GDINO},
x grid style={darkgrey176},
xmin=2.75, xmax=52.25,
xtick style={color=black},
y grid style={darkgrey176},
ylabel={\bf \footnotesize Recall},
ymin=0.512217947411224, ymax=0.881915083219426,
ytick style={color=black},
ytick={0.1,0.2,0.3,0.4,0.5,0.6,0.7,0.8,0.9},
]
\addplot [line width=1.5pt, opacity=0.75,  black]
table {%
5 0.55328660546633
10 0.707028730491497
15 0.80760329199032
20 0.85151058333747
30 0.865110667955417
40 0.806965612104668
50 0.689022401151131
};
\addplot [line width=1.5pt, opacity=0.75,  slategrey119119153]
table {%
5 0.531879777576196
10 0.684792969516814
15 0.782177999721279
20 0.826223975735119
30 0.840787607696971
40 0.785485656574921
50 0.672375708335023
};
\addplot [line width=1.5pt, opacity=0.75,  orangered2381020]
table {%
5 0.529022362675233
10 0.681239307679837
15 0.778652986552651
20 0.820795888573896
30 0.835256679152282
40 0.781598596089186
50 0.672219824118314
};

\nextgroupplot[
tick align=outside,
tick pos=left,
title={\bf \footnotesize YOLOE},
x grid style={darkgrey176},
xmin=2.75, xmax=52.25,
xtick style={color=black},
y grid style={darkgrey176},
ymin=0.700792087407615, ymax=0.939183361394704,
ytick style={color=black},
ytick={0.1,0.2,0.3,0.4,0.5,0.6,0.7,0.8,0.9},
]
\addplot [line width=1.5pt, opacity=0.75,  black]
table {%
5 0.928347394395291
10 0.904289926661582
15 0.884080065030842
20 0.865158985980554
30 0.823582075425143
40 0.780803553326301
50 0.731891587347405
};
\addplot [line width=1.5pt, opacity=0.75,  slategrey119119153]
table {%
5 0.904867283064906
10 0.881620503407814
15 0.862434375537699
20 0.844359116287234
30 0.805450105677881
40 0.76489352751732
50 0.717629516743131
};
\addplot [line width=1.5pt, opacity=0.75,  orangered2381020]
table {%
5 0.890986668101373
10 0.869144778681164
15 0.850851072643878
20 0.832688351609518
30 0.795806591862989
40 0.756673081012875
50 0.711628054407029
};

\nextgroupplot[
tick align=outside,
tick pos=left,
title={\bf \footnotesize YOLOW},
x grid style={darkgrey176},
xmin=2.75, xmax=52.25,
xtick style={color=black},
y grid style={darkgrey176},
ymin=0.705522178570323, ymax=0.936896925920406,
ytick style={color=black},
ytick={0.1,0.2,0.3,0.4,0.5,0.6,0.7,0.8,0.9},
]
\addplot [line width=1.5pt, opacity=0.75,  black]
table {%
5 0.926379891949948
10 0.903876686015826
15 0.884516569898605
20 0.864491230447582
30 0.821034164237983
40 0.780356410505001
50 0.733901820633351
};
\addplot [line width=1.5pt, opacity=0.75,  slategrey119119153]
table {%
5 0.902620949710915
10 0.880833390069032
15 0.862740615341917
20 0.843587664895409
30 0.802564289927393
40 0.764283840936251
50 0.720344467461577
};
\addplot [line width=1.5pt, opacity=0.75,  orangered2381020]
table {%
5 0.894685784963332
10 0.873926247097644
15 0.856016689747858
20 0.83715100922862
30 0.796718530212335
40 0.759496290212589
50 0.716039212540782
};

\nextgroupplot[
tick align=outside,
tick pos=left,
title={\bf \footnotesize VOC Overall},
x grid style={darkgrey176},
xmin=2.75, xmax=52.25,
xtick style={color=black},
y grid style={darkgrey176},
ymin=0.691941135210384, ymax=0.868408161733527,
ytick style={color=black},
ytick={0.1,0.2,0.3,0.4,0.5,0.6,0.7,0.8,0.9},
]
\addplot [line width=1.5pt, opacity=0.75, black]
table {%
	5 0.802671297270523
	10 0.838398447722968
	15 0.858733308973256
	20 0.860386933255202
	30 0.836575635872848
	40 0.789375191978657
	50 0.718271936377296
};
\addplot [line width=1.5pt, opacity=0.75, slategrey119119153]
table {%
	5 0.779789336784006
	10 0.81574895433122
	15 0.835784330200299
	20 0.838056918972587
	30 0.816267334434082
	40 0.771554341676164
	50 0.703449897513244
};
\addplot [line width=1.5pt, opacity=0.75, orangered2381020]
table {%
	5 0.771564938579979
	10 0.808103444486215
	15 0.828506916314796
	20 0.830211749804011
	30 0.809260600409202
	40 0.76592265577155
	50 0.699962363688708
};

\nextgroupplot[
ylabel={\bf \footnotesize Precision},
tick align=outside,
tick pos=left,
x grid style={darkgrey176},
xmin=2.75, xmax=52.25,
xtick style={color=black},
y grid style={darkgrey176},
ymin=0.00954063875974447, ymax=0.893714477047121,
ytick style={color=black},
xtick={10,30,50},
xticklabels={0.1,0.3,0.5},
ytick={0.0,0.1,0.2,0.3,0.4,0.5,0.6,0.7,0.8,0.9},
yticklabels={\tiny 0.0,\tiny 0.1,\tiny 0.2,\tiny 0.3,\tiny 0.4,\tiny 0.5,\tiny 0.6,\tiny 0.7,\tiny 0.8,\tiny 0.9},
]
\addplot [line width=1.5pt, opacity=0.75,  black]
table {%
5 0.0497303586818979
10 0.128808823824451
15 0.231598763723088
20 0.34222428425746
30 0.538218926773006
40 0.699974280404503
50 0.799158515554974
};
\addlegendentry{\bf  \scriptsize AL \cite{griff25AL}}
\addplot [line width=1.5pt, opacity=0.75,  slategrey119119153]
table {%
5 0.0857136046180775
10 0.216460054018451
15 0.338619045382095
20 0.452116089900558
30 0.632192010562879
40 0.760042958343155
50 0.83409024458204
};
\addlegendentry{\bf  \scriptsize TTN}
\addplot [line width=1.5pt, opacity=0.75,  orangered2381020]
table {%
5 0.141639386285382
10 0.306800506898294
15 0.441075787532423
20 0.545859089752579
30 0.696080827750605
40 0.796443665343644
50 0.853524757124967
};
\addlegendentry{\bf  \scriptsize TTN$_{\text{\tiny D}}$}

\nextgroupplot[
tick align=outside,
tick pos=left,
x grid style={darkgrey176},
xmin=2.75, xmax=52.25,
xtick style={color=black},
y grid style={darkgrey176},
ymin=0.378837575149429, ymax=0.874533812197455,
ytick style={color=black},
xtick={10,30,50},
xticklabels={0.1,0.3,0.5},
xlabel={\bf Confidence Threshold},
ytick={0.1,0.2,0.3,0.4,0.5,0.6,0.7,0.8,0.9},
]
\addplot [line width=1.5pt, opacity=0.75,  black]
table {%
5 0.401369222287976
10 0.506391653930586
15 0.573170093634231
20 0.622351273402392
30 0.693574111979249
40 0.748150418055929
50 0.794475249295299
};
\addplot [line width=1.5pt, opacity=0.75,  slategrey119119153]
table {%
5 0.498984727855851
10 0.590004762934002
15 0.647160389074377
20 0.689153754221398
30 0.748536389619551
40 0.793703301729235
50 0.8313590028968
};
\addplot [line width=1.5pt, opacity=0.75,  orangered2381020]
table {%
5 0.575802270629956
10 0.65378412622781
15 0.700659979402693
20 0.734565921720883
30 0.784362573991732
40 0.82149753694761
50 0.852002165058908
};

\nextgroupplot[
tick align=outside,
tick pos=left,
x grid style={darkgrey176},
xmin=2.75, xmax=52.25,
xtick style={color=black},
y grid style={darkgrey176},
ymin=0.45498169501851, ymax=0.880760216115269,
ytick style={color=black},
xtick={10,30,50},
xticklabels={0.1,0.3,0.5},
ytick={0.1,0.2,0.3,0.4,0.5,0.6,0.7,0.8,0.9},
]
\addplot [line width=1.5pt, opacity=0.75,  black]
table {%
5 0.474335264159272
10 0.569945000432824
15 0.628192854775805
20 0.669223219945586
30 0.730753048182195
40 0.780622469277059
50 0.820632766821215
};
\addplot [line width=1.5pt, opacity=0.75,  slategrey119119153]
table {%
5 0.564962075616121
10 0.645573718383748
15 0.69277198806695
20 0.725694915287137
30 0.775963383854067
40 0.816390485194129
50 0.848526641234064
};
\addplot [line width=1.5pt, opacity=0.75,  orangered2381020]
table {%
5 0.625968295528453
10 0.693434894574288
15 0.732913338607442
20 0.759387402973443
30 0.80089871551856
40 0.835334456126986
50 0.861406646974507
};

\nextgroupplot[
tick align=outside,
tick pos=left,
x grid style={darkgrey176},
xmin=2.75, xmax=52.25,
xtick style={color=black},
y grid style={darkgrey176},
ymin=0.281119969642561, ymax=0.883002835119948,
ytick style={color=black},
xtick={10,30,50},
xticklabels={0.1,0.3,0.5},
ytick={0.1,0.2,0.3,0.4,0.5,0.6,0.7,0.8,0.9},
]
\addplot [line width=1.5pt, opacity=0.75, black]
table {%
	5 0.308478281709715
	10 0.401715159395953
	15 0.477653904044375
	20 0.544599592535146
	30 0.65418202897815
	40 0.742915722579163
	50 0.804755510557163
};
\addplot [line width=1.5pt, opacity=0.75, slategrey119119153]
table {%
	5 0.383220136030017
	10 0.484012845112067
	15 0.559517140841141
	20 0.622321586469698
	30 0.718897261345499
	40 0.790045581755507
	50 0.837991962904301
};
\addplot [line width=1.5pt, opacity=0.75, orangered2381020]
table {%
	5 0.447803317481264
	10 0.551339842566798
	15 0.624883035180853
	20 0.679937471482302
	30 0.760447372420299
	40 0.81775855280608
	50 0.855644523052794
};
\end{groupplot}

\end{tikzpicture}
	\end{minipage}%
	\vspace{-2.1em}
	\caption{{\bf Recall and Precision across VLM Models and Confidence Thresholds on VOC}. 
		YOLOW, YOLOE have higher recall and precision at low thresholds values, likely due to better internal non-maximum suppression of predicted labels.
		TTN prunes away predicted label failures, which incidentally decreases recall but more substantially increases precision.
		Additional metric plots across all datasets provided in \cref{fig:metric}.
	}
	\label{fig:vocmetric}
\end{figure}
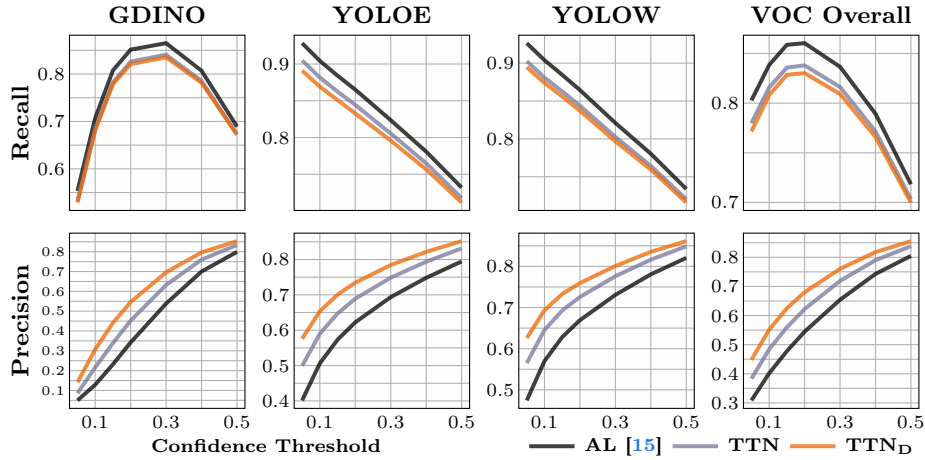

We first evaluate the effectiveness of the TTN as a standalone pseudo-label filter. 
As shown on VOC in \cref{fig:vocmetric}, TTN consistently improves precision across all VLM models and confidence thresholds. 
While pruning naturally results in an incidental decrease in recall, the more substantial precision increases lead to overall $F_1$-score improvements across all datasets (\cref{tab:metricmap}). 
Specifically, TTN and TTN$_\text{\tiny D}$ improve Overall $F_1$ by 4\% and 7\%, respectively, advancing toward the 45\% gain set by the LIG$^*$ Oracle.
Notably, LIG$^*$ exhibits an epsilon improvement in recall due to matching disambiguation.
Essentially, by rejecting high-entropy labels, we reduce Hungarian matching interference and enable higher-fidelity pairings between the accepted pseudo-labels and ground truth during evaluation.

We also evaluate the $F_2$ score, which favors recall \big({\footnotesize$\frac{\text{5TP}}{\text{5TP+4FN+FP}}$}$\in[0,1]$\big), to measure the trade-off between pruning and signal preservation. 
TTN$_\text{\tiny D}$ improves Overall $F_2$ by 3\%, demonstrating that its task-specific logic ``detoxifies'' the label set while preserving high-recall candidates for downstream training.

\setlength{\tabcolsep}{4.3pt}
\begin{table} [t!]
	\centering
	\caption{{\bf Combined Label and Downstream Model Evaluation}. Results are averaged over VLM model, confidence threshold, and class, except for Labels, which is the total number for all classes. Overall is additionally averaged over all datasets.}
	\vspace{-0.75em}
	\scriptsize
	\begin{tabular}{| c| l | r |c | c | c | c| c| c| }
		\hline 
		\rowcolor{tableheader}   & \rule{0pt}{3ex}  &  \multicolumn{7}{ c |}{\bf \footnotesize Mean of VLM, Confidence Threshold, \& Class}  \\ 
		\rowcolor{tableheader}  \rule{0pt}{2.5ex} \bf \footnotesize Dataset & \multicolumn{1}{ c |}{\bf \footnotesize Method} & \multicolumn{1}{ c }{\bf \footnotesize Labels} & \multicolumn{1}{ c }{\bf \footnotesize Recall} & \multicolumn{1}{ c }{\bf \footnotesize Prec.} & \multicolumn{1}{ c }{\bf \footnotesize $F_1$} & \multicolumn{1}{ c }{\bf \footnotesize $F_2$} & \multicolumn{1}{ c }{\bf \scriptsize mAP$_{50}$} & \multicolumn{1}{ c |}{\bf \scriptsize mAP$_{50:95}$}  \\ \hline
		\noalign{\vskip1.0pt} \hline
		\rowcolor{ligtable} \cellcolor{gray!0}	&	 LIG$^*$	&	39,008	&	0.816	&	0.914	&	0.845	&	0.825	&	0.707	& 0.500		\\ \cline{2-9}
		&	TTN$_{\text{\tiny D}}$	&	61,337	&	0.788	&	\bf 0.677	&	\bf 0.695	&\bf 	0.734	&	\bf 0.685	&	\bf 0.486		\\ \cline{2-9}
		\bf \footnotesize VOC	&	TTN	&	78,403	&	0.794	&	0.628	&	0.662	&	0.716	&	0.675	&	0.478	\\ \cline{2-9}
		&	AL \cite{griff25AL}	&	134,899	&	\bf 0.815	&	0.562	&	0.615	&	0.690	&	0.656	&	0.466	\\ \hline
		\noalign{\vskip1.0pt} \hline
		\rowcolor{ligtable} \cellcolor{gray!0}		&	LIG$^*$	&	627,792	&	0.594	&	0.908	&	0.689	&	0.626	&	0.454	& 0.315	\\ \cline{2-9}
		&	TTN$_{\text{\tiny D}}$	&	1,565,002	&	0.582	&	0.537	&	0.511	&	\bf 0.534	& \bf	0.436	&	0.302		\\ \cline{2-9}
		\bf \footnotesize COCO	&	TTN	&	1,192,383	&	0.538	&	\bf 0.571	&	\bf 0.513	&	0.515	&	0.434	&\bf	0.303	\\ \cline{2-9}
		
		&	AL	&	1,834,962	&	\bf 0.593	&	0.514	&	0.499	&	0.531	&	0.432	&	0.299\\ \hline
		\noalign{\vskip1.0pt} \hline
		\rowcolor{ligtable} \cellcolor{gray!0}		&	LIG$^*$	&	290,613	&	0.137	&	0.600	&	0.191	&	0.154	&	0.059	& 0.041	\\ \cline{2-9}
		&	TTN$_{\text{\tiny D}}$	&	1,659,673	&	0.136	&	\bf 0.116	&	\bf 0.079	&	\bf 0.087	&	\bf 0.060	&	\bf 0.042		\\ \cline{2-9}
		\bf \footnotesize LVIS	&	TTN	&	1,508,817	&	0.131	&\bf 	0.116	&\bf 	0.079	&	0.086	&	0.059	&	\bf 0.042	\\ \cline{2-9}
		
		&	AL	&	1,777,051	&	\bf 0.137	&	0.112	&	0.078	&	\bf 0.087	&	0.059	&	\bf 0.042	\\ \hline
		\noalign{\vskip1.0pt} \hline
		\rowcolor{ligtable} \cellcolor{gray!0}	&	LIG$^*$	&	426,726	&	0.267	&	0.837	&	0.376	&	0.301	&	0.306 &	0.178		\\ \cline{2-9}
		&	TTN$_{\text{\tiny D}}$	&	773,755	&	0.243	&	\bf 0.459	&	\bf 0.262	&	0.235	&	\bf 	0.269	&	\bf 	0.156	\\ \cline{2-9}
		\bf \footnotesize BDD	&	TTN	&	854,393	&	0.242	&	0.428	&	0.249	&	0.227	&	0.252	&	0.148	\\ \cline{2-9}
		&	AL	&	1,084,056	&	\bf 0.267	&	0.411	&	0.256	&	\bf 0.241	&	0.253	&	0.147\\ \hline
		\noalign{\vskip1.0pt} \hline
		\rowcolor{ligtable} \cellcolor{gray!0}		&	LIG$^*$	&	346,035	&	0.454	&	0.815	&	0.525	&	0.476	&	0.382	& 0.259		\\ \cline{2-9}
		&	TTN$_{\text{\tiny D}}$	&	1,014,942	&	0.437	&	\bf 0.447	&	\bf 0.387	&	\bf 0.398	&		\bf 	0.362	&		\bf 0.247		\\ \cline{2-9}
		\bf \footnotesize Overall	&	TTN	&	908,499	&	0.426	&	0.436	&	0.376	&	0.386	&	0.355	&	0.243		\\ \cline{2-9}
		&	AL	&	1,207,742	&	\bf 0.453	&	0.400	&	0.362	&	0.387	&		0.350	&	0.238	\\ \hline
	\end{tabular}
	\label{tab:metricmap}
\end{table}

\subsection{Detector Training and Evaluation on Detoxified Labels}

We further evaluate the utility of our pruning framework by training downstream object detectors on the detoxified pseudo-labeled datasets. 
Specifically, for each dataset, we train a YOLO11n detector \cite{yolo11_ultralytics} on unlabeled training images paired with TTN-pruned pseudo-labels ($\tilde{\sD}$ in \cref{eq:downstream}). 
After training, these detectors are then evaluated on their respective human-labeled validation sets. 
Critically, these validation sets (both images and labels) are entirely disjoint and unseen by both the TTN and the downstream detectors, ensuring a robust measure of how TTN pruning improves generalization from VLM supervision.

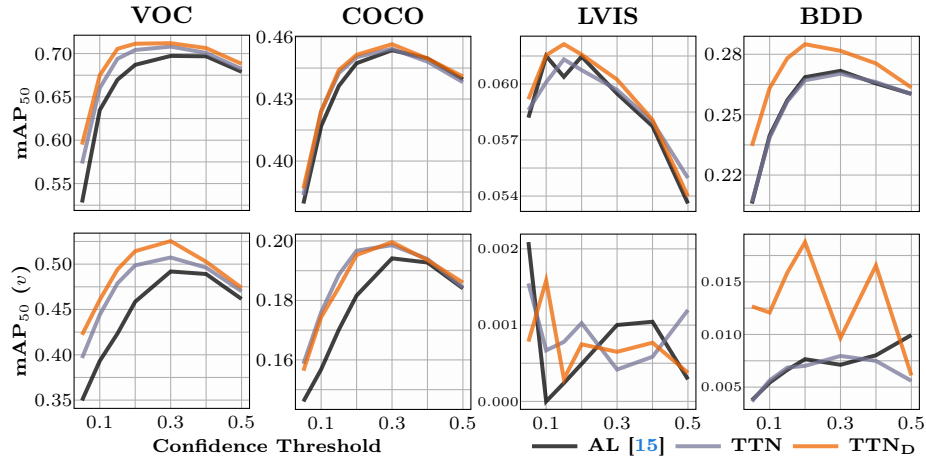
\begin{figure}[t!]
	\begin{minipage}{0.264\textwidth}
		\begin{tikzpicture}
	
	\definecolor{darkgrey176}{RGB}{176,176,176}
	\definecolor{gainsboro247}{RGB}{253,253,253}
	\definecolor{lightgrey204}{RGB}{204,204,204}
	
	\definecolor{lightgrey204}{RGB}{204,204,204}
	\definecolor{orangered2381020}{RGB}{238,102,0}
	\definecolor{slategrey119119153}{RGB}{119,119,153}
	
	\tikzstyle{every node}=[
	font=\scriptsize,
	]
	
	\begin{groupplot}[
		group style={
			group size=1 by 5,
			vertical sep=0.3cm,
		},
		legend cell align={left},
		legend style={
			fill opacity=0.4,
			draw opacity=1,
			text opacity=1,
			at={(0.16,0.04)},
			anchor=south west,
			draw=none,
			fill=gainsboro247,
			at={(2.55,-0.33)}, 
			legend columns=3,
			column sep = 1pt,
			fill opacity=0,
		},
		width=3.9cm,
		height=3.9cm,
		ytick align=inside,
		grid=both,
		x grid style={darkgrey176},
		y grid style={darkgrey176},
		title style={at={(0.5,1.125)},anchor=north},
		xtick style={draw=none},
		ytick style={draw=none},
		xminorticks=true,
		minor x tick num=1,
		yminorticks=true,
		minor y tick num=1,
		axis background/.style={fill=gainsboro247},
		yticklabel style = {xshift=1.8ex},
		xticklabel style = {yshift=1.2ex},
		y label style={at={(axis description cs:0.23, 0.5)}},
		xticklabels={},
		x label style={at={(axis description cs:1.1, 0.1)}}, 
		yticklabel style={/pgf/number format/fixed zerofill,/pgf/number format/fixed},
]
\nextgroupplot[
title={\bf \footnotesize VOC},
tick align=outside,
tick pos=left,
x grid style={darkgrey176},
xmin=0.0275, xmax=0.5225,
xtick style={color=black},
y grid style={darkgrey176},
ylabel={\bf mAP$_{50}$},
ymin=0.518957388654372, ymax=0.72116986951955,
ytick style={color=black}
]
\addplot [line width=1.5pt, opacity=0.75,  black]
table {%
0.05 0.528148865057335
0.1 0.634772700227172
0.15 0.669823921524255
0.2 0.686987918913502
0.3 0.697254868860603
0.4 0.696824556833834
0.5 0.678998157163896
};
\addplot [line width=1.5pt, opacity=0.75,  slategrey119119153]
table {%
0.05 0.57309728198465
0.1 0.660205352944207
0.15 0.693961448628902
0.2 0.704131400854395
0.3 0.708014587367649
0.4 0.700766066892746
0.5 0.68224583201107
};
\addplot [line width=1.5pt, opacity=0.75,  orangered2381020]
table {%
0.05 0.594961623185282
0.1 0.675004642193686
0.15 0.705519551771505
0.2 0.711318140059466
0.3 0.711978393116587
0.4 0.706298220728596
0.5 0.688167794500336
};

\nextgroupplot[
legend cell align={left},
tick align=outside,
tick pos=left,
x grid style={darkgrey176},
xmin=0.0275, xmax=0.5225,
xtick style={color=black},
y grid style={darkgrey176},
ylabel={\bf mAP$_{50}$ ($v$)},
ytick style={color=black},
xtick={0.1,0.3,0.5},
xticklabels={0.1,0.3,0.5},
xlabel={\bf Confidence Threshold},
ymin=0.340875001520156, ymax=0.53420074201452,
ytick style={color=black}
]
\addplot [line width=1.5pt, opacity=0.75,  black]
table {%
	0.05 0.349662535178991
	0.1 0.393196882129064
	0.15 0.423431418826823
	0.2 0.458567068640171
	0.3 0.491912262819532
	0.4 0.489096698792596
	0.5 0.461562050220932
};
\addlegendentry{\bf \scriptsize AL \cite{griff25AL}}
\addplot [line width=1.5pt, opacity=0.75,  slategrey119119153]
table {%
	0.05 0.396372808363959
	0.1 0.444136472911475
	0.15 0.478828411668619
	0.2 0.498640955736312
	0.3 0.50734837524283
	0.4 0.496173895784477
	0.5 0.470118604840141
};
\addlegendentry{\bf  \scriptsize TTN}
\addplot [line width=1.5pt, opacity=0.75,  orangered2381020]
table {%
	0.05 0.422014275918334
	0.1 0.460597328884738
	0.15 0.494106813141464
	0.2 0.514326529095923
	0.3 0.525413208355686
	0.4 0.502681929344607
	0.5 0.473919646183176
};
\addlegendentry{\bf  \scriptsize TTN$_{\text{\tiny D}}$}
\end{groupplot}

\end{tikzpicture}
	\end{minipage}%
	\begin{minipage}{0.24\textwidth}
		\begin{tikzpicture}
	
	\definecolor{darkgrey176}{RGB}{176,176,176}
	\definecolor{gainsboro247}{RGB}{253,253,253}
	\definecolor{lightgrey204}{RGB}{204,204,204}
	
	\definecolor{lightgrey204}{RGB}{204,204,204}
	\definecolor{orangered2381020}{RGB}{238,102,0}
	\definecolor{slategrey119119153}{RGB}{119,119,153}
	
	\tikzstyle{every node}=[
	font=\scriptsize,
	]
	
	\begin{groupplot}[
		group style={
			group size=1 by 5,
			vertical sep=0.3cm,
		},
		legend cell align={left},
		legend style={
			fill opacity=0.4,
			draw opacity=1,
			text opacity=1,
			at={(0.16,-0.01)},
			anchor=south west,
			draw=none,
			fill=gainsboro247
		},
		width=3.9cm,
		height=3.9cm,
		ytick align=inside,
		grid=both,
		x grid style={darkgrey176},
		y grid style={darkgrey176},
		title style={at={(0.5,1.125)},anchor=north},
		xtick style={draw=none},
		ytick style={draw=none},
		xminorticks=true,
		minor x tick num=1,
		yminorticks=true,
		minor y tick num=1,
		axis background/.style={fill=gainsboro247},
		yticklabel style = {xshift=1.8ex},
		xticklabel style = {yshift=1.2ex},
		y label style={at={(axis description cs:0.23, 0.5)}},
		xticklabels={},
		x label style={at={(axis description cs:1.18, 0.115)}},
		yticklabel style={/pgf/number format/fixed zerofill,/pgf/number format/fixed},
		]
\nextgroupplot[
tick align=outside,
tick pos=left,
title={\bf \footnotesize COCO},
x grid style={darkgrey176},
xmin=0.0275, xmax=0.5225,
xtick style={color=black},
y grid style={darkgrey176},
ymin=0.375569127189007, ymax=0.460381629376387,
ytick={0.37,0.40,0.43,0.46},
ytick style={color=black},
]
\addplot [line width=1.5pt, opacity=0.75,  black]
table {%
0.05 0.379424240924797
0.1 0.416872116444932
0.15 0.436347276625309
0.2 0.447333470410905
0.3 0.45362155714261
0.4 0.449659085928796
0.5 0.439557895498401
};
\addplot [line width=1.5pt, opacity=0.75,  slategrey119119153]
table {%
0.05 0.383439716228031
0.1 0.423979738750087
0.15 0.442558812681382
0.2 0.450267296995898
0.3 0.454287814507252
0.4 0.448085433761121
0.5 0.438146047893272
};
\addplot [line width=1.5pt, opacity=0.75,  orangered2381020]
table {%
	0.05 0.386686868595746
	0.1 0.424154564270664
	0.15 0.444001412487045
	0.2 0.451426032385978
	0.3 0.456526515640597
	0.4 0.449630838797904
	0.5 0.440907002404614
};

\nextgroupplot[
legend cell align={left},
legend style={
  fill opacity=0.8,
  draw opacity=1,
  text opacity=1,
  at={(0.97,0.03)},
  anchor=south east,
  draw=lightgrey204
},
tick align=outside,
tick pos=left,
x grid style={darkgrey176},
xmin=0.0275, xmax=0.5225,
xtick style={color=black},
y grid style={darkgrey176},
ytick style={color=black},
xtick={0.1,0.3,0.5},
xticklabels={0.1,0.3,0.5},
xlabel={\textcolor{white}{|}},
ymin=0.143342206907118, ymax=0.202279985780532,
ytick style={color=black}
]
\addplot [line width=1.5pt, opacity=0.75,  black]
table {%
	0.05 0.14602119685591
	0.1 0.156813698006951
	0.15 0.170175052293
	0.2 0.181634211261356
	0.3 0.194152000033028
	0.4 0.192751321522289
	0.5 0.184053323565268
};
\addplot [line width=1.5pt, opacity=0.75,  slategrey119119153]
table {%
	0.05 0.158703233530696
	0.1 0.176009481993604
	0.15 0.188695542714761
	0.2 0.196726747547
	0.3 0.19859011227733
	0.4 0.19391577907616
	0.5 0.184352249977896
};
\addplot [line width=1.5pt, opacity=0.75,  orangered2381020]
table {%
	0.05 0.156356354119693
	0.1 0.174358686110912
	0.15 0.184358865200059
	0.2 0.195275888951922
	0.3 0.199600995831741
	0.4 0.19356148576311
	0.5 0.186021059047108
};
\end{groupplot}

\end{tikzpicture}
	\end{minipage}%
	\begin{minipage}{0.242\textwidth}
		\begin{tikzpicture}
	
	\definecolor{darkgrey176}{RGB}{176,176,176}
	\definecolor{gainsboro247}{RGB}{253,253,253}
	\definecolor{lightgrey204}{RGB}{204,204,204}
	
	\definecolor{lightgrey204}{RGB}{204,204,204}
	\definecolor{orangered2381020}{RGB}{238,102,0}
	\definecolor{slategrey119119153}{RGB}{119,119,153}
	
	\tikzstyle{every node}=[
	font=\scriptsize,
	]
	
	\begin{groupplot}[
		group style={
			group size=1 by 5,
			vertical sep=0.3cm,
		},
		legend cell align={left},
		legend style={
			fill opacity=0.4,
			draw opacity=1,
			text opacity=1,
			at={(0.32,-0.01)},
			anchor=south west,
			draw=none,
			fill=gainsboro247
		},
		width=3.9cm,
		height=3.9cm,
		ytick align=inside,
		grid=both,
		x grid style={darkgrey176},
		y grid style={darkgrey176},
		title style={at={(0.5,1.125)},anchor=north},
		xtick style={draw=none},
		ytick style={draw=none},
		xminorticks=true,
		minor x tick num=1,
		yminorticks=true,
		minor y tick num=1,
		axis background/.style={fill=gainsboro247},
		yticklabel style = {xshift=1.8ex},
		xticklabel style = {yshift=1.2ex},
		y label style={at={(axis description cs:0.18, 0.75)}},
		xticklabels={},
		scaled ticks=false,
		yticklabel style={/pgf/number format/fixed},
		x label style={at={(axis description cs:0.85, 0.081)}},
		]
\nextgroupplot[
tick align=outside,
tick pos=left,
title={\bf \footnotesize LVIS},
x grid style={darkgrey176},
xmin=0.0275, xmax=0.5225,
xtick style={color=black},
y grid style={darkgrey176},
ymin=0.0531760455901073, ymax=0.0625623120430961,
ytick style={color=black},
xlabel={\textcolor{white}{A}},
ytick={0.054,0.057,0.06,0.063},
yticklabels={\tiny0.054,\tiny0.057,\tiny0.060,\tiny0.063},
]
\addplot [line width=1.5pt, opacity=0.75,  black]
table {%
0.05 0.0581980195108916
0.1 0.0615002605042572
0.15 0.0603838800322796
0.2 0.0614704393528882
0.3 0.0594868629963133
0.4 0.0577425291597004
0.5 0.0536026940652432
};
\addplot [line width=1.5pt, opacity=0.75,  slategrey119119153]
table {%
	0.05 0.0586078303214523
	0.1 0.0600971084689279
	0.15 0.061318820098087
	0.2 0.0607445761698374
	0.3 0.0596875261577388
	0.4 0.0579759582962494
	0.5 0.0549661829765793
};
\addplot [line width=1.5pt, opacity=0.75, orangered2381020]
table {%
	0.05 0.0591870569786722
	0.1 0.061544951070851
	0.15 0.0621356635679603
	0.2 0.0615784290255676
	0.3 0.0602405751587019
	0.4 0.0580814241370981
	0.5 0.053985727167371
};

\nextgroupplot[
legend cell align={left},
legend style={fill opacity=0.8, draw opacity=1, text opacity=1, draw=lightgrey204},
tick align=outside,
tick pos=left,
x grid style={darkgrey176},
xmin=0.0275, xmax=0.5225,
xtick style={color=black},
y grid style={darkgrey176},
xtick={0.1,0.3,0.5},
xticklabels={0.1,0.3,0.5},
ymin=-0.000104453261975445, ymax=0.00219351850148435,
ytick style={color=black},
xlabel={\textcolor{white}{I}},
ytick={0.000,0.001,0.002},
yticklabels={\tiny 0.000,\tiny0.001,\tiny0.002},
]
\addplot [line width=1.5pt, opacity=0.75,  black]
table {%
	0.05 0.00208906523950891
	0.1 0
	0.15 0.000239874537972681
	0.2 0.000490976637922128
	0.3 0.00100168737659618
	0.4 0.00104423582273441
	0.5 0.000289379174867042
};
\addplot [line width=1.5pt, opacity=0.75,  slategrey119119153]
table {%
	0.05 0.00154722607050333
	0.1 0.000668946893219315
	0.15 0.000780011438139469
	0.2 0.00102688244863727
	0.3 0.000417510500916318
	0.4 0.000585164970549372
	0.5 0.00119853726671791
};
\addplot [line width=1.5pt, opacity=0.75, orangered2381020]
table {%
	0.05 0.000783953566496866
	0.1 0.00158497084183092
	0.15 0.000294200376277889
	0.2 0.000748175646591479
	0.3 0.000650187104207623
	0.4 0.000769072374042443
	0.5 0.00037500430635346
};
\end{groupplot}

\end{tikzpicture}
	\end{minipage}%
	\begin{minipage}{0.2425\textwidth}
		\begin{tikzpicture}
	
	\definecolor{darkgrey176}{RGB}{176,176,176}
	\definecolor{gainsboro247}{RGB}{253,253,253}
	\definecolor{lightgrey204}{RGB}{204,204,204}
	
	\definecolor{lightgrey204}{RGB}{204,204,204}
	\definecolor{orangered2381020}{RGB}{238,102,0}
	\definecolor{slategrey119119153}{RGB}{119,119,153}
	
	\tikzstyle{every node}=[
	font=\scriptsize,
	]
	
	\begin{groupplot}[
		group style={
			group size=1 by 5,
			vertical sep=0.3cm,
		},
		legend cell align={left},
		legend style={
			fill opacity=0.4,
			draw opacity=1,
			text opacity=1,
			at={(0.32,0.096)},
			anchor=south west,
			draw=none,
			fill=gainsboro247
		},
		width=3.9cm,
		height=3.9cm,
		ytick align=inside,
		grid=both,
		x grid style={darkgrey176},
		y grid style={darkgrey176},
		title style={at={(0.5,1.125)},anchor=north},
		xtick style={draw=none},
		ytick style={draw=none},
		xminorticks=true,
		minor x tick num=1,
		yminorticks=true,
		minor y tick num=1,
		axis background/.style={fill=gainsboro247},
		yticklabel style = {xshift=1.8ex},
		xticklabel style = {yshift=1.2ex},
		y label style={at={(axis description cs:0.18, 0.5)}},
		xticklabels={},
		scaled ticks=false,
		yticklabel style={/pgf/number format/fixed},
		x label style={at={(axis description cs:0.6, 0.115)}},
		]
\nextgroupplot[
tick align=outside,
tick pos=left,
title={\bf \footnotesize BDD},
x grid style={darkgrey176},
xmin=0.0275, xmax=0.5225,
xtick style={color=black},
y grid style={darkgrey176},
ymin=0.201888997650319, ymax=0.28895967706361,
ytick={0.19,0.22,0.25,0.28},
ytick style={color=black}
]
\addplot [line width=1.5pt, opacity=0.75,  black]
table {%
0.05 0.205846755805468
0.1 0.239639458772534
0.15 0.257244091121361
0.2 0.26864038007851
0.3 0.271783876520301
0.4 0.265542044757935
0.5 0.260312090982754
};
\addplot [line width=1.5pt, opacity=0.75,  slategrey119119153]
table {%
0.05 0.206623808625266
0.1 0.238596224466994
0.15 0.256564183601741
0.2 0.267078896387552
0.3 0.270353072202736
0.4 0.266195105676502
0.5 0.260271784280872
};
\addplot [line width=1.5pt, opacity=0.75,  orangered2381020]
table {%
0.05 0.234508923259593
0.1 0.263154869098187
0.15 0.278050937001921
0.2 0.28500191890846
0.3 0.281730561720923
0.4 0.275464576213414
0.5 0.263477929949796
};

\nextgroupplot[
legend cell align={left},
legend style={
  fill opacity=0.8,
  draw opacity=1,
  text opacity=1,
  at={(0.97,0.03)},
  anchor=south east,
  draw=lightgrey204
},
tick align=outside,
tick pos=left,
x grid style={darkgrey176},
xmin=0.0275, xmax=0.5225,
xtick style={color=black},
y grid style={darkgrey176},
xtick={0.1,0.3,0.5},
xticklabels={0.1,0.3,0.5},
xlabel={\textcolor{white}{|}},
ymin=0.00287982447230311, ymax=0.0195259701215984,
ytick={0.0,0.005,0.01,0.015,0.02},
yticklabels={\tiny0.0,\tiny0.005,\tiny0.010,\tiny0.015},
ytick style={color=black}
]
\addplot [line width=1.5pt, opacity=0.75,  black]
table {%
	0.05 0.00375583606967493
	0.1 0.00542402388326594
	0.15 0.00663931588122533
	0.2 0.00765204067323682
	0.3 0.00712078073018423
	0.4 0.00802212892839601
	0.5 0.00994754457808801
};
\addplot [line width=1.5pt, opacity=0.75,  slategrey119119153]
table {%
	0.05 0.00363646745636199
	0.1 0.0056380227405784
	0.15 0.00685638237597705
	0.2 0.00702092967220141
	0.3 0.00795923745507675
	0.4 0.00749134837107178
	0.5 0.00559231100871265
};
\addplot [line width=1.5pt, opacity=0.75,  orangered2381020]
table {%
	0.05 0.0126681316844506
	0.1 0.0120828659761263
	0.15 0.0158788570481124
	0.2 0.0187693271375395
	0.3 0.00967150133596694
	0.4 0.0165551118734931
	0.5 0.00611329906983562
};
\end{groupplot}

\end{tikzpicture}
	\end{minipage}%
	\vspace{-2.1em}
	\caption{{\bf Downstream Model Evaluation across all Confidence Thresholds and Datasets}.
		Results are average over all VLM models (see \cref{fig:vocmetric} right).
		Row two results evaluate the vulnerable classes ($v$, \cref{sec:vulnerable}).
		mAP$_{50:95}$ plots are provided in \cref{sec:supexp}.
	}
	\label{fig:mapeff}
\end{figure}

As shown in \cref{tab:metricmap}, training on TTN-pruned datasets results in consistent mAP$_{50}$ and mAP$_{50:95}$ improvements across all VLM architectures and datasets. 
Remarkably, although TTN$_\text{\tiny D}$ does not reject nearly as many pseudo-labels as the LIG$^*$ Oracle (see Labels in \cref{tab:metricmap}), TTN$_\text{\tiny D}$ closes 38\% of the mAP$_{50}$ and 43\% of the mAP$_{50:95}$ gap between AL and LIG$^*$ performance. 
This indicates that TTN$_\text{\tiny D}$ successfully rejects many of the VLM hallucinations that are particularly disruptive for downstream training.
As illustrated in \cref{fig:mapeff}, mAP$_{50}$ gains are particularly pronounced at lower confidence thresholds ($\tau$). 
By detoxifying the low-$\tau$, high-recall regime, the TTN effectively expands the usable training distribution, allowing the downstream detector to learn from a larger, yet cleaner, set of examples. 
With TTN's generalization across categories established, we focus on the most challenging, transfer-vulnerable classes in \cref{sec:vulnerable}.

\subsection{Category Revival: Rescuing Vulnerable Classes}
\label{sec:vulnerable}

To understand TTN performance on challenge categories, we perform a targeted evaluation for AL's \cite{griff25AL} maximum and non-zero minimum $F_1$-score classes in \cref{tab:highlowclass}.
While TTN $F_1$ gains are consistent across VLM models and datasets for the Max. $F_1$ classes, gains for the challenge classes are particularly significant (e.g., $>300\%$ for GDINO Chair).
For the Min.~$F_1$ class mean, specifically, \textbf{zero-shot TTN increases $F_1$ by 28\% and TTN$_\text{\tiny D}$ increases $F_1$ by 44\%}, effectively ``raising the floor'' for VLM-driven labeling.
Remarkably, LIG$^*$ Oracle pruning increases Min.~$F_1$ by 432\%, indicating systemic VLM hallucinations that would have induced severe inductive interference in downstream models.

\setlength{\tabcolsep}{1.4 pt}
\begin{table} [t!]
	\centering
	\caption{{\bf $F_1$ Score on Max. \& Min. (Non-Zero) AL Classes at $\tau=0.1$}.
		Relative to AL \cite{griff25AL}, TTN and TTN$_\text{\tiny D}$ increase $F_1$ by 4\% and 7\% respectively for the best performing classes (top) and 28\% and 44\% for the worst performing classes (bottom).
	}
	\vspace{-0.75em}
	\scriptsize
	\begin{tabular}{| l| c| c| c |  c| c| c | c| c| c | c |  }
		\hline 
		\rowcolor{tableheader}  \rule{0pt}{3ex} & \multicolumn{3}{c|}{\bf \footnotesize VOC $F_1$} & \multicolumn{3}{c|}{\bf \footnotesize COCO $F_1$} & \multicolumn{3}{c|}{\bf \footnotesize BDD $F_1$} & \\
		\rowcolor{tableheader} \multicolumn{1}{|c|}{\scriptsize\bf Method} & \multicolumn{1}{c}{\bf \tiny GDINO} &  \multicolumn{1}{c}{\bf \tiny YOLOE} & \bf \tiny YOLOW &  \multicolumn{1}{c}{\bf \tiny GDINO} & \multicolumn{1}{c}{\bf \tiny YOLOE} & \bf \tiny YOLOW &  \multicolumn{1}{c}{\bf \tiny GDINO}& \multicolumn{1}{c}{\bf \tiny YOLOE}& \bf \tiny YOLOW & \bf\scriptsize Mean \\ \hline
		\noalign{\vskip1.0pt} \cline{2-10}
		\multicolumn{1}{c|}{\rule{0pt}{2.5ex} \scriptsize\bf Max. $F_1$ }	&	\scriptsize\bf Cat	&\scriptsize	\bf Cat	&\scriptsize	\bf Cat	&\scriptsize\bf 	Cat	&\scriptsize\bf 	Bear	&\scriptsize	\bf Giraffe	&\scriptsize	\bf Car	&\scriptsize\bf 	Car	&	\scriptsize\bf Person	&	\multicolumn{1}{c}{}	\\ \hline
		\rowcolor{ligtable} LIG$^*$	&	0.747	&	0.940	&	0.948	&	0.675	&	0.892	&	0.935	&	0.693	&	0.577	&	0.656	&	0.785 \\\hline
		TTN$_{\text{\tiny D}}$	&\bf 	0.687	&\bf 	0.907	&	\bf 0.915	&	\bf 0.602	&	0.819	&	0.873	&\bf 	0.478	&	0.519	&	\bf 0.545	&	\bf 0.705	\\\hline
		TTN	&	0.649	&	0.893	&	0.903	&	0.579	&	\bf 0.848	&	\bf 0.878	&	0.443	&	0.479	&	0.518	&	0.688	\\\hline
		AL \cite{griff25AL}        	&	0.585	&	0.822	&	0.851	&	0.528	&	0.784	&	0.861	&	0.438	&	\bf 0.535	&	0.543	&	0.661	\\ \hline
		\noalign{\vskip1.0pt} \cline{2-10}
		\multicolumn{1}{c|}{\rule{0pt}{2.5ex} \scriptsize\bf Min.	$F_1$}&\scriptsize	\bf Chair	&\scriptsize	\bf Chair	&\scriptsize	\bf Sofa	&	\bf \scriptsize Handbag	&\scriptsize\bf 	Toaster	&\scriptsize	\bf Toaster	&\bf \scriptsize	Train	&\bf \scriptsize	Motor	&\scriptsize	\bf Train	&		\multicolumn{1}{c}{}	\\ \hline
		\rowcolor{ligtable} LIG$^*$	&	0.641	&	0.892	&	0.940	&	0.516	&	0.840	&	0.686	&	0.323	&	0.018	&	0.319	&	0.575 \\\hline
		TTN$_{\text{\tiny D}}$	&	\bf 0.204	&	\bf 0.520	&	\bf 0.436	&	0.047	&	\bf 0.194	&\bf 	0.186	&	\bf 0.005	&	\bf 0.016	&\bf 	0.116	&\bf 	0.192	\\\hline
		TTN	&	0.172	&	0.489	&	0.391	&\bf 	0.051	&	0.180	&	0.175	&	0.002	&	0.014	&	0.058	&	0.170	\\\hline
		AL 	&	0.054	&	0.364	&	0.319	&	0.045	&	0.174	&	0.166	&	0.002	&	\bf 0.016	&	0.058	&	0.133	\\ \hline
	\end{tabular}
	\label{tab:highlowclass}
\end{table}

\setlength{\tabcolsep}{4.4pt}
\begin{table} [t!]
	\centering
	\caption{{\bf Vulnerable-Class ($v$) Performance Evaluation}. 
		We evaluate pseudo-label quality and downstream detector performance on the transfer-vulnerable classes (bottom quartile of AL \cite{griff25AL} mAP$_{50}$). 
		TTN and TTN$_\text{\tiny D}$ consistently outperform AL, mitigating the inductive interference caused by vision-language model hallucinations.
	}
	\vspace{-0.75em}
	\scriptsize
	\begin{tabular}{| c| l | r |c | c | c | c| c| c| }
		\hline 
		\rowcolor{tableheader}   & \rule{0pt}{3ex}  &  \multicolumn{7}{ c |}{\bf \footnotesize Mean of VLM, Confidence Threshold, \& Class}  \\ 
		\rowcolor{tableheader}  \rule{0pt}{2.5ex} \bf \footnotesize Dataset & \multicolumn{1}{ c |}{\bf \footnotesize Method} & \multicolumn{1}{ c }{\bf \footnotesize Labels} & \multicolumn{1}{ c }{\bf \footnotesize Recall} & \multicolumn{1}{ c }{\bf \footnotesize Prec.} & \multicolumn{1}{ c }{\bf \footnotesize $F_1$} & \multicolumn{1}{ c }{\bf \footnotesize $F_2$} & \multicolumn{1}{ c }{\bf \scriptsize mAP$_{50}$} & \multicolumn{1}{ c |}{\bf \scriptsize mAP$_{50:95}$}  \\ \hline
		\noalign{\vskip1.0pt} \hline
		\rowcolor{ligtable} \cellcolor{gray!0}	&	LIG$^*$	&	7,340	&	0.716	&	0.898	&	0.760	&	0.729	&	0.522 &	0.328		\\ \cline{2-9}
		&	TTN$_{\text{\tiny D}}$	&	19,265	&	0.682	&	\bf 0.513	&	\bf 0.512	&	\bf 0.576	&	\bf 0.485	&	\bf 0.308	\\  \cline{2-9}
		\bf \footnotesize VOC$_v$	&	TTN	&	22,084	&	0.686	&	0.483	&	0.490	&	0.562	&	0.470	&	0.300	\\ \cline{2-9}
		&	AL \cite{griff25AL}	&	47,139	&	\bf 0.715	&	0.425	&	0.442	&	0.528	&	0.438	&	0.281	\\ \hline
		\noalign{\vskip1.0pt} \hline
		\rowcolor{ligtable} \cellcolor{gray!0}	&	LIG$^*$	&	89,503	&	0.422	&	0.898	&	0.545	&	0.461	&	0.204 &	0.124		\\ \cline{2-9}
		&	TTN$_{\text{\tiny D}}$	&	446,132	&	0.413	&	0.392	&	0.338	&	\bf 0.355	&	0.184	&	0.112	\\  \cline{2-9}
		\bf \footnotesize COCO$_v$	&	TTN	&	319,843	&	0.380	&	\bf 0.410	&	\bf 0.340	&	0.345	&	\bf 0.185	&	\bf 0.113	\\ \cline{2-9}
		&	AL	&	544,203	&	\bf 0.421	&	0.381	&	0.333	&	0.353	&	0.175	&	0.106	\\ \hline
		\noalign{\vskip1.0pt} \hline
		\rowcolor{ligtable} \cellcolor{gray!0}	&	LIG$^*$	&	266	&	0.027	&	0.269	&	0.045	&	0.032	&	0.0004	& 0.0003	\\ \cline{2-9}
		&	TTN$_{\text{\tiny D}}$	&	27,025	&	\bf 0.027	&	\bf 0.034	&	\bf 0.013	&	\bf 0.014	&	0.0007	&	0.0006	\\  \cline{2-9}
		\bf \footnotesize LVIS$_v$	&	TTN	&	29,297	&	0.026	&	0.032	&	\bf 0.013	&	0.013	&	\bf 0.0009	&	\bf 0.0007	\\ \cline{2-9}
		&	AL	&	31,402	&	\bf 0.027	&	0.032	&	0.012	&	\bf 0.014	&	0.0007	&	0.0006	\\ \hline
		\noalign{\vskip1.0pt} \hline
		\rowcolor{ligtable} \cellcolor{gray!0}	&	LIG$^*$	&	453	&	0.077	&	0.576	&	0.123	&	0.091	&	0.036 &	0.016	\\ \cline{2-9}
		&	TTN$_{\text{\tiny D}}$	&	18,535	&	0.075	&	\bf 0.074	&	\bf 0.035	&	\bf 0.042	&	\bf 0.013	&	\bf 0.006	\\  \cline{2-9}
		\bf \footnotesize BDD$_v$	&	TTN	&	32,899	&	0.076	&	0.035	&	0.022	&	0.031	&	0.006	&	0.003	\\ \cline{2-9}
		&	AL	&	34,888	&	\bf 0.077	&	0.036	&	0.022	&	0.032	&	0.007	&	0.003	\\ \hline
		\noalign{\vskip1.0pt} \hline
		\rowcolor{ligtable} \cellcolor{gray!0}	&	LIG$^*$	&	24,390	&	0.311	&	0.660	&	0.368	&	0.328	&	0.191 &	0.117		\\ \cline{2-9}
		&	TTN$_{\text{\tiny D}}$	&	127,739	&	0.299	&	\bf 0.253	&\bf 	0.225	&\bf 	0.247	&\bf 	0.171	&\bf 	0.107	\\ \cline{2-9}
		\bf \footnotesize Overall$_v$	&	TTN	&	101,031	&	0.292	&	0.240	&	0.216	&	0.238	&	0.166	&	0.104	\\ \cline{2-9}
		&	AL	&	164,408	&	\bf 0.310	&	0.218	&	0.202	&	0.231	&	0.155	&	0.098	\\ \hline
	\end{tabular}
	\label{tab:metricmapv}
\end{table}

Motivated by this finding, we perform a broader analysis of these challenge categories in downstream model training.
Specifically, we evaluate transfer-\textbf{vulnerable classes} ($v$), defined per-VLM as the bottom quartile of classes ranked by AL's \cite{griff25AL} downstream mAP$_{50}$ at $\tau=0.1$.
As shown in \cref{fig:mapeff}, TTN pruning consistently improves mAP$_{50}$ across confidence thresholds and datasets on these failing categories. 
Remarkably, \textbf{TTN$_\text{\tiny D}$ approximately doubles precision, $F_1$, mAP$_{50}$, and mAP$_{50:95}$} on BDD$_v$ with only a 3\% reduction in recall after pruning 53\% of the labels, as shown in \cref{tab:metricmapv}.
Across all datasets, zero-shot TTN and TTN$_\text{\tiny D}$ improve Overall mAP$_{50:95}$ by 6\% and 9\%, respectively.

By selectively pruning disruptive candidates, both TTNs facilitate a particularly striking \textbf{Category Revival}, where the downstream model achieves non-trivial mAP on classes that previously exhibited zero recall. 
For example, at a $\tau=0.1$ threshold on LVIS, the AL baseline exhibits 0 recall for the 259 vulnerable classes regardless of VLM. 
The zero-shot TTN recovers recall for 37 of these categories (27, 10 for YOLOE, YOLOW), while $\text{TTN}_{\text{\tiny D}}$ recovers 28 (17, 11 for YOLOE, YOLOW).
This disproportionate improvement suggests that many transfer-vulnerable classes do not suffer from a lack of data, but rather from an exceptionally high noise-to-signal ratio. 
This recovery of the most challenging classes shows that the TTN's semantic pruning is a critical component for achieving robust generalization in fully autonomous pseudo-labeling pipelines.

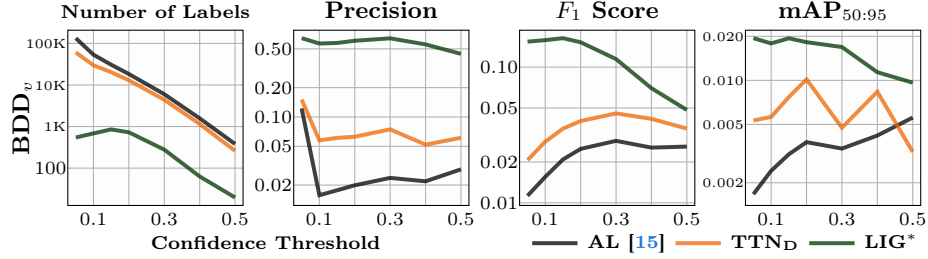
\begin{figure} [t!]
	\begin{minipage}{1\textwidth}
		\begin{tikzpicture}
	
	\definecolor{darkgrey176}{RGB}{176,176,176}
	\definecolor{gainsboro247}{RGB}{253,253,253}
	\definecolor{lightgrey204}{RGB}{204,204,204}
	
	\definecolor{lightgrey204}{RGB}{204,204,204}
	\definecolor{orangered2381020}{RGB}{238,102,0}
	\definecolor{slategrey119119153}{RGB}{119,119,153}
	\definecolor{darkgreen0510}{RGB}{0,51,0}
	
	\tikzstyle{every node}=[
	font=\scriptsize,
	]
	
	\begin{groupplot}[
		group style={
			group size=4 by 1,
			horizontal sep=0.67cm,
			vertical sep=0.3cm,
		},
		legend cell align={left},
		legend style={
			draw opacity=1,
			text opacity=1,
			anchor=south west,
			draw=none,
			fill=gainsboro247,
			at={(2.59,-0.33)}, 
			legend columns=3,
			column sep = 1pt,
			fill opacity=0,
		},
		width=3.9cm,
		height=3.9cm,
		ytick align=inside,
		grid=both,
		x grid style={darkgrey176},
		y grid style={darkgrey176},
		title style={at={(0.5,1.125)},anchor=north},
		xtick style={draw=none},
		ytick style={draw=none},
		xminorticks=true,
		minor x tick num=1,
		yminorticks=true,
		minor y tick num=1,
		axis background/.style={fill=gainsboro247},
		yticklabel style = {xshift=1.8ex},
		xticklabel style = {yshift=1.2ex},
		y label style={at={(axis description cs:0.27, 0.5)}},
		xticklabels={},
		x label style={at={(axis description cs:-0.16, 0.1)}},
		x label style={at={(axis description cs:1.13, 0.1)}},
		yticklabel style={/pgf/number format/fixed zerofill,/pgf/number format/fixed},
		]
\nextgroupplot[
tick align=outside,
tick pos=left,
title={\bf \scriptsize Number of Labels},
x grid style={darkgrey176},
xmin=2.75, xmax=52.25,
xtick style={color=black},
y grid style={darkgrey176},
ylabel={\bf \footnotesize BDD$_v$},
ymin=12.6507460803539, ymax=207796.264432195,
ymode=log,
ytick style={color=black},
ytick={2,10,100,1000,10000,100000},
yticklabels={2,10,100,\tiny 1\textrm{K},\tiny 10\textrm{K},\tiny 100\textrm{K}},
xtick={10,30,50},
xticklabels={0.1,0.3,0.5},
xlabel={\bf Confidence Threshold},
title style={at={(0.5,1.1)},anchor=north},
]
\addplot [line width=1.5pt, opacity=0.75,  black]
table {%
5 133666.666666667
10 53513.6666666667
15 30715
20 18360.6666666667
30 6004.33333333333
40 1577
50 381
};
\addlegendentry{\bf \scriptsize AL \cite{griff25AL}}
\addplot [line width=1.5pt, opacity=0.75,  orangered2381020]
table {%
5 60868.6666666667
10 29766
15 20323.3333333333
20 12963.6666666667
30 4412.66666666667
40 1149.66666666667
50 260.333333333333
};
\addlegendentry{\bf  \scriptsize TTN$_{\text{\tiny D}}$}
\addplot [line width=1.5pt, opacity=0.75,  darkgreen0510]
table {%
5 545.666666666667
10 684.333333333333
15 853.666666666667
20 726.666666666667
30 278
40 63
50 19.6666666666667
};
\addlegendentry{\bf  \scriptsize LIG$^*$}

\nextgroupplot[
tick align=outside,
tick pos=left,
title={\bf \footnotesize Precision},
x grid style={darkgrey176},
xmin=2.75, xmax=52.25,
xtick style={color=black},
y grid style={darkgrey176},
ymin=0.0123483349364516, ymax=0.7775350799716,
ymode=log,
ytick style={color=black},
ytick={0.02,0.05,0.1,0.2,0.5},
yticklabels={0.02,0.05,0.10,0.20,0.50},
xtick={10,30,50},
xticklabels={0.1,0.3,0.5},
]
\addplot [line width=1.5pt, opacity=0.75,  black]
table {%
5 0.122370077535494
10 0.0157283048605216
15 0.017677747868113
20 0.0198935135508089
30 0.0236867177877554
40 0.0218510371122909
50 0.0290148067487191
};
\addplot [line width=1.5pt, opacity=0.75,  orangered2381020]
table {%
5 0.150871254894361
10 0.0574720925069897
15 0.0610555037034888
20 0.0626485207977172
30 0.0746309980339829
40 0.0520105011297682
50 0.061122651502893
};
\addplot [line width=1.5pt, opacity=0.75,  darkgreen0510]
table {%
5 0.644083528062649
10 0.567766355016421
15 0.575363089070254
20 0.605774243413835
30 0.641542274664986
40 0.555555555555555
50 0.444444444444444
};

\nextgroupplot[
tick align=outside,
tick pos=left,
title={\bf \footnotesize $F_1$ Score},
x grid style={darkgrey176},
xmin=2.75, xmax=52.25,
xtick style={color=black},
y grid style={darkgrey176},
ymin=0.00956240807171253, ymax=0.186977188044595,
ymode=log,
ytick style={color=black},
xtick={10,30,50},
xticklabels={0.1,0.3,0.5},
ytick={0.01,0.02,0.05,0.1,0.2},
yticklabels={ 0.01, 0.02, 0.05, 0.10, 0.20},
]
\addplot [line width=1.5pt, opacity=0.75,  black]
table {%
5 0.0112663545508079
10 0.0155346831445511
15 0.0208583840165348
20 0.0249697363064287
30 0.0286177614470248
40 0.0255119769180309
50 0.0259708978497948
};
\addplot [line width=1.5pt, opacity=0.75,  orangered2381020]
table {%
5 0.02064175892584
10 0.0282112813200525
15 0.0352031707323557
20 0.0400630066398604
30 0.0456186901022625
40 0.0415715760128972
50 0.0352356339810374
};
\addplot [line width=1.5pt, opacity=0.75,  darkgreen0510]
table {%
5 0.154066614402541
10 0.157944811421428
15 0.16334157711853
20 0.152420022662266
30 0.114572484858497
40 0.0699190413796605
50 0.048323920988832
};

\nextgroupplot[
tick align=outside,
tick pos=left,
title={\bf \footnotesize mAP$_{50:95}$},
x grid style={darkgrey176},
xmin=0.0275, xmax=0.5225,
xtick style={color=black},
y grid style={darkgrey176},
ymin=0.00139979296372098, ymax=0.021997453411958,
ytick style={color=black},
ymode=log,
xtick={0.1,0.3,0.5},
xticklabels={0.1,0.3,0.5},
ytick={0.002,0.005,0.01,0.02},
yticklabels={ \tiny 0.002, \tiny 0.005, \tiny 0.010, \tiny 0.020},
]
\addplot [line width=1.5pt, opacity=0.75,  black]
table {%
0.05 0.00167268911833001
0.1 0.00241298539895842
0.15 0.00312902545314092
0.2 0.00378737562416271
0.3 0.00342340915319404
0.4 0.00420125641953512
0.5 0.00557521086837139
};
\addplot [line width=1.5pt, opacity=0.75,  orangered2381020]
table {%
0.05 0.00532429772607402
0.1 0.00563265038566981
0.15 0.00768609382615549
0.2 0.0101571738430333
0.3 0.00475100726080307
0.4 0.00836609607032111
0.5 0.00324938792183686
};
\addplot [line width=1.5pt, opacity=0.75,  darkgreen0510]
table {%
0.05 0.0194086238360556
0.1 0.0178772606796067
0.15 0.019375352787128
0.2 0.0182265375178434
0.3 0.0168885440049411
0.4 0.011366951647775
0.5 0.00964332402331082
};
\end{groupplot}

\end{tikzpicture}
	\end{minipage}%
	\vspace{-2.1em}
	\caption{
		{\bf The Post-Engine Interrogation Frontier} (log scale).
		The pruning gap on BDD$_v$ (left) shows the \textit{sheer magnitude} of VLM hallucinations that the LIG$^*$
		Oracle successfully discards, whereas even high-confidence filtering retains these errors.
		Precision, $F_1$ score, and mAP$_{50:95}$ performance across confidence thresholds $\tau$ (right) shows that TTN$_\text{\tiny D}$ significantly narrows the performance gap in the high-noise, low-$\tau$ regime.
	}
	\label{fig:frontbdd}
\end{figure}

\section{Conclusion and the Post-Engine Interrogation Frontier}

In this paper, we introduced the Label Imitation Game (LIG), a Turing-inspired framework that formalizes pseudo-label pruning as a dataset-wide adversarial interrogation.
By developing the Turing Test Network (TTN), we demonstrate that a model trained strictly on image classification can effectively judge and ``detoxify'' complex object detection pseudo-labels. 
Our experiments across four diverse datasets show that the TTN successfully suppresses systemic foundation model hallucinations, yielding zero-shot $F_1$-score gains of 28\% for the most challenging categories---rising to 44\% with task-specific fine-tuning ($\text{TTN}_{\text{\tiny D}}$).
Crucially, both models facilitate Category Revival, enabling downstream detectors to recover from zero recall on dozens of transfer-vulnerable classes.
Our results establish that global semantic-contextual logic is a robust alternative to local spatial-geometric verification. 
By releasing our LIG framework and pre-trained TTN weights, we provide the vision community with a lightweight, task-agnostic module that enhances pseudo-labeling rigor at scale.

Looking forward, a significant \textbf{Post-Engine Interrogation Frontier} remains for future work.
As shown in \cref{fig:frontbdd}, the optimal pruning regime includes low-confidence labels, where the most authentic signal is often discarded by traditional filtering.
$\text{TTN}_{\text{\tiny D}}$ substantially narrows this performance gap, but additional gains must be won from increasingly subtle and difficult-to-identify hallucinations.
We postulate that integrating multi-model consensus and LIG-aware spatial redundancy reduction will close this gap. 
To broaden applicability, expanding the TTN's interrogative capacity to video \cite{finepseudo24eccv} and 3D modalities \cite{sal24ECCV} provides new frontiers for fully autonomous, noise-robust machine learning.

%
%
\bibliographystyle{splncs04}
\bibliography{ttn_ref}

\setcounter{section}{0}
\setcounter{figure}{0}
\setcounter{table}{0}
\setcounter{equation}{0}

\renewcommand{\thesection}{S\arabic{section}}
\renewcommand{\thefigure}{S\arabic{figure}}
\renewcommand{\thetable}{S\arabic{table}}
\renewcommand{\theequation}{S\arabic{equation}}

\renewcommand{\theHsection}{S\arabic{section}}
\renewcommand{\theHfigure}{S\arabic{figure}}
\renewcommand{\theHtable}{S\arabic{table}}
\renewcommand{\theHequation}{S\arabic{equation}}

\newpage
\begin{center}
	\vspace{12mm}
	\Large{\bf Supplementary Material} \\ \vspace{7mm}
	\Large\textbf{The Label Imitation Game: Turing Test Network for Zero-Shot Pseudo-Label Pruning} \\
	\vspace{7mm}
	\normalsize Brent A. Griffin and Jason J. Corso
	\vspace{1mm}
\end{center}

\section*{Contents of the Supplementary Material}
\begin{enumerate}
	\item[] \textbf{Section \ref{sec:supnet}: Network Architecture Specifications}
	\item[] \textbf{Section \ref{sec:compare}: Extended Method Comparisons}
	\item[] \textbf{Section \ref{sec:ablate}: Methodological Validation and Sensitivity Analysis}
	\item[] \textbf{Section \ref{sec:compute}: Computational Efficiency \& Scalability}
	\item[] \textbf{Section \ref{sec:supexp}: Extended Performance Benchmarks} 
	\item[] \textbf{Section \ref{sec:supsetup}: Experimental Protocol} 
\end{enumerate}
\hrule
\vspace{1em}

\section{Network Architecture Specifications}
\label{sec:supnet}

\begin{figure}[t!]
	\centering
	\includegraphics[width=0.975\textwidth]{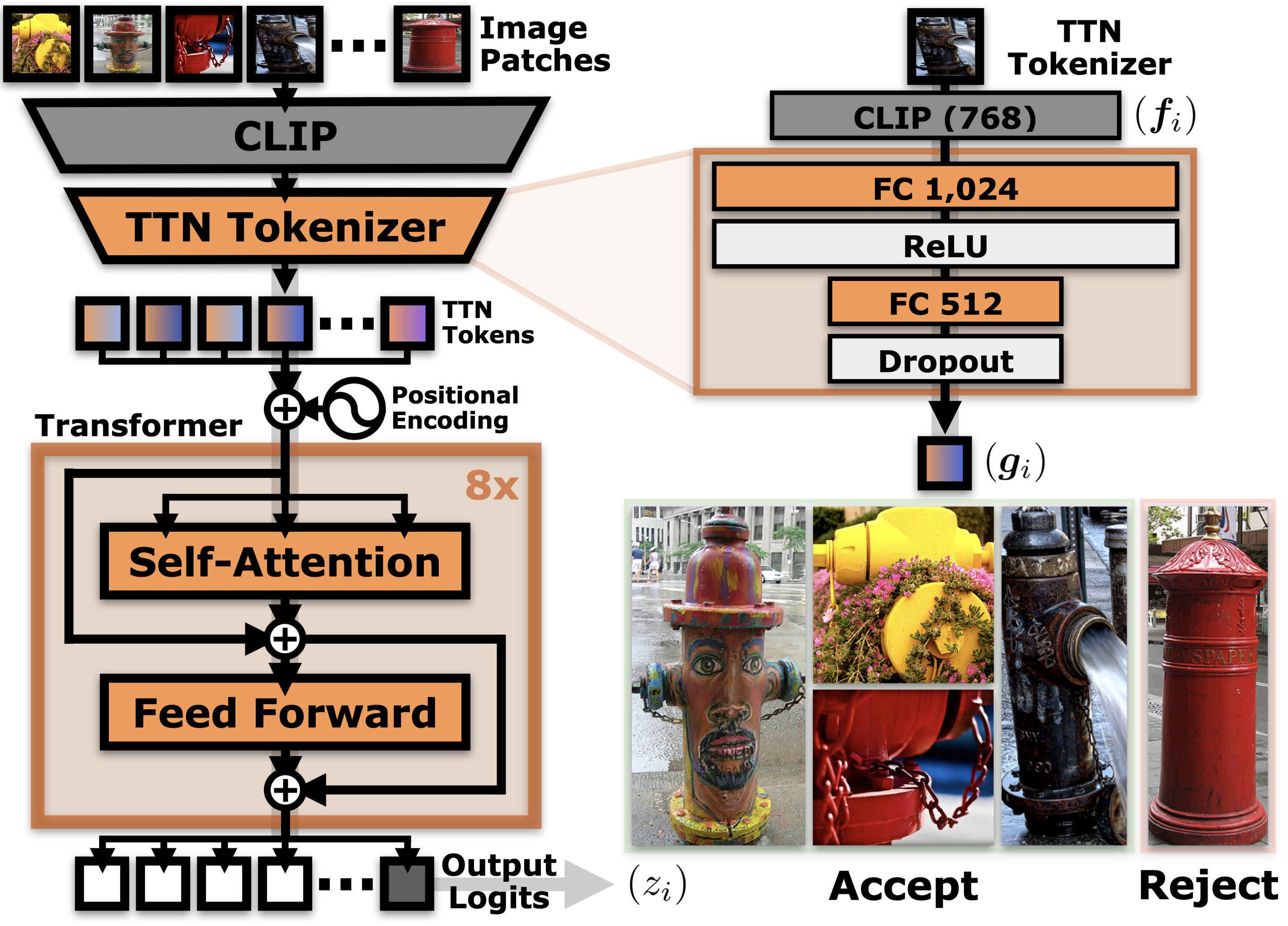}
	\vspace{-0.8em}
	\caption{{\bf Turing Test Network (TTN) Architecture Specifications}.
		The TTN processes image patches via a custom tokenizer (\cref{sec:net2}) and an eight-layer transformer (\cref{sec:net3}).
		Using non-masked self-attention, the TTN evaluates relational and sequential context to produce Accept/Reject logits ($z_i$) for each input without any class labels, enabling zero-shot task transfer from image classification training to object detection pseudo-label pruning.
		The TTN learns 26.5~\textrm{M} parameter weights (orange).
	}
	\label{fig:supschematic}
\end{figure}

To elaborate on \cref{sec:ttn}, we provide the full architectural details of the Turing Test Network (TTN), which consists of three stages: a frozen feature extractor, a custom-learned tokenizer, and a transformer-based reasoning block.
These architectural designs enable the TTN to build and evaluate semantic and spatial context without class labels, thus facilitating zero-shot transfer (\cref{fig:supschematic}).

\subsection{Stage 1: Frozen Feature Extractor}
\label{sec:net1}
TTN uses a pre-trained vision-language model to establish a high-level semantic representation of labels. 
Specifically, we use the CLIP ViT-L/14 model \cite{clip} to encode the underlying image patches defined by each label. 
We keep the encoder frozen for a robust, general-purpose initial feature space $\vf_i \in \sR^{\text{768}}$.

\subsection{Stage 2: Custom-Learned TTN Tokenizer}
\label{sec:net2}
To bridge the gap between static CLIP embeddings and the transformer reasoning space, a lightweight multi-layer perceptron (MLP) transforms the initial features into ``TTN Tokens.'' 
The MLP sequentially processes $\vf_i$ through a fully-connected 1,024-D layer, a ReLU activation, a fully-connected 512-D layer, and, finally, a Dropout layer ($p=0.2$).
The output tokens $\vg_i(\vf_i) \in \sR^{\text{512}}$ (\cref{eq:g}) are the condensed visual and semantic information TTN evaluates during the Label Imitation Game (LIG).

\subsection{Stage 3: Non-Masked Transformer-based Reasoning Block}
\label{sec:net3}
The core of TTN is an eight-layer transformer that evaluates each token $\vg_i \in \sG$. 
Unlike standard classifiers that process examples in isolation, this block utilizes non-masked Multi-Head Self-Attention (MHSA) to evaluate many tokenized labels across an arbitrary number of data examples simultaneously (\cref{eq:setg}).
To preserve relational and sequential context across tokens, we add learned positional encodings via an embedding table to the transformer input.
The MHSA block splits the input tokens into eight heads, while the subsequent feed-forward block reuses the tokenizer's MLP design with new weights.
To ensure stable gradient flow, each transformer layer has residual connections where the input is normalized \textit{only} for sub-layer operations and added to each sub-layer output. 
The final output state of each token is passed through a final layer normalization and linear projection layer to produce the Accept/Reject logits ($z_i$ in \cref{eq:loss}).

\section{Extended Method Comparisons}
\label{sec:compare}

We provide additional method comparisons to supplement the related work (\cref{sec:relate}) and experimental evaluation (\cref{sec:exp}) of the main paper.
To provide more context for related work, we compare the methodological setup and procedural requirements for several representative methods in \cref{tab:method}.
Importantly, only TTN and the AL baseline are zero-shot compatible with existing object detection frameworks.
All other methods require ground-truth detection labels or, more commonly, invasive changes to the underlying model architecture, training pipeline, or even the core methodology itself for detection compatibility.

\begin{table}[t]
	\setlength{\tabcolsep}{4.4pt}
	\centering
	\caption{{\bf Methodological Setup \& Procedural Requirements Comparison}}
	\label{tab:rel}
	\vspace{-0.75em}
	\begin{tabular}{| l | c | c | c |}
		\hline 
		\rowcolor{tableheader} 	&   \bf Developed    &   \multicolumn{2}{c |}{\bf Operates \textit{Without}: \rule{0pt}{2.25ex}}  \\
		\rowcolor{tableheader} 	& \bf  for Object &   \multicolumn{1}{ c }{\bf Architecture or} & \bf  Detection  \\
		\rowcolor{tableheader} 	\multicolumn{1}{| c |}{\bf \footnotesize Method} &    \bf Detection &  \multicolumn{1}{ c }{\bf Pipeline Changes} &  \multicolumn{1}{ c |}{\bf GT Labels} \\ \hline
		\bf AL \cite{griff25AL}, {\bf TTN} & \bf Yes & \bf Yes & \bf Yes \\ \hline
		CLIP kNN Prune, TTN$_{\text{\tiny D}}$ & \bf Yes & \bf Yes & No \\ \hline
		SoftTeacher \cite{Xu_2021_ICCV} & \bf Yes & No & No \\ \hline
		Co-teaching \cite{NEURIPS2018_a19744e2}, FixMatch \cite{NEURIPS2020_06964dce}, &  &  &  \\ 
		kNN-CLIP \cite{gui2024knnclip}, MentorNet \cite{pmlr-v80-jiang18c} & No & No & No \\ \hline
	\end{tabular}
	\label{tab:method}
\end{table}

To expand the method evaluation in \cref{sec:exp}, following kNN-CLIP \cite{gui2024knnclip}, we implement kNN baselines that prune or relabel pseudo-labels based on the majority-vote consistency among the 16 nearest ground-truth CLIP embeddings (using $\leq$ 100 ground-truth labels per class).
We also implement kNN baselines with TTN Token embeddings ($\vg_i(\vf_i) \in \sR^{\text{512}}$ in \cref{sec:net2}) to compare their label pruning efficacy directly against the frozen CLIP embeddings ($\vf_i \in \sR^{\text{768}}$ in \cref{sec:net1}).
As shown in \cref{tab:compare}, both TTN and TTN$_\text{\tiny D}$ Token-based kNN outperform CLIP, which supports the decision to include the TTN Tokenizer in the full TTN architecture (see \cref{fig:supschematic}).
Furthermore, while TTN Token and CLIP kNN pruning outperform the AL baseline, zero-shot pruning with the full TTN architecture and corresponding semantic logic surpasses all kNN baselines \textit{despite their use of ground-truth labels}.
Notably, CLIP kNN relabeling \textit{decreases} AL performance, which further motivates our focus on pruning-based approaches.

\begin{table}[t]
	\setlength{\tabcolsep}{5.4 pt}
	\centering
	\caption{{\bf Extended Method Evaluation}. Highlighted rows correspond to methods from the main paper (left), and bold font indicates best performance (right).}
	\vspace{-0.75em}
	\footnotesize
	\begin{tabular}{| l | c | c | c | c | c | c | }
		\hline 
		\rowcolor{tableheader} \rule{0pt}{2.25ex} & & \bf GT & \multicolumn{4}{ c |}{\bf \footnotesize Mean of VLM, Confidence}  \\ 
		\rowcolor{tableheader} \rule{0pt}{2.25ex} & \bf Label & \bf Labels & \multicolumn{4}{ c |}{\bf \footnotesize Threshold, Class, \& Dataset}  \\ 
		\rowcolor{tableheader} \rule{0pt}{2.25ex} \bf \quad {\tiny ~} Method & \bf Operation &  \bf per Class & \multicolumn{1}{ c }{\bf \footnotesize Recall} & \multicolumn{1}{ c }{\bf \footnotesize Prec.} & \multicolumn{1}{ c }{\bf \footnotesize $F_1$} & \multicolumn{1}{ c |}{\bf \footnotesize $F_2$}  \\ \hline
		\rowcolor{defaultrow} {\bf TTN$_{\text{\tiny D}}$} & Prune & $\leq$ 100	&	0.437	&	0.447	&	\bf 0.387	&	\bf 0.398	\\	\hline
		\rowcolor{defaultrow} {\bf TTN } & Prune & \bf zero-shot &	0.426	&	0.436	&	0.376	&	0.386	\\	\hline
		TTN$_{\text{\tiny D}}$ Token kNN & Prune & $\leq$ 100	&	0.367	&	0.508	&	0.373	&	0.360	\\	\hline
		TTN Token kNN & Prune & $\leq$ 100	&	0.357	&	0.512	&	0.367	&	0.352	\\	\hline
		CLIP kNN & Prune & $\leq$ 100	&	0.347	& \bf 0.523	&	0.366	&	0.346	\\	\hline
		\rowcolor{defaultrow} \bf AL& --  & \bf zero-shot &	\bf 0.453	&	0.400	&	0.362	&	0.387	\\	\hline
		CLIP kNN & Relabel & $\leq$ 100	&	0.402	&	0.285	&	0.275	&	0.312	\\	\hline
	\end{tabular}
	\label{tab:compare}
\end{table}

\section{Methodological Validation and Sensitivity Analysis}
\label{sec:ablate}

\begin{figure}
	\centering
	\begin{minipage}{1\textwidth}
		\begin{tikzpicture}
	
	\definecolor{darkgrey176}{RGB}{176,176,176}
	\definecolor{gainsboro247}{RGB}{253,253,253}
	\definecolor{lightgrey204}{RGB}{204,204,204}
	
	\definecolor{lightgrey204}{RGB}{204,204,204}
	\definecolor{orangered2381020}{RGB}{238,102,0}
	\definecolor{slategrey119119153}{RGB}{119,119,153}
	\definecolor{darkgreen0510}{RGB}{0,51,0}
	
	\tikzstyle{every node}=[
	font=\scriptsize,
	]
	
	\begin{groupplot}[
		group style={
			group size=4 by 1,
			horizontal sep=0.7cm,
			vertical sep=0.3cm,
		},
		legend cell align={left},
		legend style={
			draw opacity=1,
			text opacity=1,
			anchor=south west,
			draw=none,
			fill=gainsboro247,
			at={(2.63,-0.33)},
			legend columns=3,
			column sep = 1pt,
			fill opacity=0,
		},
		width=3.9cm,
		height=3.9cm,
		ytick align=inside,
		grid=both,
		x grid style={darkgrey176},
		y grid style={darkgrey176},
		title style={at={(0.5,1.125)},anchor=north},
		xtick style={draw=none},
		ytick style={draw=none},
		xminorticks=true,
		minor x tick num=1,
		yminorticks=true,
		minor y tick num=1,
		axis background/.style={fill=gainsboro247},
		yticklabel style = {xshift=1.6ex},
		xticklabel style = {yshift=1.2ex},
		xticklabels={},
		x label style={at={(axis description cs:1.1, 0.1)}},
		yticklabel style={/pgf/number format/fixed zerofill,/pgf/number format/fixed},
		]
\nextgroupplot[
tick align=outside,
tick pos=left,
title={\bf \footnotesize Recall},
x grid style={darkgrey176},
xmin=0.05, xmax=3,
xtick style={color=black},
y grid style={darkgrey176},
ymin=0.3, ymax=0.5,
xmode=log,
ytick style={color=black},
ytick={0.3,0.35,0.4,0.45,0.5},
yticklabels={ 0.30, 0.35, 0.40, 0.45, 0.50},
xtick={0.1,0.2,0.5,1,2},
xticklabels={ 0.1~, ~0.2, 0.5, 1, 2},
xlabel={\bf \footnotesize $p_w$ (log scale)},
]
\addplot [line width=1.5pt, opacity=0.75,  black]
table {%
0.01 0.453
1 0.453
3 0.453
};
\addlegendentry{\bf \scriptsize AL \cite{griff25AL}}
\addplot [line width=1.5pt, opacity=0.75,  slategrey119119153]
table {%
2	0.406
1	0.426
0.5	0.371
0.2	0.325
0.1	0.453
};
\addlegendentry{\bf  \scriptsize TTN}
\addplot [line width=1.5pt, opacity=0.75,  orangered2381020]
table {%
2	0.341
1	0.397
0.5	0.418
0.2	0.437
0.1	0.431
};
\addlegendentry{\bf  \scriptsize TTN$_{\text{\tiny D}}$}

\nextgroupplot[
tick align=outside,
tick pos=left,
title={\bf \footnotesize Precision},
x grid style={darkgrey176},
xmin=2.75, xmax=52.25,
xmin=0.05, xmax=3,
xtick style={color=black},
y grid style={darkgrey176},
ymin=0.0123483349364516, ymax=0.7775350799716,
xmode=log,
ymin=0.375, ymax=0.55,
ytick style={color=black},
ytick={0.4,0.45,0.5,0.55},
yticklabels={ 0.40, 0.45, 0.50, 0.55},
xtick={0.1,0.2,0.5,1,2},
xticklabels={ 0.1~, ~0.2, 0.5, 1, 2},
]
\addplot [line width=1.5pt, opacity=0.75,  black]
table {%
	0.01 0.4
	1 0.4
	3 0.4
};
\addplot [line width=1.5pt, opacity=0.75,  slategrey119119153]
table {%
2	0.452
1	0.436
0.5	0.470
0.2	0.470
0.1	0.400
};
\addplot [line width=1.5pt, opacity=0.75,  orangered2381020]
table {%
2	0.536
1	0.494
0.5	0.471
0.2	0.447
0.1	0.456
};

\nextgroupplot[
tick align=outside,
tick pos=left,
title={\bf \footnotesize $F_1$ Score},
x grid style={darkgrey176},
xtick style={color=black},
xmin=0.05, xmax=3,
y grid style={darkgrey176},
xmode=log,
ymin=0.31, ymax=0.4,
ytick style={color=black},
ytick={0.32,0.34,0.36,0.38,0.4},
yticklabels={ 0.32,  0.34,  0.36, 0.38, 0.40},
xtick={0.1,0.2,0.5,1,2},
xticklabels={ 0.1~, ~0.2, 0.5, 1, 2},
]
\addplot [line width=1.5pt, opacity=0.75,  black]
table {%
	0.01 0.362
	1 0.362
	3 0.362
};
\addplot [line width=1.5pt, opacity=0.75,  slategrey119119153]
table {%
2	0.374
1	0.376
0.5	0.361
0.2	0.328
0.1	0.362
};
\addplot [line width=1.5pt, opacity=0.75,  orangered2381020]
table {%
2	0.370
1	0.390
0.5	0.390
0.2	0.387
0.1	0.388
};

\nextgroupplot[
tick align=outside,
tick pos=left,
title={\bf \footnotesize $F_2$ Score},
x grid style={darkgrey176},
xmin=0.05, xmax=3,
xtick style={color=black},
y grid style={darkgrey176},
xmode=log,
ymin=0.31, ymax=0.41,
ytick style={color=black},
ytick={0.3,0.32,0.34,0.36,0.38,0.4},
yticklabels={ 0.30,  0.32,  0.34,  0.36, 0.38, 0.40},
xtick={0.1,0.2,0.5,1,2},
xticklabels={ 0.1~, ~0.2, 0.5, 1, 2},
]
\addplot [line width=1.5pt, opacity=0.75,  black]
table {%
	0.01  0.387
	1  0.387
	3 0.387
};
\addplot [line width=1.5pt, opacity=0.75,  slategrey119119153]
table {%
2	0.377
1	0.386
0.5	0.354
0.2	0.315
0.1	0.387
};
\addplot [line width=1.5pt, opacity=0.75,  orangered2381020]
table {%
2	0.344
1	0.382
0.5	0.392
0.2	0.398
0.1	0.396
};

\end{groupplot}

\end{tikzpicture}
	\end{minipage}%
	\vspace{-2.1em}
	\caption{
		{\bf Pruning Aggression ($p_w$) vs. Mean Performance}
	}
	\label{fig:pw}
\end{figure}
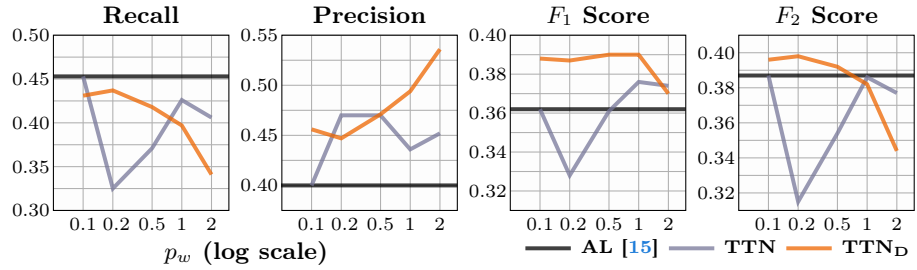

To quantitatively validate design decisions from the main paper, we include supplementary results for methodological ablations and sensitivity analyses in \cref{fig:pw} and \cref{tab:ablate}.
While TTN is robust to many variations, ablations show TTN's chosen pruning aggression ($p_w=1$), sample distribution ($\pi_\text{1,2}=\frac{1}{2}$), token design ($\vg_i \in \sR^{512}$), and number of input tokens (10 Reference, 1 Candidate) enable TTN to train from scratch on classification data then perform zero-shot pruning on detection labels with high $F_1$ \textit{and} $F_2$ scores.
Notably, training TTN from scratch with the lowest pruning aggression ($p_w=0.1$) causes TTN to simply accept \textit{all} labels, which results in identical performance to the AL baseline.

\setlength{\tabcolsep}{5.9 pt}
\begin{table}
	\centering
	\caption{{\bf TTN \& TTN$_{\text{\tiny D}}$ Method Ablation Evaluation}. Highlighted rows correspond to default settings (left), and bold font indicates best performance (right).}
	\vspace{-0.75em}
	\footnotesize
	\begin{tabular}{| l | l | c | c | c | c | }
		\hline 
		\rowcolor{tableheader} \rule{0pt}{2.25ex} & & \multicolumn{4}{ c |}{\bf \footnotesize Mean of VLM, Confidence}  \\ 
		\rowcolor{tableheader} \rule{0pt}{2.25ex} & & \multicolumn{4}{ c |}{\bf \footnotesize Threshold, Class, \& Dataset}  \\ 
		\rowcolor{tableheader} \rule{0pt}{2.25ex} \footnotesize \bf Method & \multicolumn{1}{c |}{\bf Ablation Details} & \multicolumn{1}{ c }{\bf \footnotesize Recall} & \multicolumn{1}{ c }{\bf \footnotesize Prec.} & \multicolumn{1}{ c }{\bf \footnotesize $F_1$} & \multicolumn{1}{ c |}{\bf \footnotesize $F_2$} \\ \hline
		\multicolumn{5}{l}{\bf Training Sample Distribution $\pi_k = P(\sS=k)$, $\sum \pi_k =1$} \rule{0pt}{3ex} \\ \hline
		\rowcolor{defaultrow}	{\bf TTN$_{\text{\tiny D}}$} 	& {$\pi_\text{1}=\frac{1}{2}$, $\pi_\text{2}=\frac{1}{4}$, $\pi_\text{3,4}=\frac{1}{8}$}	\rule{0pt}{2.25ex} &	0.437	&	0.447	&	\bf 0.387	&	\bf 0.398	\\	\hline
		TTN$_{\text{\tiny D}}$	&  {OCG-Pseudo} {$\pi_\text{1,3}=\frac{1}{2}$, $\pi_\text{2,4}=0$} \rule{0pt}{2.25ex}	&	0.431	&	0.451	&	0.385	&	0.394	\\	\hline
		TTN$_{\text{\tiny D}}$	& Uniform {$\pi_\text{1,2,3,4}=\frac{1}{4}$} \rule{0pt}{2.25ex} &	\bf 0.439	&	0.442	&	0.384	&	0.396	\\	\hline
		TTN$_{\text{\tiny D}}$	&  {OCG-Human} {$\pi_\text{1,2}=\frac{1}{2}$, $\pi_\text{3,4}=0$} \rule{0pt}{2.25ex} &	0.434	&	0.441	&	0.380	&	0.393	\\	\hline
		TTN$_{\text{\tiny D}}$	&  {PLG} {$\pi_\text{1,4}=\frac{1}{2}$, $\pi_\text{2,3}=0$} \rule{0pt}{2.25ex} &	0.408	&	\bf 0.459	&	0.372	&	0.376	\\	\hline \hline
		\rowcolor{defaultrow}	{\bf TTN}	&	 {$\pi_\text{1,2}=\frac{1}{2}$, $\pi_\text{3,4}=0$} \rule{0pt}{2.25ex} &	\bf 0.426	&	0.436	&	\bf 0.376	&	\bf 0.386	\\	\hline
		TTN	& {$\pi_\text{1}=\frac{1}{4}$, $\pi_\text{2}=\frac{3}{4}$, $\pi_\text{3,4}=0$}	\rule{0pt}{2.25ex}	&	0.415	&	0.442	&	0.374	&	0.381	\\	\hline
		TTN	& {$\pi_\text{1}=\frac{3}{4}$, $\pi_\text{2}=\frac{1}{4}$, $\pi_\text{3,4}=0$}	\rule{0pt}{2.25ex}	&	0.385	&	\bf 0.467	&	0.373	&	0.366	\\	\hline
		\multicolumn{5}{l}{\bf Token Complexity $\vg_i \in \sR^{n}$}  \rule{0pt}{3ex} \\ \hline
		\rowcolor{defaultrow}	{\bf TTN}	&   $n=$~512	&	\bf 0.426	&	0.436	&	\bf 0.376	&	\bf 0.386	\\	\hline
		TTN	&   $n=$~256	&	0.424	&	0.438	&	\bf 0.376	&	0.385	\\	\hline
		TTN	&   $n=$~1,024	&	0.387	&	\bf 0.447	&	0.365	&	0.363	\\	\hline
		\multicolumn{5}{l}{\bf Number of Reference \& Candidate Token Inputs}  \rule{0pt}{3ex} \\ \hline
		TTN	&   10 Reference, 2 Candidate	&	0.406	&		\bf 0.455	&		\bf 0.377	&	0.378	\\	\hline
		\rowcolor{defaultrow}	{\bf TTN}	&   10 Reference, 1 Candidate	&		\bf 0.426	&	0.436	&	0.376	&	\bf 0.386	\\	\hline
		TTN	&   20 Reference, 1 Candidate	&	0.399	&	0.454	&	0.373	&	0.373	\\	\hline
		TTN 	&  {\tiny ~~}5 Reference, 1 Candidate	&	0.404	&	0.437	&	0.367	&	0.372	\\	\hline
		TTN 	&  10 Reference, 5 Candidate	&	0.382	&	0.441	&	0.357	&	0.357	\\	\hline
		\multicolumn{5}{l}{\bf Pruning Reference Set Configuration (\& Mean Accuracy)} \rule{0pt}{3ex} \\ \hline
		TTN	&   $\leq$ 80 ~~{\tiny ~}~GT Reference ~{\tiny ~}~(100\%)	&	\bf 0.442	&	0.427	&	\bf 0.379	&	\bf 0.395	\\	\hline
		\rowcolor{defaultrow}	{\bf TTN}	&   $\leq$ 1,000 Self-Referential (63.6\%)	&	0.426	&	0.436	&	0.376	&	0.386	\\	\hline
		TTN 	&  $\leq$ 100 {\tiny ~~}~Self-Referential (70.9\%)&	0.414	&	\bf 0.448	&	0.376	&	0.381	\\	\hline
	\end{tabular}
	\label{tab:ablate}
\end{table}

Subsequent to TTN training, the pruning reference set ablations in \cref{tab:ablate} reveal a trade-off.
Reducing the self-referential pruning set size from $\le$ 1,000 to $\le$~100 highest-confidence samples improves the reference set's accuracy ($ 63.6\% \rightarrow 70.9\%$) but degrades final pruning performance ($0.386 \rightarrow 0.381~F_2$ score) due to a reduction in contextual breadth. 
However, if available, using $\le$ 80 higher-quality ground-truth labels bridges this contextual gap, enabling TTN to approach the pruning performance of fine-tuned TTN$_\text{\tiny D}$ (3.95 and 3.98 $F_2$ scores respectively).

For fine-tuned TTN$_\text{\tiny D}$, increasing $p_w$ increases precision but also decreases recall, with the chosen $p_w=0.2$ achieving the highest $F_2$ score in \cref{fig:pw}. 
TTN$_\text{\tiny D}$ train distribution ablations reveal that OCG-Pseudo ($\pi_3$) is the best individual game---followed by OCG-Human and PLG---yet the nominal TTN$_{\text{\tiny D}}$ performs best in \cref{tab:ablate}, validating that using all games is optimal.

\section{Computational Efficiency \& Scalability}
\label{sec:compute}

To establish the computational efficiency and scalability of TTN pruning, we provide a runtime comparison for several TTN configurations in \cref{tab:runtime}.
While runtime depends on the number of pseudo-labels, generating the underlying CLIP embeddings remains the primary computational bottleneck, requiring more time than TTN pruning in all cases.
Standard TTN pruning averages 31 minutes per dataset, which scales down to less than one minute via class-wise parallelization.

\setlength{\tabcolsep}{8.3 pt}
\begin{table}
	\centering
	\caption{{\bf Runtime Comparison} on a single NVIDIA L40S GPU.
	TTN$^*$ is estimated runtime if pruning in parallel by class.
	Speedup via data parallelization is also viable.
}
	\vspace{-0.75em}
	\footnotesize
	\begin{tabular}{| l | r | r | r | }
		\hline 
		\rowcolor{tableheader}  \multicolumn{1}{| c |}{\footnotesize \bf Full Dataset Process} &  \multicolumn{1}{ c |}{\footnotesize \bf Min. \rule{0pt}{2.5ex}} &  \multicolumn{1}{ c |}{\footnotesize \bf Max.} &  \multicolumn{1}{ c |}{\footnotesize \bf Mean} \\ \hline
		\rowcolor{ligtable} TTN$^{*}_{\tiny ~~}$ Pruning  $|\sG^{(c^*)}| \leq$ 1,000 Reference \rule{0pt}{2.5ex} &	0.8 s	&	0.33 h	&	54.5 s	\\	\hline
		TTN$_{\text{\tiny D}}$ Pruning $|\sG^{(c^*)}| \leq$ 80 ~~{\tiny ~}~Reference \rule{0pt}{2.5ex}	& \bf	6.1 \textrm{s}	&	\bf 0.35 \textrm{h}	& \bf	167.7 s	\\	\hline
		TTN  ~{\tiny ~}Pruning $|\sG^{(c^*)}| \leq$ 100 {\tiny ~~} Reference \rule{0pt}{2.5ex}	&	8.1 s	&	0.42 h	&	205.8 s	\\	\hline
		TTN  ~{\tiny ~}Pruning $|\sG^{(c^*)}| \leq$ 1,000 Reference \rule{0pt}{2.5ex}	&	63.4 s	&	4.31 h	&	0.52 h	\\	\hline
		Generate CLIP Embeddings \rule{0pt}{2.25ex}	&	354.6 s	&	12.95 h	&	1.90 h	\\	\hline \hline
		Number of Pseudo-Labels	\rule{0pt}{2.25ex} &	37,800	&	10,078,051	&	1,182,061	\\	\hline
	\end{tabular}
	\label{tab:runtime}
\end{table}

\section{Extended Performance Benchmarks}
\label{sec:supexp}

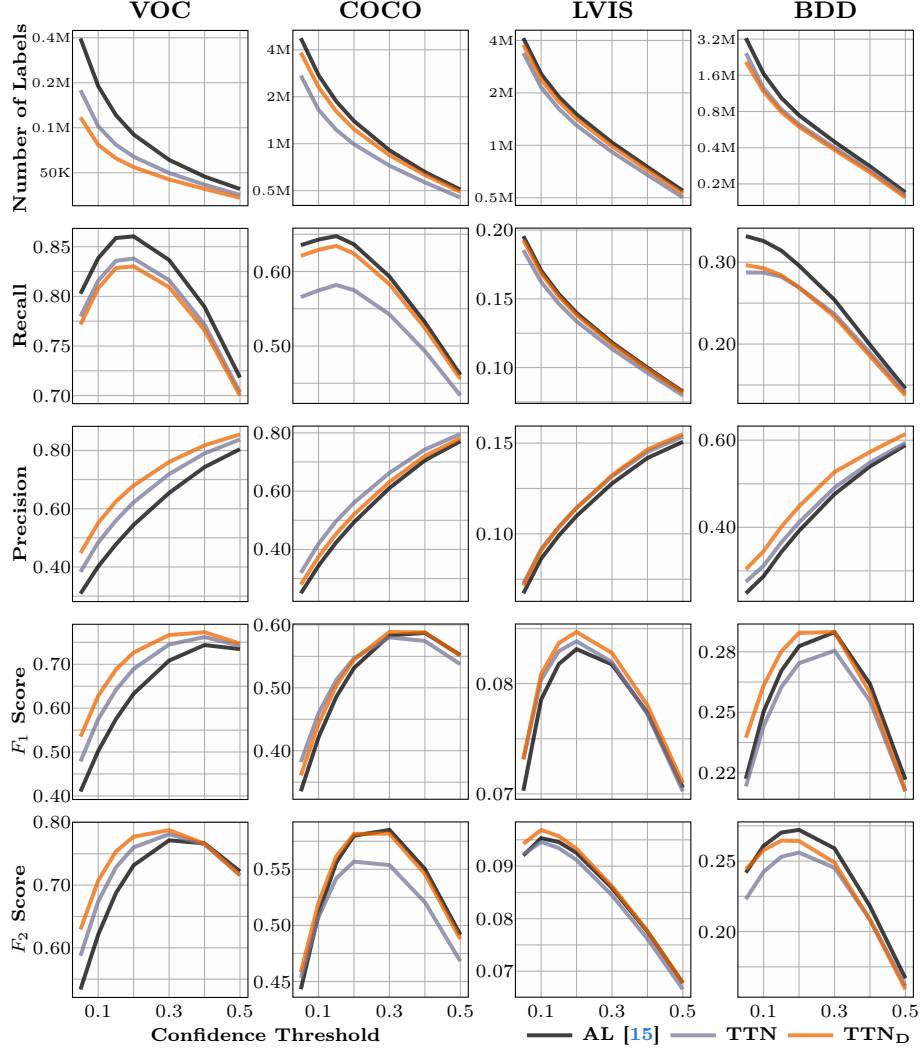
\begin{figure}[t!]
	\begin{minipage}{0.26\textwidth}
		\begin{tikzpicture}

\definecolor{darkgrey176}{RGB}{176,176,176}
\definecolor{gainsboro247}{RGB}{253,253,253}
\definecolor{lightgrey204}{RGB}{204,204,204}

\definecolor{lightgrey204}{RGB}{204,204,204}
\definecolor{orangered2381020}{RGB}{238,102,0}
\definecolor{slategrey119119153}{RGB}{119,119,153}

\tikzstyle{every node}=[
	font=\scriptsize,
]

\begin{groupplot}[
	group style={
		group size=1 by 5,
		vertical sep=0.3cm,
		},
		legend cell align={left},
		legend style={
			draw opacity=1,
			text opacity=1,
			anchor=south west,
			draw=none,
			fill=gainsboro247,
			at={(2.53,-0.33)},
			legend columns=3,
			column sep = 1pt,
			fill opacity=0,
		},
		width=3.9cm,
		height=3.9cm,
		ytick align=inside,
		grid=both,
		x grid style={darkgrey176},
		y grid style={darkgrey176},
		title style={at={(0.5,1.125)},anchor=north},
		xtick style={draw=none},
		ytick style={draw=none},
		xminorticks=true,
		minor x tick num=1,
		yminorticks=true,
		minor y tick num=1,
		axis background/.style={fill=gainsboro247},
		yticklabel style = {xshift=1.8ex},
		xticklabel style = {yshift=1.2ex},
		y label style={at={(axis description cs:0.23, 0.5)}},
		xticklabels={},
		x label style={at={(axis description cs:1.1, 0.1)}},
		yticklabel style={/pgf/number format/fixed zerofill,/pgf/number format/fixed},
]
\nextgroupplot[
legend cell align={left},
log basis y={10},
tick align=outside,
tick pos=left,
title={\bf \footnotesize VOC},
x grid style={darkgrey176},
xmin=2.75, xmax=52.25,
xtick style={color=black},
y grid style={darkgrey176},
ylabel={\bf \scriptsize Number of Labels},
ymin=30101.8413012817, ymax=447399.771783098,
ymode=log,
ytick style={color=black},
ytick={50000,100000,200000,400000},
yticklabels={
  \(\displaystyle $\text{\tiny 50}\textrm{\tiny K}$\),
    \(\displaystyle $\text{\tiny 0.1}\textrm{\tiny M}$\),
  \(\displaystyle $\text{\tiny 0.2}\textrm{\tiny M}$\),
  \(\displaystyle $\text{\tiny 0.4}\textrm{\tiny M}$\),
}
]
\addplot [line width=1.5pt, opacity=0.75, black]
table {%
5 395747.666666667
10 189853.666666667
15 121474.333333333
20 90117
30 60998.6666666667
40 47172
50 38930.6666666667
};
\addplot [line width=1.5pt, opacity=0.75, slategrey119119153]
table {%
5 178360.333333333
10 102401.666666667
15 77012.6666666667
20 63998
30 49846
40 41623
50 35582.3333333333
};
\addplot [line width=1.5pt, opacity=0.75, orangered2381020]
table {%
5 117345
10 76883.6666666667
15 62480.3333333333
20 54519.3333333333
30 45106.6666666667
40 38991
50 34030.6666666667
};

\nextgroupplot[
tick align=outside,
tick pos=left,
ylabel={\bf \scriptsize Recall},
x grid style={darkgrey176},
xmin=2.75, xmax=52.25,
xtick style={color=black},
y grid style={darkgrey176},
ymin=0.691941135210384, ymax=0.868408161733527,
ytick style={color=black}
]
\addplot [line width=1.5pt, opacity=0.75, black]
table {%
5 0.802671297270523
10 0.838398447722968
15 0.858733308973256
20 0.860386933255202
30 0.836575635872848
40 0.789375191978657
50 0.718271936377296
};
\addplot [line width=1.5pt, opacity=0.75, slategrey119119153]
table {%
5 0.779789336784006
10 0.81574895433122
15 0.835784330200299
20 0.838056918972587
30 0.816267334434082
40 0.771554341676164
50 0.703449897513244
};
\addplot [line width=1.5pt, opacity=0.75, orangered2381020]
table {%
5 0.771564938579979
10 0.808103444486215
15 0.828506916314796
20 0.830211749804011
30 0.809260600409202
40 0.76592265577155
50 0.699962363688708
};

\nextgroupplot[
tick align=outside,
tick pos=left,
ylabel={\bf \scriptsize Precision},
x grid style={darkgrey176},
xmin=2.75, xmax=52.25,
xtick style={color=black},
y grid style={darkgrey176},
ymin=0.281119969642561, ymax=0.883002835119948,
ytick style={color=black}
]
\addplot [line width=1.5pt, opacity=0.75, black]
table {%
	5 0.308478281709715
	10 0.401715159395953
	15 0.477653904044375
	20 0.544599592535146
	30 0.65418202897815
	40 0.742915722579163
	50 0.804755510557163
};
\addplot [line width=1.5pt, opacity=0.75, slategrey119119153]
table {%
	5 0.383220136030017
	10 0.484012845112067
	15 0.559517140841141
	20 0.622321586469698
	30 0.718897261345499
	40 0.790045581755507
	50 0.837991962904301
};
\addplot [line width=1.5pt, opacity=0.75, orangered2381020]
table {%
	5 0.447803317481264
	10 0.551339842566798
	15 0.624883035180853
	20 0.679937471482302
	30 0.760447372420299
	40 0.81775855280608
	50 0.855644523052794
};

\nextgroupplot[
tick align=outside,
tick pos=left,
ylabel={\bf \scriptsize $F_1$ Score},
x grid style={darkgrey176},
xmin=2.75, xmax=52.25,
xtick style={color=black},
y grid style={darkgrey176},
ymin=0.391739862868378, ymax=0.791172235504142,
ytick style={color=black}
]
\addplot [line width=1.5pt, opacity=0.75, black]
table {%
5 0.409895879806367
10 0.502441517552384
15 0.574980981694514
20 0.632837188343371
30 0.70836338131792
40 0.743791581610878
50 0.734590289302567
};
\addplot [line width=1.5pt, opacity=0.75, slategrey119119153]
table {%
5 0.478507661820429
10 0.57499822189929
15 0.641441377026233
20 0.68949597898146
30 0.745036733173561
40 0.762059783457733
50 0.741951005372204
};
\addplot [line width=1.5pt, opacity=0.75, orangered2381020]
table {%
5 0.535205507769558
10 0.627868690056753
15 0.687973899400165
20 0.726783587724413
30 0.766666370745434
40 0.773016218566152
50 0.74717966164022
};

\nextgroupplot[
tick align=outside,
tick pos=left,
ylabel={\bf \scriptsize $F_2$ Score},
x grid style={darkgrey176},
xmin=2.75, xmax=52.25,
xtick style={color=black},
y grid style={darkgrey176},
ymin=0.520752546667141, ymax=0.800412130236723,
ytick style={color=black},
xtick={10,30,50},
xticklabels={0.1,0.3,0.5},
xlabel={\bf Confidence Threshold},
]
\addplot [line width=1.5pt, opacity=0.75, black]
table {%
5 0.533464345920304
10 0.620963230921971
15 0.687201826149562
20 0.73163554423834
30 0.771459978787054
40 0.766617892650551
50 0.721893664640764
};
\addlegendentry{\bf \scriptsize AL \cite{griff25AL}}
\addplot [line width=1.5pt, opacity=0.75, slategrey119119153]
table {%
5 0.587147774176866
10 0.674018361157897
15 0.728352706478949
20 0.760195278779139
30 0.78128534913202
40 0.764988611191193
50 0.716061396512619
};
\addlegendentry{\bf  \scriptsize TTN}
\addplot [line width=1.5pt, opacity=0.75, orangered2381020]
table {%
5 0.629229445847001
10 0.707066133360821
15 0.753713014299086
20 0.777290685286163
30 0.78770033098356
40 0.766155591585184
50 0.715881321467151
};
\addlegendentry{\bf  \scriptsize TTN$_{\text{\tiny D}}$}
\end{groupplot}

\end{tikzpicture}
	\end{minipage}%
	\begin{minipage}{0.241\textwidth}
		\begin{tikzpicture}
	
	\definecolor{darkgrey176}{RGB}{176,176,176}
	\definecolor{gainsboro247}{RGB}{253,253,253}
	\definecolor{lightgrey204}{RGB}{204,204,204}
	
	\definecolor{lightgrey204}{RGB}{204,204,204}
	\definecolor{orangered2381020}{RGB}{238,102,0}
	\definecolor{slategrey119119153}{RGB}{119,119,153}
	
	\tikzstyle{every node}=[
	font=\scriptsize,
	]
	
	\begin{groupplot}[
		group style={
			group size=1 by 5,
			vertical sep=0.3cm,
		},
		legend cell align={left},
		legend style={
			fill opacity=0.4,
			draw opacity=1,
			text opacity=1,
			at={(0.32,0.096)},
			anchor=south west,
			draw=none,
			fill=gainsboro247
		},
		width=3.9cm,
		height=3.9cm,
		ytick align=inside,
		grid=both,
		x grid style={darkgrey176},
		y grid style={darkgrey176},
		title style={at={(0.5,1.125)},anchor=north},
		xtick style={draw=none},
		ytick style={draw=none},
		xminorticks=true,
		minor x tick num=1,
		yminorticks=true,
		minor y tick num=1,
		axis background/.style={fill=gainsboro247},
		yticklabel style = {xshift=1.8ex},
		xticklabel style = {yshift=1.2ex},
		y label style={at={(axis description cs:0.18, 0.5)}},
	x label style={at={(axis description cs:1.18, 0.1)}},
		xticklabels={},
		yticklabel style={/pgf/number format/fixed zerofill,/pgf/number format/fixed},
		]
\nextgroupplot[
title={\bf \footnotesize COCO},
legend cell align={left},
legend style={fill opacity=0.8, draw opacity=1, text opacity=1, draw=lightgrey204},
log basis y={10},
tick align=outside,
tick pos=left,
x grid style={darkgrey176},
xmin=2.75, xmax=52.25,
xtick style={color=black},
y grid style={darkgrey176},
ymin=401799.327240942, ymax=5344251.58323348,
ymode=log,
ytick style={color=black},
ytick={500000,1000000,2000000,4000000},
yticklabels={
	\(\displaystyle $\text{\tiny 0.5}\textrm{\tiny M}$\),
	\(\displaystyle $\text{\tiny 1}\textrm{\tiny M}$\),
	\(\displaystyle $\text{\tiny 2}\textrm{\tiny M}$\),
	\(\displaystyle $\text{\tiny 4}\textrm{\tiny M}$\),
}
]
\addplot [line width=1.5pt, opacity=0.75, black]
table {%
5 4751180.66666667
10 2753161.66666667
15 1868628.33333333
20 1397346.33333333
30 907936
40 658433
50 508047
};
\addplot [line width=1.5pt, opacity=0.75, slategrey119119153]
table {%
5 2739054.66666667
10 1649410
15 1226135
20 991520.666666667
30 723575.666666667
40 565028
50 451954.333333333
};
\addplot [line width=1.5pt, opacity=0.75, orangered2381020]
table {%
5 3835078.66666667
10 2303608.33333333
15 1611997.33333333
20 1238841.66666667
30 844034
40 630124.333333333
50 491330
};

\nextgroupplot[
tick align=outside,
tick pos=left,
x grid style={darkgrey176},
xmin=2.75, xmax=52.25,
xtick style={color=black},
y grid style={darkgrey176},
ymin=0.422963312731985, ymax=0.658205632964975,
ytick style={color=black}
]
\addplot [line width=1.5pt, opacity=0.75, black]
table {%
5 0.635356588245282
10 0.642885564418261
15 0.647512800227112
20 0.636417346923011
30 0.593239326425953
40 0.531835699925165
50 0.461330424763019
};
\addplot [line width=1.5pt, opacity=0.75, slategrey119119153]
table {%
5 0.565667489514488
10 0.574621461053851
15 0.581954504568301
20 0.574934818495491
30 0.542386655198447
40 0.493107800017975
50 0.433656145469848
};
\addplot [line width=1.5pt, opacity=0.75, orangered2381020]
table {%
5 0.620832034671137
10 0.62917340454989
15 0.634324955576416
20 0.624048179765773
30 0.583043393829112
40 0.524148256077392
50 0.455907954221114
};

\nextgroupplot[
tick align=outside,
tick pos=left,
x grid style={darkgrey176},
xmin=2.75, xmax=52.25,
xtick style={color=black},
y grid style={darkgrey176},
ymin=0.221940798392137, ymax=0.823260172661399,
ytick style={color=black}
]
\addplot [line width=1.5pt, opacity=0.75, black]
table {%
	5 0.249273497222558
	10 0.344259171060757
	15 0.424561024583793
	20 0.493334193186005
	30 0.610695780641732
	40 0.705625509996373
	50 0.771230471615822
};
\addplot [line width=1.5pt, opacity=0.75, slategrey119119153]
table {%
	5 0.319263113376123
	10 0.419718610584217
	15 0.497441824514894
	20 0.56030846590189
	30 0.662736703520897
	40 0.742869214090883
	50 0.795927473830978
};
\addplot [line width=1.5pt, opacity=0.75, orangered2381020]
table {%
	5 0.279413374537117
	10 0.375167680220571
	15 0.453934563951612
	20 0.520121617964612
	30 0.631310943797547
	40 0.720792567202972
	50 0.78147511286005
};

\nextgroupplot[
tick align=outside,
tick pos=left,
x grid style={darkgrey176},
xmin=2.75, xmax=52.25,
xtick style={color=black},
y grid style={darkgrey176},
ymin=0.322801444913054, ymax=0.601426700503873,
ytick style={color=black}
]
\addplot [line width=1.5pt, opacity=0.75, black]
table {%
5 0.335466229258091
10 0.421382797981297
15 0.485656891340509
20 0.531491059792265
30 0.583274395827735
40 0.587262350148943
50 0.551934551309662
};
\addplot [line width=1.5pt, opacity=0.75, slategrey119119153]
table {%
5 0.381844158394203
10 0.45938820442955
15 0.512605370037891
20 0.54690178630717
30 0.580043287066763
40 0.574430756007729
50 0.537475850204598
};
\addplot [line width=1.5pt, opacity=0.75, orangered2381020]
table {%
5 0.360771394429856
10 0.443232354354453
15 0.503423744102309
20 0.54483253282813
30 0.588761916158836
40 0.588031523414518
50 0.551045818656997
};

\nextgroupplot[
tick align=outside,
tick pos=left,
x grid style={darkgrey176},
xmin=2.75, xmax=52.25,
xtick style={color=black},
y grid style={darkgrey176},
ymin=0.436059366241781, ymax=0.592203082030918,
ytick style={color=black},
xtick={10,30,50},
xticklabels={0.1,0.3,0.5},
xlabel={\textcolor{white}{|}},
]
\addplot [line width=1.5pt, opacity=0.75, black]
table {%
5 0.44315680786856
10 0.510385539477041
15 0.55619661906265
20 0.579937235834354
30 0.585105640404139
40 0.550285481464968
50 0.491639837736022
};
\addplot [line width=1.5pt, opacity=0.75, slategrey119119153]
table {%
5 0.453098245571368
10 0.507523260168719
15 0.541964306029817
20 0.556833720942534
30 0.553772271546379
40 0.520604600293124
50 0.468075385041828
};
\addplot [line width=1.5pt, opacity=0.75, orangered2381020]
table {%
5 0.458322385358838
10 0.520265228166552
15 0.561627719901501
20 0.581507406379743
30 0.581827895997715
40 0.545780717214233
50 0.487775633894772
};
\end{groupplot}

\end{tikzpicture}
	\end{minipage}%
	\begin{minipage}{0.241\textwidth}
		\begin{tikzpicture}
	
	\definecolor{darkgrey176}{RGB}{176,176,176}
	\definecolor{gainsboro247}{RGB}{253,253,253}
	\definecolor{lightgrey204}{RGB}{204,204,204}
	
	\definecolor{lightgrey204}{RGB}{204,204,204}
	\definecolor{orangered2381020}{RGB}{238,102,0}
	\definecolor{slategrey119119153}{RGB}{119,119,153}
	
	\tikzstyle{every node}=[
	font=\scriptsize,
	]
	
	\begin{groupplot}[
		group style={
			group size=1 by 5,
			vertical sep=0.3cm,
		},
		legend cell align={left},
		legend style={
			fill opacity=0.4,
			draw opacity=1,
			text opacity=1,
			at={(0.32,0.096)},
			anchor=south west,
			draw=none,
			fill=gainsboro247
		},
		width=3.9cm,
		height=3.9cm,
		ytick align=inside,
		grid=both,
		x grid style={darkgrey176},
		y grid style={darkgrey176},
		title style={at={(0.5,1.125)},anchor=north},
		xtick style={draw=none},
		ytick style={draw=none},
		xminorticks=true,
		minor x tick num=1,
		yminorticks=true,
		minor y tick num=1,
		axis background/.style={fill=gainsboro247},
		yticklabel style = {xshift=1.8ex},
		xticklabel style = {yshift=1.2ex},
		y label style={at={(axis description cs:0.18, 0.5)}},
		xticklabels={},
		scaled ticks=false,
		yticklabel style={/pgf/number format/fixed},
		x label style={at={(axis description cs:0.85, 0.1)}},
		yticklabel style={/pgf/number format/fixed zerofill,/pgf/number format/fixed},
		]
\nextgroupplot[
title={\bf \footnotesize LVIS},
legend cell align={left},
legend style={fill opacity=0.8, draw opacity=1, text opacity=1, draw=lightgrey204},
log basis y={10},
tick align=outside,
tick pos=left,
x grid style={darkgrey176},
xmin=2.75, xmax=52.25,
xtick style={color=black},
y grid style={darkgrey176},
ymin=451580.68140199, ymax=4581860.98070565,
ymode=log,
ytick style={color=black},
ytick={500000,1000000,2000000,4000000},
yticklabels={
	\(\displaystyle $\text{\tiny 0.5}\textrm{\tiny M}$\),
	\(\displaystyle $\text{\tiny 1}\textrm{\tiny M}$\),
	\(\displaystyle $\text{\tiny 2}\textrm{\tiny M}$\),
	\(\displaystyle $\text{\tiny 4}\textrm{\tiny M}$\),
}
]
\addplot [line width=1.5pt, opacity=0.75, black]
table {%
5 4123829.5
10 2569288.5
15 1900988.5
20 1506647
30 1038225.5
40 751933.5
50 548443.5
};
\addplot [line width=1.5pt, opacity=0.75, slategrey119119153]
table {%
5 3389887.5
10 2156508.5
15 1620347
20 1301008
30 916279.5
40 675948
50 501737.5
};
\addplot [line width=1.5pt, opacity=0.75, orangered2381020]
table {%
5 3782298.5
10 2390816.5
15 1783372
20 1421911.5
30 988950.5
40 721340
50 529018.5
};

\nextgroupplot[
tick align=outside,
tick pos=left,
x grid style={darkgrey176},
xmin=2.75, xmax=52.25,
xtick style={color=black},
y grid style={darkgrey176},
ymin=0.0738997463526117, ymax=0.201501510213041,
ytick style={color=black},
]
\addplot [line width=1.5pt, opacity=0.75, black]
table {%
5 0.195701430037567
10 0.170687470367729
15 0.153515597485003
20 0.139836660079715
30 0.118285698583887
40 0.0999119086586565
50 0.0824574062113158
};
\addplot [line width=1.5pt, opacity=0.75, slategrey119119153]
table {%
5 0.185469780656589
10 0.162444582214198
15 0.146477841157265
20 0.133679892713457
30 0.113574273100944
40 0.0961434507434492
50 0.0796998265280858
};
\addplot [line width=1.5pt, opacity=0.75, orangered2381020]
table {%
5 0.192920107053257
10 0.168680669347415
15 0.15179911196413
20 0.138327545531844
30 0.117082338959855
40 0.0990093932708338
50 0.0817769354600326
};

\nextgroupplot[
tick align=outside,
tick pos=left,
x grid style={darkgrey176},
xmin=2.75, xmax=52.25,
xtick style={color=black},
y grid style={darkgrey176},
ymin=0.0628341886990102, ymax=0.159363489524729,
ytick style={color=black},
ytick={0.05,0.1,0.15},
]
\addplot [line width=1.5pt, opacity=0.75, black]
table {%
5 0.0672218841910883
10 0.0863052044769047
15 0.099028844662963
20 0.109844553543732
30 0.12773057322818
40 0.14176756543504
50 0.150700543316964
};
\addplot [line width=1.5pt, opacity=0.75, slategrey119119153]
table {%
5 0.0726061806929535
10 0.0914752357885806
15 0.103889462081336
20 0.114385434088566
30 0.131636407336446
40 0.145111238802914
50 0.153582871115765
};
\addplot [line width=1.5pt, opacity=0.75, orangered2381020]
table {%
5 0.0714448610708873
10 0.0909270072021103
15 0.10352254570491
20 0.114133277603263
30 0.132227003947736
40 0.146423062493082
50 0.154975794032651
};

\nextgroupplot[
tick align=outside,
tick pos=left,
x grid style={darkgrey176},
xmin=2.75, xmax=52.25,
xtick style={color=black},
y grid style={darkgrey176},
ymin=0.0695161966134031, ymax=0.0854146288779393,
ytick style={color=black},
ytick={0.07,0.08,0.09},
]
\addplot [line width=1.5pt, opacity=0.75, black]
table {%
5 0.0702901570581985
10 0.0785421683665559
15 0.0817810477239672
20 0.0831343810491203
30 0.0817466648787263
40 0.0773639211217152
50 0.070555590854212
};
\addplot [line width=1.5pt, opacity=0.75, slategrey119119153]
table {%
5 0.0731510023426095
10 0.080360506432365
15 0.0829502345323804
20 0.0838515788175172
30 0.0819583776852576
40 0.0772244232478479
50 0.0702388526254274
};
\addplot [line width=1.5pt, opacity=0.75, orangered2381020]
table {%
5 0.0731350189493137
10 0.0808586428669497
15 0.0836897822769527
20 0.084691972865915
30 0.0827800151505375
40 0.0780386013016942
50 0.0709827921137429
};

\nextgroupplot[
tick align=outside,
tick pos=left,
x grid style={darkgrey176},
xmin=2.75, xmax=52.25,
xtick style={color=black},
y grid style={darkgrey176},
ymin=0.0650495239684029, ymax=0.0984054982636198,
ytick style={color=black},
xtick={10,30,50},
xticklabels={0.1,0.3,0.5},
xlabel={\textcolor{white}{|}},
]
\addplot [line width=1.5pt, opacity=0.75, black]
table {%
5 0.0920028215605468
10 0.0953220114931597
15 0.0945818808386982
20 0.0925251394471933
30 0.0858200457509671
40 0.0775273282001192
50 0.0677439107838158
};
\addplot [line width=1.5pt, opacity=0.75, slategrey119119153]
table {%
5 0.0922408063691621
10 0.0945918330661807
15 0.0934266336102766
20 0.0911771483426294
30 0.0844476375419971
40 0.0761736622283246
50 0.0665657046181855
};
\addplot [line width=1.5pt, opacity=0.75, orangered2381020]
table {%
5 0.0942426781862895
10 0.0968893176138372
15 0.0957040880041277
20 0.0933092288312391
30 0.0861920171244091
40 0.0776747458281286
50 0.0677664309169981
};
\end{groupplot}

\end{tikzpicture}
	\end{minipage}%
	\begin{minipage}{0.2425\textwidth}
		\begin{tikzpicture}
	
	\definecolor{darkgrey176}{RGB}{176,176,176}
	\definecolor{gainsboro247}{RGB}{253,253,253}
	\definecolor{lightgrey204}{RGB}{204,204,204}
	
	\definecolor{lightgrey204}{RGB}{204,204,204}
	\definecolor{orangered2381020}{RGB}{238,102,0}
	\definecolor{slategrey119119153}{RGB}{119,119,153}
	
	\tikzstyle{every node}=[
	font=\scriptsize,
	]
	
	\begin{groupplot}[
		group style={
			group size=1 by 5,
			vertical sep=0.3cm,
		},
		legend cell align={left},
		legend style={
			fill opacity=0.4,
			draw opacity=1,
			text opacity=1,
			at={(0.32,0.096)},
			anchor=south west,
			draw=none,
			fill=gainsboro247
		},
		width=3.9cm,
		height=3.9cm,
		ytick align=inside,
		grid=both,
		x grid style={darkgrey176},
		y grid style={darkgrey176},
		title style={at={(0.5,1.125)},anchor=north},
		xtick style={draw=none},
		ytick style={draw=none},
		xminorticks=true,
		minor x tick num=1,
		yminorticks=true,
		minor y tick num=1,
		axis background/.style={fill=gainsboro247},
		yticklabel style = {xshift=1.8ex},
		xticklabel style = {yshift=1.2ex},
		y label style={at={(axis description cs:0.18, 0.5)}},
		xticklabels={},
		scaled ticks=false,
		yticklabel style={/pgf/number format/fixed},
		x label style={at={(axis description cs:0.6, 0.1)}},
		yticklabel style={/pgf/number format/fixed zerofill,/pgf/number format/fixed},
		]
\nextgroupplot[
title={\bf \footnotesize BDD},
legend cell align={left},
legend style={fill opacity=0.8, draw opacity=1, text opacity=1, draw=lightgrey204},
log basis y={10},
tick align=outside,
tick pos=left,
x grid style={darkgrey176},
xmin=2.75, xmax=52.25,
xtick style={color=black},
y grid style={darkgrey176},
ymin=132352.116833333, ymax=3794844.47045679,
ymode=log,
ytick style={color=black},
ytick={200000,400000,800000,1600000,3200000},
yticklabels={
	\(\displaystyle $\text{\tiny 0.2}\textrm{\tiny M}$\),
	\(\displaystyle $\text{\tiny 0.4}\textrm{\tiny M}$\),
	\(\displaystyle $\text{\tiny 0.8}\textrm{\tiny M}$\),
	\(\displaystyle $\text{\tiny 1.6}\textrm{\tiny M}$\),
	\(\displaystyle $\text{\tiny 3.2}\textrm{\tiny M}$\),
}
]
\addplot [line width=1.5pt, opacity=0.75, black]
table {%
5 3257959.33333333
10 1647305
15 1040828.33333333
20 744152
30 447921.666666667
40 280278.666666667
50 169949.333333333
};
\addplot [line width=1.5pt, opacity=0.75, slategrey119119153]
table {%
5 2448069.33333333
10 1249964
15 835564.333333333
20 625700.333333333
30 400472.666666667
40 259886.666666667
50 161095.333333333
};
\addplot [line width=1.5pt, opacity=0.75, orangered2381020]
table {%
5 2068677
10 1170038.66666667
15 794611
20 596697.666666667
30 383014.666666667
40 249083.333333333
50 154162.666666667
};

\nextgroupplot[
tick align=outside,
tick pos=left,
x grid style={darkgrey176},
xmin=2.75, xmax=52.25,
xtick style={color=black},
y grid style={darkgrey176},
ymin=0.127511975010996, ymax=0.341600359099689,
ytick style={color=black},
ytick={0.1,0.2,0.3},
]
\addplot [line width=1.5pt, opacity=0.75, black]
table {%
5 0.331869068913839
10 0.325827369139732
15 0.31420122674707
20 0.295849825718143
30 0.253971167889281
40 0.199036879660478
50 0.145697280672686
};
\addplot [line width=1.5pt, opacity=0.75, slategrey119119153]
table {%
5 0.287469483742212
10 0.287249047553848
15 0.282253217312419
20 0.269318001479323
30 0.236711929057249
40 0.189195207814013
50 0.140605552529473
};
\addplot [line width=1.5pt, opacity=0.75, orangered2381020]
table {%
5 0.296502257984635
10 0.292595267506149
15 0.284082691157959
20 0.268821355714162
30 0.233462175636789
40 0.185295058559528
50 0.137243265196846
};

\nextgroupplot[
tick align=outside,
tick pos=left,
x grid style={darkgrey176},
xmin=2.75, xmax=52.25,
xtick style={color=black},
y grid style={darkgrey176},
ymin=0.230060005748329, ymax=0.632349448945773,
ytick style={color=black},
ytick={0.2,0.4,0.6},
]
\addplot [line width=1.5pt, opacity=0.75, black]
table {%
5 0.248345889530031
10 0.287631862608733
15 0.343279657136359
20 0.391738767903575
30 0.476058997579072
40 0.53977039686177
50 0.588577746135815
};
\addplot [line width=1.5pt, opacity=0.75, slategrey119119153]
table {%
5 0.275155127248203
10 0.311733064032291
15 0.365364279000334
20 0.411253705731723
30 0.490845036666317
40 0.54873733637362
50 0.593651581209817
};
\addplot [line width=1.5pt, opacity=0.75, orangered2381020]
table {%
5 0.303625568870583
10 0.345314394895599
15 0.400933210492762
20 0.447933256418856
30 0.526726833742732
40 0.572953807389279
50 0.614063565164071
};

\nextgroupplot[
tick align=outside,
tick pos=left,
x grid style={darkgrey176},
xmin=2.75, xmax=52.25,
xtick style={color=black},
y grid style={darkgrey176},
ymin=0.206907306168905, ymax=0.293749646911837,
ytick style={color=black},
ytick={0.22,0.25,0.28},
]
\addplot [line width=1.5pt, opacity=0.75, black]
table {%
5 0.217016107425892
10 0.25035590606527
15 0.270546029033133
20 0.28273964557573
30 0.289699380508965
40 0.264062583179691
50 0.216612941339701
};
\addplot [line width=1.5pt, opacity=0.75, slategrey119119153]
table {%
5 0.213296256713309
10 0.243172243383991
15 0.262417502473346
20 0.274314903056493
30 0.280508602505685
40 0.255607331408391
50 0.210854685293584
};
\addplot [line width=1.5pt, opacity=0.75, orangered2381020]
table {%
5 0.237382264010345
10 0.263269619461141
15 0.280167139731645
20 0.289319752812234
30 0.289802267787159
40 0.259784352338425
50 0.21091195377437
};

\nextgroupplot[
tick align=outside,
tick pos=left,
x grid style={darkgrey176},
xmin=2.75, xmax=52.25,
xtick style={color=black},
y grid style={darkgrey176},
ymin=0.153503446849623, ymax=0.277773338177534,
ytick style={color=black},
xtick={10,30,50},
xticklabels={0.1,0.3,0.5},
xlabel={\textcolor{white}{|}},
]
\addplot [line width=1.5pt, opacity=0.75, black]
table {%
5 0.241799902658914
10 0.260836469386804
15 0.270233169161147
20 0.272124706753538
30 0.259018501239182
40 0.218083531991649
50 0.166864207387926
};
\addplot [line width=1.5pt, opacity=0.75, slategrey119119153]
table {%
5 0.222945775669122
10 0.242375596970091
15 0.253076901885992
20 0.256033962647345
30 0.245281083682174
40 0.208606359776321
50 0.161477238339856
};
\addplot [line width=1.5pt, opacity=0.75, orangered2381020]
table {%
5 0.2441455519881
10 0.257761097517061
15 0.264577922720064
20 0.264219682601111
30 0.248961316889441
40 0.208293232273892
50 0.159152078273619
};
\end{groupplot}

\end{tikzpicture}
	\end{minipage}%
	\vspace{-2.1em}
	\caption{{\bf Label Evaluation across all Confidence Thresholds and Datasets}.
		Results averaged over VLM model \& class, but Number of Labels is total for all classes.
	}
	\label{fig:metric}
\end{figure}

We provide an extended suite of quantitative results supporting the efficacy of the Label Imitation Game (LIG) and corresponding TTN model. 
Relative to the experimental evaluation in \cref{sec:exp}, the following figures offer a more granular breakdown of performance across all four AL benchmark \cite{griff25AL} datasets (VOC, COCO, LVIS, and BDD).
For pseudo-label evaluation, we plot the number of labels, recall, precision, $F_1$ score, and $F_2$ score across all datasets in \cref{fig:metric}.
For downstream model evaluation, we plot the mAP$_{50}$ results corresponding to each individual VLM on VOC in \cref{fig:vocmap}.
For breadth, we plot the mAP$_{50}$ and mAP$_{50:95}$ results across all datasets in \cref{fig:map}.
Finally, we plot the mAP$_{50}$ and mAP$_{50:95}$ results for the transfer-vulnerable classes identified in \cref{sec:vulnerable} of the main paper across all datasets in \cref{fig:mapv}.

\begin{figure}[t!]
	\begin{minipage}{1\textwidth}
		\begin{tikzpicture}
	
	\definecolor{darkgrey176}{RGB}{176,176,176}
	\definecolor{gainsboro247}{RGB}{253,253,253}
	\definecolor{lightgrey204}{RGB}{204,204,204}
	
	\definecolor{lightgrey204}{RGB}{204,204,204}
	\definecolor{orangered2381020}{RGB}{238,102,0}
	\definecolor{slategrey119119153}{RGB}{119,119,153}
	
	\tikzstyle{every node}=[
	font=\scriptsize,
	]
	
	\begin{groupplot}[
		group style={
			group size=4 by 1,
			horizontal sep=0.66cm,
		},
		legend cell align={left},
		legend style={
			fill opacity=0.4,
			draw opacity=1,
			text opacity=1,
			at={(0.16,0.04)},
			anchor=south west,
			draw=none,
			fill=gainsboro247,
			at={(2.58,-0.33)},
legend columns=3,
column sep = 1pt,
fill opacity=0,
yticklabel style={/pgf/number format/fixed zerofill,/pgf/number format/fixed},
},
		width=3.9cm,
		height=3.9cm,
		ytick align=inside,
		grid=both,
		x grid style={darkgrey176},
		y grid style={darkgrey176},
		title style={at={(0.5,1.125)},anchor=north},
		xtick style={draw=none},
		ytick style={draw=none},
		xminorticks=true,
		minor x tick num=1,
		yminorticks=true,
		minor y tick num=1,
		axis background/.style={fill=gainsboro247},
		yticklabel style = {xshift=1.8ex},
		xticklabel style = {yshift=1.2ex},
		y label style={at={(axis description cs:0.25, 0.5)}},
		xticklabels={},
x label style={at={(axis description cs:-0.17, 0.1)}},
		]
\nextgroupplot[
tick align=outside,
tick pos=left,
title={\bf \footnotesize GDINO},
x grid style={darkgrey176},
xmin=0.0275, xmax=0.5225,
xtick style={color=black},
y grid style={darkgrey176},
ylabel={\bf mAP$_{50}$},
ymin=0.169526253757373, ymax=0.732123046624858,
ytick style={color=black},
xtick={0.1,0.3,0.5},
xticklabels={0.1,0.3,0.5},
]
\addplot [line width=1.5pt, opacity=0.75,  black]
table {%
0.05 0.195098835251349
0.1 0.481442762647939
0.15 0.589432604929476
0.2 0.637121408992923
0.3 0.675880822160058
0.4 0.690645309859582
0.5 0.673985946089972
};
\addlegendentry{\bf  \scriptsize AL \cite{griff25AL}}
\addplot [line width=1.5pt, opacity=0.75,  slategrey119119153]
table {%
0.05 0.292871627189704
0.1 0.540933008716186
0.15 0.641995030044866
0.2 0.672579641273079
0.3 0.692590850171025
0.4 0.695800960530937
0.5 0.672212924469775
};
\addlegendentry{\bf  \scriptsize TTN}
\addplot [line width=1.5pt, opacity=0.75,  orangered2381020]
table {%
0.05 0.35026279583503
0.1 0.578735095448923
0.15 0.66744544833006
0.2 0.69095827981893
0.3 0.703394788725886
0.4 0.706550465130881
0.5 0.677804093608749
};
\addlegendentry{\bf  \scriptsize TTN$_{\text{\tiny D}}$}

\nextgroupplot[
tick align=outside,
tick pos=left,
title={\bf \footnotesize YOLOE},
x grid style={darkgrey176},
xmin=0.0275, xmax=0.5225,
xtick style={color=black},
y grid style={darkgrey176},
ymin=0.680184459651142, ymax=0.72257167414642,
ytick style={color=black},
xtick={0.1,0.3,0.5},
xticklabels={0.1,0.3,0.5},
xlabel={\bf Confidence Threshold},
]
\addplot [line width=1.5pt, opacity=0.75,  black]
table {%
0.05 0.688615279968037
0.1 0.705987967917236
0.15 0.701697444279976
0.2 0.708910488535897
0.3 0.70783178237348
0.4 0.702562289319267
0.5 0.682111151219109
};
\addplot [line width=1.5pt, opacity=0.75,  slategrey119119153]
table {%
0.05 0.706127316751302
0.1 0.715948848113847
0.15 0.71639422236684
0.2 0.714020867359599
0.3 0.715335365357623
0.4 0.699219932251479
0.5 0.689555368707609
};
\addplot [line width=1.5pt, opacity=0.75,  orangered2381020]
table {%
0.05 0.715030566650751
0.1 0.718656020173769
0.15 0.71996411994598
0.2 0.715520465197343
0.3 0.720644982578452
0.4 0.703750178815102
0.5 0.70158778318898
};

\nextgroupplot[
tick align=outside,
tick pos=left,
title={\bf \footnotesize YOLOW},
x grid style={darkgrey176},
xmin=0.0275, xmax=0.5225,
xtick style={color=black},
y grid style={darkgrey176},
ymin=0.678484788539812, ymax=0.731561672681268,
ytick style={color=black},
xtick={0.1,0.3,0.5},
xticklabels={0.1,0.3,0.5},
]
\addplot [line width=1.5pt, opacity=0.75,  black]
table {%
0.05 0.700732479952618
0.1 0.716887370116343
0.15 0.718341715363312
0.2 0.714931859211686
0.3 0.708052002048271
0.4 0.697266071322653
0.5 0.680897374182605
};
\addplot [line width=1.5pt, opacity=0.75,  slategrey119119153]
table {%
0.05 0.720292902012944
0.1 0.723734202002589
0.15 0.723495093475001
0.2 0.725793693930506
0.3 0.716117546574298
0.4 0.707277307895822
0.5 0.684969202855828
};
\addplot [line width=1.5pt, opacity=0.75,  orangered2381020]
table {%
0.05 0.719591507070065
0.1 0.727622810958365
0.15 0.729149087038475
0.2 0.727475675162124
0.3 0.711895408045423
0.4 0.708594018239804
0.5 0.68511150670328
};

\nextgroupplot[
title={\bf \footnotesize VOC Overall},
tick align=outside,
tick pos=left,
x grid style={darkgrey176},
xmin=0.0275, xmax=0.5225,
xtick style={color=black},
y grid style={darkgrey176},
ymin=0.518957388654372, ymax=0.72116986951955,
ytick style={color=black},
xtick={0.1,0.3,0.5},
xticklabels={0.1,0.3,0.5},
]
\addplot [line width=1.5pt, opacity=0.75,  black]
table {%
	0.05 0.528148865057335
	0.1 0.634772700227172
	0.15 0.669823921524255
	0.2 0.686987918913502
	0.3 0.697254868860603
	0.4 0.696824556833834
	0.5 0.678998157163896
};
\addplot [line width=1.5pt, opacity=0.75,  slategrey119119153]
table {%
	0.05 0.57309728198465
	0.1 0.660205352944207
	0.15 0.693961448628902
	0.2 0.704131400854395
	0.3 0.708014587367649
	0.4 0.700766066892746
	0.5 0.68224583201107
};
\addplot [line width=1.5pt, opacity=0.75,  orangered2381020]
table {%
	0.05 0.594961623185282
	0.1 0.675004642193686
	0.15 0.705519551771505
	0.2 0.711318140059466
	0.3 0.711978393116587
	0.4 0.706298220728596
	0.5 0.688167794500336
};
\end{groupplot}

\end{tikzpicture}
	\end{minipage}%
	\vspace{-2.1em}
	\caption{{\bf Downstream Model Evaluation across VLM Models and Confidence Thresholds on VOC}. 
		VOC Overall denotes the average over all VLM models.
	}
	\label{fig:vocmap}
\end{figure}
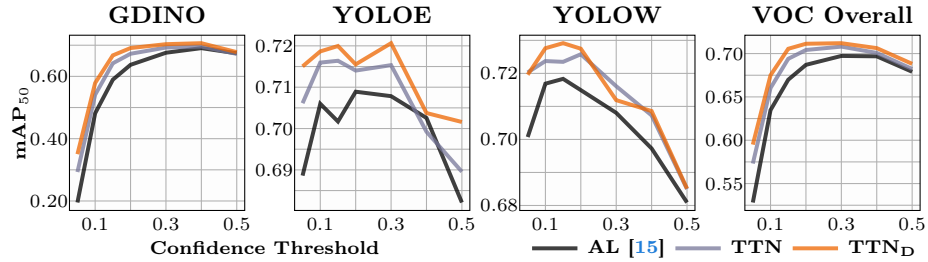

\begin{figure}[t!]
	\begin{minipage}{0.264\textwidth}
		\begin{tikzpicture}
	
	\definecolor{darkgrey176}{RGB}{176,176,176}
	\definecolor{gainsboro247}{RGB}{253,253,253}
	\definecolor{lightgrey204}{RGB}{204,204,204}
	
	\definecolor{lightgrey204}{RGB}{204,204,204}
	\definecolor{orangered2381020}{RGB}{238,102,0}
	\definecolor{slategrey119119153}{RGB}{119,119,153}
	
	\tikzstyle{every node}=[
	font=\scriptsize,
	]
	
	\begin{groupplot}[
		group style={
			group size=1 by 5,
			vertical sep=0.3cm,
		},
		legend cell align={left},
		legend style={
			fill opacity=0.4,
			draw opacity=1,
			text opacity=1,
			at={(0.16,0.04)},
			anchor=south west,
			draw=none,
			fill=gainsboro247,
			at={(2.56,-0.33)},
			legend columns=3,
			column sep = 1pt,
			fill opacity=0,
		},
		width=3.9cm,
		height=3.9cm,
		ytick align=inside,
		grid=both,
		x grid style={darkgrey176},
		y grid style={darkgrey176},
		title style={at={(0.5,1.125)},anchor=north},
		xtick style={draw=none},
		ytick style={draw=none},
		xminorticks=true,
		minor x tick num=1,
		yminorticks=true,
		minor y tick num=1,
		axis background/.style={fill=gainsboro247},
		yticklabel style = {xshift=1.8ex},
		xticklabel style = {yshift=1.2ex},
		y label style={at={(axis description cs:0.23, 0.5)}},
		xticklabels={},
		x label style={at={(axis description cs:1.1, 0.1)}},
		yticklabel style={/pgf/number format/fixed zerofill,/pgf/number format/fixed},
]
\nextgroupplot[
title={\bf \footnotesize VOC},
tick align=outside,
tick pos=left,
x grid style={darkgrey176},
xmin=0.0275, xmax=0.5225,
xtick style={color=black},
y grid style={darkgrey176},
ylabel={\bf mAP$_{50}$},
ymin=0.518957388654372, ymax=0.72116986951955,
ytick style={color=black}
]
\addplot [line width=1.5pt, opacity=0.75,  black]
table {%
0.05 0.528148865057335
0.1 0.634772700227172
0.15 0.669823921524255
0.2 0.686987918913502
0.3 0.697254868860603
0.4 0.696824556833834
0.5 0.678998157163896
};
\addplot [line width=1.5pt, opacity=0.75,  slategrey119119153]
table {%
0.05 0.57309728198465
0.1 0.660205352944207
0.15 0.693961448628902
0.2 0.704131400854395
0.3 0.708014587367649
0.4 0.700766066892746
0.5 0.68224583201107
};
\addplot [line width=1.5pt, opacity=0.75,  orangered2381020]
table {%
0.05 0.594961623185282
0.1 0.675004642193686
0.15 0.705519551771505
0.2 0.711318140059466
0.3 0.711978393116587
0.4 0.706298220728596
0.5 0.688167794500336
};

\nextgroupplot[
legend cell align={left},
tick align=outside,
tick pos=left,
x grid style={darkgrey176},
xmin=0.0275, xmax=0.5225,
xtick style={color=black},
y grid style={darkgrey176},
ylabel={\bf mAP$_{50:95}$},
ymin=0.358648534972151, ymax=0.513961595415543,
ytick style={color=black},
xtick={0.1,0.3,0.5},
xticklabels={0.1,0.3,0.5},
xlabel={\bf Confidence Threshold},
]
\addplot [line width=1.5pt, opacity=0.75,  black]
table {%
0.05 0.36570821953776
0.1 0.447145520638628
0.15 0.476040820430036
0.2 0.490099048816664
0.3 0.499048754318168
0.4 0.496784146043497
0.5 0.483676687279642
};
\addlegendentry{\bf \scriptsize AL \cite{griff25AL}}
\addplot [line width=1.5pt, opacity=0.75,  slategrey119119153]
table {%
0.05 0.398482006450567
0.1 0.465681628470089
0.15 0.490927688107828
0.2 0.502007909955806
0.3 0.505323944071683
0.4 0.499321224614193
0.5 0.485199829621554
};
\addlegendentry{\bf  \scriptsize TTN}
\addplot [line width=1.5pt, opacity=0.75,  orangered2381020]
table {%
0.05 0.411931288149251
0.1 0.478231886022331
0.15 0.502618880692896
0.2 0.506901910849934
0.3 0.50682476663095
0.4 0.504851431124225
0.5 0.491747792421561
};
\addlegendentry{\bf  \scriptsize TTN$_{\text{\tiny D}}$}
\end{groupplot}

\end{tikzpicture}
	\end{minipage}%
	\begin{minipage}{0.241\textwidth}
		\begin{tikzpicture}
	
	\definecolor{darkgrey176}{RGB}{176,176,176}
	\definecolor{gainsboro247}{RGB}{253,253,253}
	\definecolor{lightgrey204}{RGB}{204,204,204}
	
	\definecolor{lightgrey204}{RGB}{204,204,204}
	\definecolor{orangered2381020}{RGB}{238,102,0}
	\definecolor{slategrey119119153}{RGB}{119,119,153}
	
	\tikzstyle{every node}=[
	font=\scriptsize,
	]
	
	\begin{groupplot}[
		group style={
			group size=1 by 5,
			vertical sep=0.3cm,
		},
		legend cell align={left},
		legend style={
			fill opacity=0.4,
			draw opacity=1,
			text opacity=1,
			at={(0.16,-0.01)},
			anchor=south west,
			draw=none,
			fill=gainsboro247
		},
		width=3.9cm,
		height=3.9cm,
		ytick align=inside,
		grid=both,
		x grid style={darkgrey176},
		y grid style={darkgrey176},
		title style={at={(0.5,1.125)},anchor=north},
		xtick style={draw=none},
		ytick style={draw=none},
		xminorticks=true,
		minor x tick num=1,
		yminorticks=true,
		minor y tick num=1,
		axis background/.style={fill=gainsboro247},
		yticklabel style = {xshift=1.8ex},
		xticklabel style = {yshift=1.2ex},
		y label style={at={(axis description cs:0.23, 0.5)}},
		xticklabels={},
		x label style={at={(axis description cs:1.18, 0.1)}},
		yticklabel style={/pgf/number format/fixed zerofill,/pgf/number format/fixed},
		]
\nextgroupplot[
tick align=outside,
tick pos=left,
title={\bf \footnotesize COCO},
x grid style={darkgrey176},
xmin=0.0275, xmax=0.5225,
xtick style={color=black},
y grid style={darkgrey176},
ymin=0.375569127189007, ymax=0.460381629376387,
ytick={0.37,0.4,0.43,0.46},
ytick style={color=black}
]
\addplot [line width=1.5pt, opacity=0.75,  black]
table {%
0.05 0.379424240924797
0.1 0.416872116444932
0.15 0.436347276625309
0.2 0.447333470410905
0.3 0.45362155714261
0.4 0.449659085928796
0.5 0.439557895498401
};
\addplot [line width=1.5pt, opacity=0.75,  slategrey119119153]
table {%
0.05 0.383439716228031
0.1 0.423979738750087
0.15 0.442558812681382
0.2 0.450267296995898
0.3 0.454287814507252
0.4 0.448085433761121
0.5 0.438146047893272
};
\addplot [line width=1.5pt, opacity=0.75,  orangered2381020]
table {%
	0.05 0.386686868595746
	0.1 0.424154564270664
	0.15 0.444001412487045
	0.2 0.451426032385978
	0.3 0.456526515640597
	0.4 0.449630838797904
	0.5 0.440907002404614
};

\nextgroupplot[
legend cell align={left},
legend style={
  fill opacity=0.8,
  draw opacity=1,
  text opacity=1,
  at={(0.97,0.03)},
  anchor=south east,
  draw=lightgrey204
},
tick align=outside,
tick pos=left,
x grid style={darkgrey176},
xmin=0.0275, xmax=0.5225,
xtick style={color=black},
y grid style={darkgrey176},
ymin=0.255551697124603, ymax=0.320737361212866,
ytick style={color=black},
xtick={0.1,0.3,0.5},
xticklabels={0.1,0.3,0.5},
xlabel={\textcolor{white}{|}},
]
\addplot [line width=1.5pt, opacity=0.75,  black]
table {%
0.05 0.258514681855887
0.1 0.286800066381551
0.15 0.302043524850332
0.2 0.309781028447359
0.3 0.315031716968078
0.4 0.312762833514798
0.5 0.307436412986164
};
\addplot [line width=1.5pt, opacity=0.75,  slategrey119119153]
table {%
0.05 0.263634348101263
0.1 0.293828278371659
0.15 0.308019640640086
0.2 0.313308560805992
0.3 0.317774376481581
0.4 0.313296020075204
0.5 0.307686552209071
};
\addplot [line width=1.5pt, opacity=0.75,  orangered2381020]
table {%
	0.05 0.263821478076136
	0.1 0.29220897241894
	0.15 0.307523488494074
	0.2 0.312831976647534
	0.3 0.317710180781206
	0.4 0.313459781703306
	0.5 0.308734606006051
};
\end{groupplot}

\end{tikzpicture}
	\end{minipage}%
	\begin{minipage}{0.245\textwidth}
		\begin{tikzpicture}
	
	\definecolor{darkgrey176}{RGB}{176,176,176}
	\definecolor{gainsboro247}{RGB}{253,253,253}
	\definecolor{lightgrey204}{RGB}{204,204,204}
	
	\definecolor{lightgrey204}{RGB}{204,204,204}
	\definecolor{orangered2381020}{RGB}{238,102,0}
	\definecolor{slategrey119119153}{RGB}{119,119,153}
	
	\tikzstyle{every node}=[
	font=\scriptsize,
	]
	
	\begin{groupplot}[
		group style={
			group size=1 by 5,
			vertical sep=0.3cm,
		},
		legend cell align={left},
		legend style={
			fill opacity=0.4,
			draw opacity=1,
			text opacity=1,
			at={(0.32,0.096)},
			anchor=south west,
			draw=none,
			fill=gainsboro247
		},
		width=3.9cm,
		height=3.9cm,
		ytick align=inside,
		grid=both,
		x grid style={darkgrey176},
		y grid style={darkgrey176},
		title style={at={(0.5,1.125)},anchor=north},
		xtick style={draw=none},
		ytick style={draw=none},
		xminorticks=true,
		minor x tick num=1,
		yminorticks=true,
		minor y tick num=1,
		axis background/.style={fill=gainsboro247},
		yticklabel style = {xshift=1.8ex},
		xticklabel style = {yshift=1.2ex},
		y label style={at={(axis description cs:0.18, 0.75)}},
		xticklabels={},
		scaled ticks=false,
		yticklabel style={/pgf/number format/fixed},
		x label style={at={(axis description cs:0.85, 0.1)}},
		]
\nextgroupplot[
tick align=outside,
tick pos=left,
title={\bf \footnotesize LVIS},
x grid style={darkgrey176},
xmin=0.0275, xmax=0.5225,
xtick style={color=black},
y grid style={darkgrey176},
ymin=0.0531760455901073, ymax=0.0625623120430961,
ytick style={color=black},
xlabel={\textcolor{white}{A}},
ytick={0.054,0.057,0.06,0.063},
yticklabels={\tiny0.054,\tiny0.057,\tiny0.060,\tiny0.063},
]
\addplot [line width=1.5pt, opacity=0.75,  black]
table {%
0.05 0.0581980195108916
0.1 0.0615002605042572
0.15 0.0603838800322796
0.2 0.0614704393528882
0.3 0.0594868629963133
0.4 0.0577425291597004
0.5 0.0536026940652432
};
\addplot [line width=1.5pt, opacity=0.75,  slategrey119119153]
table {%
	0.05 0.0586078303214523
	0.1 0.0600971084689279
	0.15 0.061318820098087
	0.2 0.0607445761698374
	0.3 0.0596875261577388
	0.4 0.0579759582962494
	0.5 0.0549661829765793
};
\addplot [line width=1.5pt, opacity=0.75, orangered2381020]
table {%
	0.05 0.0591870569786722
	0.1 0.061544951070851
	0.15 0.0621356635679603
	0.2 0.0615784290255676
	0.3 0.0602405751587019
	0.4 0.0580814241370981
	0.5 0.053985727167371
};

\nextgroupplot[
legend cell align={left},
legend style={fill opacity=0.8, draw opacity=1, text opacity=1, draw=lightgrey204},
tick align=outside,
tick pos=left,
x grid style={darkgrey176},
xmin=0.0275, xmax=0.5225,
xtick style={color=black},
y grid style={darkgrey176},
ymin=0.0384534121267385, ymax=0.0437067293419547,
ytick style={color=black},
xtick={0.1,0.3,0.5},
xticklabels={0.1,0.3,0.5},
xlabel={\textcolor{white}{|}},
ytick={0.036,0.038,0.04,0.042,0.044},
yticklabels={\tiny0.054,\tiny0.057,\tiny 0.040,\tiny0.042},
]
\addplot [line width=1.5pt, opacity=0.75,  black]
table {%
0.05 0.0399114770973109
0.1 0.0425969553969033
0.15 0.0422512099802471
0.2 0.0434289065344712
0.3 0.0425439747923692
0.4 0.0416472863078775
0.5 0.0386921992728847
};
\addplot [line width=1.5pt, opacity=0.75,  slategrey119119153]
table {%
	0.05 0.0404803522142172
	0.1 0.041876293520902
	0.15 0.043262780216393
	0.2 0.0430933047418147
	0.3 0.042659081296169
	0.4 0.0419168062367779
	0.5 0.0400384479412723
};
\addplot [line width=1.5pt, opacity=0.75,  orangered2381020]
table {%
	0.05 0.0406920638426684
	0.1 0.0427416436094721
	0.15 0.0434679421958085
	0.2 0.0433417330543975
	0.3 0.0429859048065179
	0.4 0.0419588716694249
	0.5 0.039019297782703
};
\end{groupplot}

\end{tikzpicture}
	\end{minipage}%
	\begin{minipage}{0.2425\textwidth}
		\begin{tikzpicture}
	
	\definecolor{darkgrey176}{RGB}{176,176,176}
	\definecolor{gainsboro247}{RGB}{253,253,253}
	\definecolor{lightgrey204}{RGB}{204,204,204}
	
	\definecolor{lightgrey204}{RGB}{204,204,204}
	\definecolor{orangered2381020}{RGB}{238,102,0}
	\definecolor{slategrey119119153}{RGB}{119,119,153}
	
	\tikzstyle{every node}=[
	font=\scriptsize,
	]
	
	\begin{groupplot}[
		group style={
			group size=1 by 5,
			vertical sep=0.3cm,
		},
		legend cell align={left},
		legend style={
			fill opacity=0.4,
			draw opacity=1,
			text opacity=1,
			at={(0.32,0.096)},
			anchor=south west,
			draw=none,
			fill=gainsboro247
		},
		width=3.9cm,
		height=3.9cm,
		ytick align=inside,
		grid=both,
		x grid style={darkgrey176},
		y grid style={darkgrey176},
		title style={at={(0.5,1.125)},anchor=north},
		xtick style={draw=none},
		ytick style={draw=none},
		xminorticks=true,
		minor x tick num=1,
		yminorticks=true,
		minor y tick num=1,
		axis background/.style={fill=gainsboro247},
		yticklabel style = {xshift=1.8ex},
		xticklabel style = {yshift=1.2ex},
		y label style={at={(axis description cs:0.18, 0.5)}},
		xticklabels={},
		scaled ticks=false,
		yticklabel style={/pgf/number format/fixed},
		x label style={at={(axis description cs:0.6, 0.1)}},
		]
\nextgroupplot[
tick align=outside,
tick pos=left,
title={\bf \footnotesize BDD},
x grid style={darkgrey176},
xmin=0.0275, xmax=0.5225,
xtick style={color=black},
y grid style={darkgrey176},
ymin=0.201888997650319, ymax=0.28895967706361,
ytick={0.19,0.22,0.25,0.28},
ytick style={color=black}
]
\addplot [line width=1.5pt, opacity=0.75,  black]
table {%
0.05 0.205846755805468
0.1 0.239639458772534
0.15 0.257244091121361
0.2 0.26864038007851
0.3 0.271783876520301
0.4 0.265542044757935
0.5 0.260312090982754
};
\addplot [line width=1.5pt, opacity=0.75,  slategrey119119153]
table {%
0.05 0.206623808625266
0.1 0.238596224466994
0.15 0.256564183601741
0.2 0.267078896387552
0.3 0.270353072202736
0.4 0.266195105676502
0.5 0.260271784280872
};
\addplot [line width=1.5pt, opacity=0.75,  orangered2381020]
table {%
0.05 0.234508923259593
0.1 0.263154869098187
0.15 0.278050937001921
0.2 0.28500191890846
0.3 0.281730561720923
0.4 0.275464576213414
0.5 0.263477929949796
};

\nextgroupplot[
legend cell align={left},
legend style={
  fill opacity=0.8,
  draw opacity=1,
  text opacity=1,
  at={(0.97,0.03)},
  anchor=south east,
  draw=lightgrey204
},
tick align=outside,
tick pos=left,
x grid style={darkgrey176},
xmin=0.0275, xmax=0.5225,
xtick style={color=black},
y grid style={darkgrey176},
ymin=0.110299249846825, ymax=0.169000042818585,
ytick style={color=black},
xtick={0.1,0.3,0.5},
xticklabels={0.1,0.3,0.5},
xlabel={\textcolor{white}{|}},
]
\addplot [line width=1.5pt, opacity=0.75,  black]
table {%
0.05 0.112967467709178
0.1 0.132793060857077
0.15 0.146689198149078
0.2 0.155124271946678
0.3 0.160152339000674
0.4 0.160309544697582
0.5 0.161876967634647
};
\addplot [line width=1.5pt, opacity=0.75,  slategrey119119153]
table {%
0.05 0.114589561262319
0.1 0.134908055267632
0.15 0.148543104039193
0.2 0.155945510973039
0.3 0.160821601479561
0.4 0.161932613106421
0.5 0.162253437940004
};
\addplot [line width=1.5pt, opacity=0.75,  orangered2381020]
table {%
	0.05 0.127245718171345
	0.1 0.146287259929557
	0.15 0.157926677016816
	0.2 0.164610235853631
	0.3 0.166148585238029
	0.4 0.166331824956233
	0.5 0.164254062268937
};
\end{groupplot}

\end{tikzpicture}
	\end{minipage}%
	\vspace{-2.1em}
	\caption{{\bf Downstream Model Evaluation across all Confidence Thresholds and Datasets}.
		Results are average over all VLM models (see \cref{fig:vocmap} right).
	}
	\label{fig:map}
\end{figure}

\begin{figure}[t!]
	\begin{minipage}{0.262\textwidth}
		\begin{tikzpicture}
	
	\definecolor{darkgrey176}{RGB}{176,176,176}
	\definecolor{gainsboro247}{RGB}{253,253,253}
	\definecolor{lightgrey204}{RGB}{204,204,204}
	
	\definecolor{lightgrey204}{RGB}{204,204,204}
	\definecolor{orangered2381020}{RGB}{238,102,0}
	\definecolor{slategrey119119153}{RGB}{119,119,153}
	
	\tikzstyle{every node}=[
	font=\scriptsize,
	]
	
	\begin{groupplot}[
		group style={
			group size=1 by 5,
			vertical sep=0.3cm,
		},
		legend cell align={left},
		legend style={
			fill opacity=0.4,
			draw opacity=1,
			text opacity=1,
			at={(0.16,0.04)},
			anchor=south west,
			draw=none,
			fill=gainsboro247,
			at={(2.54,-0.33)},
			legend columns=3,
			column sep = 1pt,
			fill opacity=0,
		},
		width=3.9cm,
		height=3.9cm,
		ytick align=inside,
		grid=both,
		x grid style={darkgrey176},
		y grid style={darkgrey176},
		title style={at={(0.5,1.125)},anchor=north},
		xtick style={draw=none},
		ytick style={draw=none},
		xminorticks=true,
		minor x tick num=1,
		yminorticks=true,
		minor y tick num=1,
		axis background/.style={fill=gainsboro247},
		yticklabel style = {xshift=1.8ex},
		xticklabel style = {yshift=1.2ex},
		y label style={at={(axis description cs:0.23, 0.5)}},
		xticklabels={},
		x label style={at={(axis description cs:1.1, 0.1)}},
		yticklabel style={/pgf/number format/fixed zerofill,/pgf/number format/fixed},
]
\nextgroupplot[
title={\bf \footnotesize VOC$_v$},
tick align=outside,
tick pos=left,
x grid style={darkgrey176},
xmin=0.0275, xmax=0.5225,
xtick style={color=black},
y grid style={darkgrey176},
ylabel={\bf mAP$_{50}$},
ymin=0.340875001520156, ymax=0.53420074201452,
ytick style={color=black}
]
\addplot [line width=1.5pt, opacity=0.75,  black]
table {%
0.05 0.349662535178991
0.1 0.393196882129064
0.15 0.423431418826823
0.2 0.458567068640171
0.3 0.491912262819532
0.4 0.489096698792596
0.5 0.461562050220932
};
\addplot [line width=1.5pt, opacity=0.75,  slategrey119119153]
table {%
0.05 0.396372808363959
0.1 0.444136472911475
0.15 0.478828411668619
0.2 0.498640955736312
0.3 0.50734837524283
0.4 0.496173895784477
0.5 0.470118604840141
};
\addplot [line width=1.5pt, opacity=0.75,  orangered2381020]
table {%
0.05 0.422014275918334
0.1 0.460597328884738
0.15 0.494106813141464
0.2 0.514326529095923
0.3 0.525413208355686
0.4 0.502681929344607
0.5 0.473919646183176
};

\nextgroupplot[
legend cell align={left},
tick align=outside,
tick pos=left,
x grid style={darkgrey176},
xmin=0.0275, xmax=0.5225,
xtick style={color=black},
y grid style={darkgrey176},
ylabel={\bf mAP$_{50:95}$},
ymin=0.215355812628127, ymax=0.339683676949363,
ytick style={color=black},
xtick={0.1,0.3,0.5},
xticklabels={0.1,0.3,0.5},
xlabel={\bf Confidence Threshold},
]
\addplot [line width=1.5pt, opacity=0.75,  black]
table {%
0.05 0.221007079188183
0.1 0.248703138879237
0.15 0.269607037592722
0.2 0.296986251749032
0.3 0.318644271431761
0.4 0.313277836586202
0.5 0.297294104505178
};
\addlegendentry{\bf \scriptsize AL \cite{griff25AL}}
\addplot [line width=1.5pt, opacity=0.75,  slategrey119119153]
table {%
0.05 0.249490685983973
0.1 0.279356940920522
0.15 0.302131761367741
0.2 0.320893918291051
0.3 0.326249911099459
0.4 0.317813882226893
0.5 0.300942813798633
};
\addlegendentry{\bf  \scriptsize TTN}
\addplot [line width=1.5pt, opacity=0.75,  orangered2381020]
table {%
0.05 0.262211032646304
0.1 0.291098067899591
0.15 0.313549954867963
0.2 0.330867795519972
0.3 0.334032410389306
0.4 0.323794292632522
0.5 0.303561565149575
};
\addlegendentry{\bf  \scriptsize TTN$_{\text{\tiny D}}$}
\end{groupplot}

\end{tikzpicture}
	\end{minipage}%
	\begin{minipage}{0.24\textwidth}
		\begin{tikzpicture}
	
	\definecolor{darkgrey176}{RGB}{176,176,176}
	\definecolor{gainsboro247}{RGB}{253,253,253}
	\definecolor{lightgrey204}{RGB}{204,204,204}
	
	\definecolor{lightgrey204}{RGB}{204,204,204}
	\definecolor{orangered2381020}{RGB}{238,102,0}
	\definecolor{slategrey119119153}{RGB}{119,119,153}
	
	\tikzstyle{every node}=[
	font=\scriptsize,
	]
	
	\begin{groupplot}[
		group style={
			group size=1 by 5,
			vertical sep=0.3cm,
		},
		legend cell align={left},
		legend style={
			fill opacity=0.4,
			draw opacity=1,
			text opacity=1,
			at={(0.16,-0.01)},
			anchor=south west,
			draw=none,
			fill=gainsboro247
		},
		width=3.9cm,
		height=3.9cm,
		ytick align=inside,
		grid=both,
		x grid style={darkgrey176},
		y grid style={darkgrey176},
		title style={at={(0.5,1.125)},anchor=north},
		xtick style={draw=none},
		ytick style={draw=none},
		xminorticks=true,
		minor x tick num=1,
		yminorticks=true,
		minor y tick num=1,
		axis background/.style={fill=gainsboro247},
		yticklabel style = {xshift=1.8ex},
		xticklabel style = {yshift=1.2ex},
		y label style={at={(axis description cs:0.23, 0.5)}},
		xticklabels={},
		x label style={at={(axis description cs:1.18, 0.1)}},
		scaled ticks=false,
		yticklabel style={/pgf/number format/fixed zerofill,/pgf/number format/fixed},
		]
\nextgroupplot[
tick align=outside,
tick pos=left,
title={\bf \footnotesize COCO$_v$},
x grid style={darkgrey176},
xmin=0.0275, xmax=0.5225,
xtick style={color=black},
y grid style={darkgrey176},
ymin=0.143342206907118, ymax=0.202279985780532,
ytick style={color=black}
]
\addplot [line width=1.5pt, opacity=0.75,  black]
table {%
0.05 0.14602119685591
0.1 0.156813698006951
0.15 0.170175052293
0.2 0.181634211261356
0.3 0.194152000033028
0.4 0.192751321522289
0.5 0.184053323565268
};
\addplot [line width=1.5pt, opacity=0.75,  slategrey119119153]
table {%
0.05 0.158703233530696
0.1 0.176009481993604
0.15 0.188695542714761
0.2 0.196726747547
0.3 0.19859011227733
0.4 0.19391577907616
0.5 0.184352249977896
};
\addplot [line width=1.5pt, opacity=0.75,  orangered2381020]
table {%
0.05 0.156356354119693
0.1 0.174358686110912
0.15 0.184358865200059
0.2 0.195275888951922
0.3 0.199600995831741
0.4 0.19356148576311
0.5 0.186021059047108
};

\nextgroupplot[
legend cell align={left},
legend style={
  fill opacity=0.8,
  draw opacity=1,
  text opacity=1,
  at={(0.97,0.03)},
  anchor=south east,
  draw=lightgrey204
},
tick align=outside,
tick pos=left,
x grid style={darkgrey176},
xmin=0.0275, xmax=0.5225,
xtick style={color=black},
y grid style={darkgrey176},
ymin=0.0841940902075503, ymax=0.125288674447807,
ytick style={color=black},
xtick={0.1,0.3,0.5},
xticklabels={0.1,0.3,0.5},
xlabel={\textcolor{white}{|}},
ytick={0.09,0.1,0.11,0.12},
yticklabels={0.09,0.10,0.11,0.12},
]
\addplot [line width=1.5pt, opacity=0.75,  black]
table {%
0.05 0.0860620258548347
0.1 0.0944204920416744
0.15 0.102118117146425
0.2 0.110208537846864
0.3 0.119065370780578
0.4 0.117660084144688
0.5 0.11301168569063
};
\addplot [line width=1.5pt, opacity=0.75,  slategrey119119153]
table {%
0.05 0.0939304865724167
0.1 0.106788165342595
0.15 0.1147107677028
0.2 0.121215142874315
0.3 0.123420738800522
0.4 0.119004570030791
0.5 0.113563634662386
};
\addplot [line width=1.5pt, opacity=0.75,  orangered2381020]
table {%
0.05 0.0932700438812619
0.1 0.104372631039513
0.15 0.112250666940729
0.2 0.119171114036913
0.3 0.121836577112429
0.4 0.11939337386581
0.5 0.114527220860394
};
\end{groupplot}

\end{tikzpicture}
	\end{minipage}%
	\begin{minipage}{0.243\textwidth}
		\begin{tikzpicture}
	
	\definecolor{darkgrey176}{RGB}{176,176,176}
	\definecolor{gainsboro247}{RGB}{253,253,253}
	\definecolor{lightgrey204}{RGB}{204,204,204}
	
	\definecolor{lightgrey204}{RGB}{204,204,204}
	\definecolor{orangered2381020}{RGB}{238,102,0}
	\definecolor{slategrey119119153}{RGB}{119,119,153}
	
	\tikzstyle{every node}=[
	font=\scriptsize,
	]
	
	\begin{groupplot}[
		group style={
			group size=1 by 5,
			vertical sep=0.3cm,
		},
		legend cell align={left},
		legend style={
			fill opacity=0.4,
			draw opacity=1,
			text opacity=1,
			at={(0.32,0.096)},
			anchor=south west,
			draw=none,
			fill=gainsboro247
		},
		width=3.9cm,
		height=3.9cm,
		ytick align=inside,
		grid=both,
		x grid style={darkgrey176},
		y grid style={darkgrey176},
		title style={at={(0.5,1.125)},anchor=north},
		xtick style={draw=none},
		ytick style={draw=none},
		xminorticks=true,
		minor x tick num=1,
		yminorticks=true,
		minor y tick num=1,
		axis background/.style={fill=gainsboro247},
		yticklabel style = {xshift=1.8ex},
		xticklabel style = {yshift=1.2ex},
		y label style={at={(axis description cs:0.18, 0.75)}},
		xticklabels={},
		scaled ticks=false,
		yticklabel style={/pgf/number format/fixed},
		x label style={at={(axis description cs:0.85, 0.1)}},
		yticklabel style={/pgf/number format/fixed zerofill,/pgf/number format/fixed},
		]
\nextgroupplot[
tick align=outside,
tick pos=left,
title={\bf \footnotesize LVIS$_v$},
x grid style={darkgrey176},
xmin=0.0275, xmax=0.5225,
xtick style={color=black},
y grid style={darkgrey176},
ymin=-0.000104453261975445, ymax=0.00219351850148435,
ytick style={color=black},
xtick={0.1,0.3,0.5},
xlabel={\textcolor{white}{A}},
ytick={0.000,0.001,0.002},
yticklabels={\tiny 0.000,\tiny0.001,\tiny0.002},
]
\addplot [line width=1.5pt, opacity=0.75,  black]
table {%
0.05 0.00208906523950891
0.1 0
0.15 0.000239874537972681
0.2 0.000490976637922128
0.3 0.00100168737659618
0.4 0.00104423582273441
0.5 0.000289379174867042
};
\addplot [line width=1.5pt, opacity=0.75,  slategrey119119153]
table {%
0.05 0.00154722607050333
0.1 0.000668946893219315
0.15 0.000780011438139469
0.2 0.00102688244863727
0.3 0.000417510500916318
0.4 0.000585164970549372
0.5 0.00119853726671791
};
\addplot [line width=1.5pt, opacity=0.75, orangered2381020]
table {%
0.05 0.000783953566496866
0.1 0.00158497084183092
0.15 0.000294200376277889
0.2 0.000748175646591479
0.3 0.000650187104207623
0.4 0.000769072374042443
0.5 0.00037500430635346
};

\nextgroupplot[
legend cell align={left},
legend style={fill opacity=0.8, draw opacity=1, text opacity=1, draw=lightgrey204},
tick align=outside,
tick pos=left,
x grid style={darkgrey176},
xmin=0.0275, xmax=0.5225,
xtick style={color=black},
y grid style={darkgrey176},
ymin=-7.51335864796703e-05, ymax=0.00157780531607308,
ytick style={color=black},
xtick={0.1,0.3,0.5},
xticklabels={0.1,0.3,0.5},
xlabel={\textcolor{white}{|}},
ytick={0.0,0.001,0.002},
yticklabels={\tiny 0.000,\tiny0.001,\tiny0.002},
]
\addplot [line width=1.5pt, opacity=0.75,  black]
table{
0.05 0.00150267172959341
0.1 0
0.15 0.000166723907602604
0.2 0.000420790781798054
0.3 0.000886001013476093
0.4 0.000943508853119702
0.5 0.000245972762710434
};
\addplot [line width=1.5pt, opacity=0.75,  slategrey119119153]
table {%
0.05 0.00101910420275278
0.1 0.000487374334652836
0.15 0.00066381113474297
0.2 0.000878574142354928
0.3 0.000330872011616987
0.4 0.000472671993272848
0.5 0.00108723953458148
};
\addplot [line width=1.5pt, opacity=0.75,  orangered2381020]
table {%
0.05 0.000670545450281195
0.1 0.00129223379347319
0.15 0.000198694333049845
0.2 0.000465483920680326
0.3 0.000503398385946965
0.4 0.000657337407189489
0.5 0.000329260768689342
};
\end{groupplot}

\end{tikzpicture}
	\end{minipage}%
	\begin{minipage}{0.242\textwidth}
		\begin{tikzpicture}
	
	\definecolor{darkgrey176}{RGB}{176,176,176}
	\definecolor{gainsboro247}{RGB}{253,253,253}
	\definecolor{lightgrey204}{RGB}{204,204,204}
	
	\definecolor{lightgrey204}{RGB}{204,204,204}
	\definecolor{orangered2381020}{RGB}{238,102,0}
	\definecolor{slategrey119119153}{RGB}{119,119,153}
	
	\tikzstyle{every node}=[
	font=\scriptsize,
	]
	
	\begin{groupplot}[
		group style={
			group size=1 by 5,
			vertical sep=0.3cm,
		},
		legend cell align={left},
		legend style={
			fill opacity=0.4,
			draw opacity=1,
			text opacity=1,
			at={(0.32,0.096)},
			anchor=south west,
			draw=none,
			fill=gainsboro247
		},
		width=3.9cm,
		height=3.9cm,
		ytick align=inside,
		grid=both,
		x grid style={darkgrey176},
		y grid style={darkgrey176},
		title style={at={(0.5,1.125)},anchor=north},
		xtick style={draw=none},
		ytick style={draw=none},
		xminorticks=true,
		minor x tick num=1,
		yminorticks=true,
		minor y tick num=1,
		axis background/.style={fill=gainsboro247},
		yticklabel style = {xshift=1.8ex},
		xticklabel style = {yshift=1.2ex},
		y label style={at={(axis description cs:0.18, 0.5)}},
		xticklabels={},
		scaled ticks=false,
		yticklabel style={/pgf/number format/fixed},
		x label style={at={(axis description cs:0.6, 0.1)}},
		yticklabel style={/pgf/number format/fixed zerofill,/pgf/number format/fixed},
		]
\nextgroupplot[
tick align=outside,
tick pos=left,
title={\bf \footnotesize BDD$_v$},
x grid style={darkgrey176},
xmin=0.0275, xmax=0.5225,
xtick style={color=black},
y grid style={darkgrey176},
ymin=0.00287982447230311, ymax=0.0195259701215984,
ytick={0.0,0.005,0.01,0.015,0.02},
yticklabels={\tiny0.0,\tiny0.005,\tiny0.010,\tiny0.015},
ytick style={color=black}
]
\addplot [line width=1.5pt, opacity=0.75,  black]
table {%
0.05 0.00375583606967493
0.1 0.00542402388326594
0.15 0.00663931588122533
0.2 0.00765204067323682
0.3 0.00712078073018423
0.4 0.00802212892839601
0.5 0.00994754457808801
};
\addplot [line width=1.5pt, opacity=0.75,  slategrey119119153]
table {%
0.05 0.00363646745636199
0.1 0.0056380227405784
0.15 0.00685638237597705
0.2 0.00702092967220141
0.3 0.00795923745507675
0.4 0.00749134837107178
0.5 0.00559231100871265
};
\addplot [line width=1.5pt, opacity=0.75,  orangered2381020]
table {%
0.05 0.0126681316844506
0.1 0.0120828659761263
0.15 0.0158788570481124
0.2 0.0187693271375395
0.3 0.00967150133596694
0.4 0.0165551118734931
0.5 0.00611329906983562
};

\nextgroupplot[
legend cell align={left},
legend style={
  fill opacity=0.8,
  draw opacity=1,
  text opacity=1,
  at={(0.97,0.03)},
  anchor=south east,
  draw=lightgrey204
},
tick align=outside,
tick pos=left,
x grid style={darkgrey176},
xmin=0.0275, xmax=0.5225,
xtick style={color=black},
y grid style={darkgrey176},
ymin=0.00115797165235057, ymax=0.0105857072806849,
ytick style={color=black},
xtick={0.1,0.3,0.5},
xticklabels={0.1,0.3,0.5},
xlabel={\textcolor{white}{|}},
ytick={-0.002,0.002,0.006,0.01},
yticklabels={\tiny0.00,\tiny0.002,\tiny0.006,\tiny0.010,},
]
\addplot [line width=1.5pt, opacity=0.75,  black]
table {%
0.05 0.00167268911833001
0.1 0.00241298539895842
0.15 0.00312902545314092
0.2 0.00378737562416271
0.3 0.00342340915319404
0.4 0.00420125641953512
0.5 0.00557521086837139
};
\addplot [line width=1.5pt, opacity=0.75,  slategrey119119153]
table {%
0.05 0.00158650509000212
0.1 0.00251683203344697
0.15 0.00315244175159329
0.2 0.00345486476812226
0.3 0.00384544323902563
0.4 0.00373650455325171
0.5 0.00290350556726956
};
\addplot [line width=1.5pt, opacity=0.75,  orangered2381020]
table {%
0.05 0.00532429772607402
0.1 0.00563265038566981
0.15 0.00768609382615549
0.2 0.0101571738430333
0.3 0.00475100726080307
0.4 0.00836609607032111
0.5 0.00324938792183686
};
\end{groupplot}

\end{tikzpicture}
	\end{minipage}%
	\vspace{-2.1em}
	\caption{{\bf Downstream Vulnerable-Class Evaluation across all Confidence Thresholds and Datasets}.
		Results are average over VLM models (see \cref{fig:vocmap} right).
	}
	\label{fig:mapv}
\end{figure}
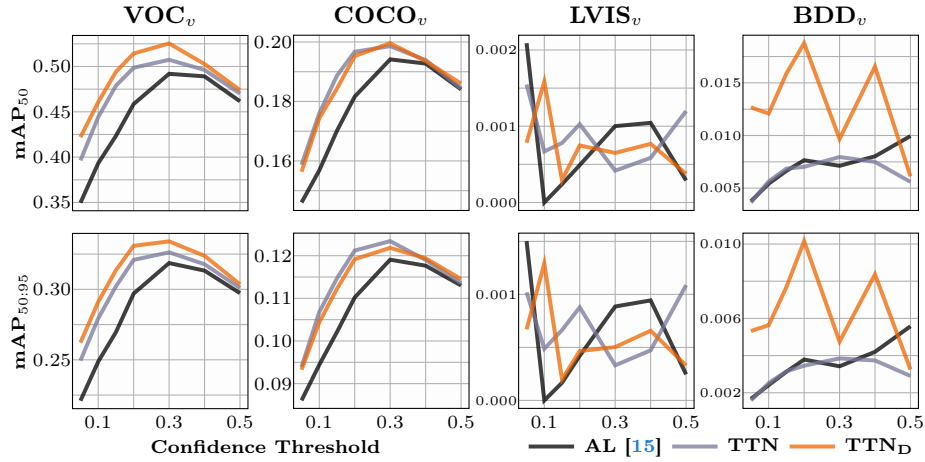

\section{Experimental Protocol}
\label{sec:supsetup}

Our experimental protocol is motivated by the comprehensive \textbf{Auto-Labeling} (AL) benchmark \cite{griff25AL}.
To supplement the initial experiment setup in \cref{sec:setup}, we provide detailed descriptions of our experimental protocol for reproducability.

\subsection{Vision-Language Model Details}
\label{sec:vlm}

We use three state-of-the-art vision-language models (VLMs) for our pseudo-labeling experiments.
YOLO-World (YOLOW) \cite{Cheng_2024_CVPR} and YOLOE \cite{wang2025yoloerealtimeseeing} are both pre-trained on Objects365 \cite{Shao_2019_ICCV}, GQA \cite{Hudson_2019_CVPR}, and Flickr30k \cite{young14} and implemented via Ultralytics \cite{jocher2023yolo}.
Grounding DINO-Tiny (GDINO)  \cite{shilong2024} is pre-trained on Objects365, GoldG \cite{Kamath_2021_ICCV}, and Cap4M \cite{Li_2022_CVPR} and implemented via Hugging Face \cite{wolf-etal-2020-transformers}.
Notably, there is no direct data leakage between these VLM pre-training data and the object detection datasets in \cref{sec:detectdata}.

\subsection{Object Detection Dataset Details}
\label{sec:detectdata}

For experiments and evaluation, we use the object detection datasets listed in \cref{tab:detectdatasetcost}, which vary in terms of application complexity and domain.
All four datasets were originally labeled by human annotators, which enables us to directly evaluate pseudo-labels (\cref{sec:suplabel}) and---after training lightweight detectors only on those pseudo-labels---downstream model performance on the unseen validation sets (\cref{sec:supdown}).

For the PASCAL {\bf Visual Object Classes} Dataset (VOC) \cite{EvEtAl10}, we combine the original train and validation splits from the 2007 and 2012 challenges as a single train set for pseudo-labeling (16,551 images); after training models on pseudo-labels, we then use the original 2007 test split (4,952 images) for validation.
For the Microsoft {\bf Common Objects in COntext} (COCO) \cite{coco}, {\bf Large Vocabulary Instance Segmentation} (LVIS) \cite{Gupta_2019_CVPR}, and {\bf Berkeley DeepDrive} (BDD) datasets \cite{Yu_2020_CVPR}, we use the standard train splits for pseudo-labeling and downstream model training then the standard validation splits for subsequent model evaluation.
For our detection-based evaluation, we do not use COCO crowd instances and all segmentation labels are converted to bounding boxes.

\vspace{0.2em}
\noindent{\bf Remarks on LVIS}.
Two properties unique to LVIS require accommodation.
First, while the training split comprises 1,203 classes, only 1,035 are represented in the validation split.
Thus, we pseudo-label all 1,203 classes for pseudo-label evaluation and downstream model training, but the downstream model evaluation for mAP includes only the 1,035 classes available for validation.

Second, LVIS uses verbose descriptions to differentiate its numerous classes.
For example, class 242 is ``chili/chili vegetable/chili pepper/chili pepper vegetable/chilli/chilli vegetable/chilly/chilly vegetable/chile/chile vegetable.''
Unfortunately, when prompted by LVIS's 1,203 verbose class descriptions, GDINO runs out of memory due to architectural constraints (additional details in the AL benchmark paper \cite{griff25AL}).
For this reason, we omit GDINO from LVIS experiments.

\setlength{\tabcolsep}{2.9pt}
\begin{table} [t]
	\centering
	\caption{{\bf Object Detection Datasets}. Labeling Cost based on AWS SageMaker.
		The total cost to pseudo-label all of the detection datasets with YOLOW is \$1.18 \cite{griff25AL}.
	}
	\vspace{-0.75em}
	\begin{tabular}{| c | r | r |r | l | }
		\hline 
		\rowcolor{tableheader} & \multicolumn{2}{ c |}{\bf Number of}   &  \multicolumn{1}{ c |}{\bf Labeling} & \\ 
		\rowcolor{tableheader}  \multicolumn{1}{| c |}{\bf Dataset} & \multicolumn{1}{ c }{\bf Classes} & \multicolumn{1}{ c |}{\bf Objects} & \multicolumn{1}{ c |}{\bf Cost} & \multicolumn{1}{ c |}{\bf Application} \\ \hline
		VOC	\cite{EvEtAl10} &	20	&	40,058	&	\$1,442	&	Basic object categories for web images	\\ \hline
		COCO \cite{coco}	&	80	&	849,945	&	\$30,598	&	Common objects, moderate complexity	\\ \hline
		LVIS \cite{Gupta_2019_CVPR}	&	1,203	&	1,270,141	&	\$45,725	&	Large vocabulary, high complexity	\\ \hline
		BDD \cite{Yu_2020_CVPR}	&	10	&	1,286,871	&	\$46,327 &	Autonomous driving views \& objects	\\ \hline
	\end{tabular}
	\label{tab:detectdatasetcost}
\end{table}

\subsection{Pseudo-Label Evaluation Details}
\label{sec:suplabel}

We directly evaluate pseudo-labels relative to ground truth on all four of the object detection datasets.
Specifically, we use
\begin{align}
	\text{Recall} &= \frac{\text{TP}}{\text{TP+FN}}~\in~[0,1], \\
	\text{Precision} &= \frac{\text{TP}}{\text{TP+FP}}~\in~[0,1], \label{eq:precision} \\
	F_1~\text{Score} &= 2\frac{\text{Precision}\cdot\text{Recall}}{\text{Precision + Recall}} = \frac{\text{2TP}}{\text{2TP + FN + FP}} \in [0,1], \\
	F_2~\text{Score} &= \frac{5\cdot\text{Precision}\cdot\text{Recall}}{4\cdot\text{Precision + Recall}} = \frac{\text{5TP}}{\text{5TP + 4FN + FP}} \in [0,1],
\end{align}
where true positives ({\footnotesize TP}) is the number of pseudo-labels with correct class label \textit{and} Intersection over Union (IoU) $>0.5$, false positives ({\footnotesize FP}) is the number of pseudo-labels failing the {\footnotesize TP} criteria, and false negatives ({\footnotesize FN}) is the number of ground-truth labels without a corresponding {\footnotesize TP}.
Notably, if $\text{\footnotesize TP, FP}=0$ in \cref{eq:precision}, $\text{\footnotesize Precision}=1$ if $\text{\footnotesize FN}=0$ and $\text{\footnotesize Precision}=0$ if $\text{\footnotesize FN}>0$.
All pseudo-label evaluation metrics are calculated on a per-class basis then averaged together.

\subsection{Downstream Model Evaluation Details}
\label{sec:supdown}

For each dataset, we train a YOLO11n detector on the unlabeled training images paired with pseudo-labels for 100 epochs using the Ultralytics framework \cite{yolo11_ultralytics}. 
Using the final training weights, we evaluate these detectors on their respective human-labeled validation sets via the \textit{mean} average precision, defined as
\begin{equation}
	\text{mAP}_{50} = \frac{1}{| \mathcal{C} |} \sum_{c_i \in \mathcal{C}} \text{AP}_{50}^{(c_i)},
	\label{eq:map}
\end{equation}
where the \textit{average precision} AP$_{50}^{(c_i)}$ is taken over a discretized precision--recall curve with $\text{IoU}>0.5$ for each object class $c_i$ \cite{EvEtAl10}. 
Similarly, the mean average precision over the COCO \cite{coco} primary metric range is defined as 
\begin{equation}
	\text{mAP}_{50:95} = \frac{1}{10} \sum_{k=0}^{9} \text{mAP}_{\text{IoU} \, = \, 0.5 + 0.05k},
\end{equation}
where each constituent $\text{mAP}_{\text{IoU}}$ is calculated according to \cref{eq:map} using the corresponding intersection-over-union threshold.

\end{document}